\documentclass[pdflatex,sn-mathphys-num]{sn-jnl}

\usepackage[T1]{fontenc}%
\usepackage[utf8]{inputenc}%
\usepackage{microtype}%
\usepackage{inconsolata}%

\usepackage{graphicx}%
\usepackage{amsmath,amssymb,amsfonts}%
\usepackage{mathtools}%
\usepackage{amsthm}%
\usepackage{booktabs}%
\usepackage{multirow}%
\usepackage{makecell}%
\usepackage{array}%
\usepackage{adjustbox}%
\usepackage[table]{xcolor}%
\usepackage{nicefrac}%
\usepackage{mleftright}%
\usepackage{textcomp}%
\usepackage{listings}%
\usepackage{alltt}%
\usepackage{fvextra}%
\usepackage{enumitem}%
\usepackage[most]{tcolorbox}%
\usepackage{subcaption}%
\usepackage{float}%
\usepackage{placeins}%
\usepackage{algorithm}%
\usepackage{algpseudocode}%

\tcbuselibrary{breakable,skins,listings}

\makeatletter
\newenvironment{snwidetable}[1][]{\begin{tableorg}[#1]\centering}{\end{tableorg}}
\makeatother

\definecolor{techXML}{HTML}{1F77B4}
\definecolor{techCOT}{HTML}{FF7F0E}
\definecolor{techVAL}{HTML}{D62728}
\definecolor{techLOG}{HTML}{17BECF}
\definecolor{techDIS}{HTML}{9467BD}
\definecolor{techPERF}{HTML}{2CA02C}

\newtcolorbox{legendbox}{
  enhanced,
  colback=white,
  colframe=black!35,
  boxrule=0.6pt,
  arc=1.2mm,
  left=5pt,right=5pt,top=4pt,bottom=4pt,
}

\theoremstyle{plain}

\theoremstyle{definition}

\theoremstyle{remark}

\begin{document}


\title[TAB-PO]{TAB-PO: Preference Optimization with a Token-Level Adaptive Barrier for Token-Critical Structured Generation}

\author[1]{\fnm{Samah} \sur{Fodeh}}\email{samah.fodeh@yale.edu}

\author[1]{\fnm{Linhai} \sur{Ma}}\email{linhai.ma@yale.edu}

\author[1]{\fnm{Ganesh} \sur{Puthiaraju}}\email{ganesh.puthiaraju@yale.edu}

\author[1]{\fnm{Srivani} \sur{Talakokkul}}\email{srivani.talakokkul@yale.edu}

\author[1]{\fnm{Afshan} \sur{Khan}}\email{afshan.khan@yale.edu}

\author[1]{\fnm{Sreeraj} \sur{Ramachandran}}\email{sreeraj.ramachandran@yale.edu}

\author[1]{\fnm{Elyas} \sur{Irankhah}}\email{elyas.irankhah@yale.edu}

\author[2]{\fnm{Aimee Kendall} \sur{Roundtree}}\email{akr@txstate.edu}

\affil[1]{\orgname{Yale University}, \orgaddress{\city{New Haven}, \state{CT}, \country{USA}}}

\affil[2]{\orgname{Texas State University}, \orgaddress{\city{San Marcos}, \state{TX}, \country{USA}}}

\abstract{Direct Preference Optimization (DPO) is effective for offline alignment but poorly matched to ontology-driven structured prediction, where preferred and rejected JSON often differ by only a few schema-defining tokens. In this low-edit-distance regime, sequence-level DPO spreads gradient mass across non-critical serialization tokens (\emph{gradient dilution}) and can reduce likelihood on rare preferred schema tokens (\emph{token erosion}). To address these limitations, we first develop a confusion-aware preference-construction strategy combining expert-curated ambiguity patterns with validation-set SFT structured-error modes, producing minimally perturbed, schema-valid negatives for ontology-level decision errors. We then introduce Token-Adaptive Barrier Preference Optimization (TAB-PO), a post-SFT objective for token-critical structured generation with a confidence-gated token-level barrier that anchors under-confident schema tokens. On SciERC, with Llama/Qwen models, TAB-PO improves ontology-critical \emph{semantic-label} and \emph{relational-linking} metrics by 11.59\% relative to SFT, wins 100\% of comparisons against strongest token-level/sequence-level DPO variants, and surpasses  strongest frontier baselines on these metrics by 14.71\% relative while improving \emph{textual grounding}.}


\maketitle


\section{Introduction}

Structured prediction lies at the core of many high-value language technologies.
In these settings, a model must generate outputs that satisfy explicit
structural constraints, respect ontology or schema dependencies, and often
ground each decision in supporting evidence from the input \cite{schmidt2025grammar}.
This requirement arises across information extraction, where errors can cause
missed or incorrectly normalized evidence \cite{leaman2015challenges}; legal
document analysis, where errors can introduce compliance risk
\cite{cejas2023nlp}; and biomedical NLP, where incorrect labels or links can
propagate scientific noise \cite{he2024prompt}. Such errors are costly because
they distort downstream evidence, violate task constraints, and weaken decision
support \cite{demnerfushman2009nlp}.

Supervised fine-tuning teaches the model the target schema and valid output
format, but many residual errors occur at hard decision boundaries between
plausible schema-valid alternatives, such as closely related ontology labels,
competing evidence spans, or alternative relational links
\cite{sainz2024gollie,luan2018multi}. Preference optimization is therefore a
natural post-SFT stage: it can contrast ambiguous correct and incorrect
structured outputs and sharpen the model's choices
\cite{ouyang2022training,rafailov2023direct}.

Direct Preference Optimization (DPO) has become a widely used approach for
post-supervised alignment of large language models, replacing explicit reward
modeling and online reinforcement learning with an offline Bradley--Terry
objective that increases the relative likelihood of preferred responses over
rejected responses \cite{rafailov2023direct}. DPO and its variants have been
effective for open-ended generation tasks such as instruction following,
dialogue, and summarization
\cite{ouyang2022training,bai2022constitutional,tunstall2023zephyr,
ivison2023camels,wang2023far}. Ontology-driven structured prediction, however,
has a different error geometry. Preferred and rejected outputs are often
near-identical serialized records that differ in only a few schema-defining
tokens while sharing most JSON scaffolding, field names, and formatting tokens.

This low-separation regime exposes a mismatch between sequence-level preference
objectives and token-critical structured correctness. Standard DPO can dilute
learning signal across non-critical serialization tokens rather than
concentrating it on the sparse tokens that determine semantic labels, textual
groundings, or relational links; we refer to this mismatch as \emph{gradient
dilution}. It can also improve the aggregate preferred-over-rejected margin
while reducing the likelihood of under-confident preferred schema tokens, a
failure mode we call \emph{preferred-token erosion}. Recent DPO variants address
sequence-level pathologies such as calibration drift, reference dependence, or
positive-likelihood degradation
\cite{gheshlaghiAzar2024psiPO,meng2024simpo,xiao2024caldpo,pal2024smaug},
and token-level variants introduce finer-grained preference signals
\cite{zeng2024token,yang2025token}. However, these methods do not explicitly
target the low-edit-distance, ontology-constrained preferences that arise in
structured extraction.

Addressing this regime requires preference pairs that reflect realistic
ontology-level errors and an objective that focuses optimization on the few
tokens that decide structured correctness while still improving the
preferred-over-rejected preference margin. Our contributions are as follows:

\begin{figure}[!t]
  \centering
  \includegraphics[width=0.99\linewidth]{ACL_workflowfinal.pdf}
  \caption{Overview of the TAB-PO framework for ontology-driven structured prediction. A modular prompt-engineered interface and completion-only supervised fine-tuning teach the model to generate schema-formatted structured records, but residual SFT errors can remain in token-critical decisions such as semantic labelling, relational linking and textual grounding. TAB-PO targets these errors using confusion-aware and expert-guided hard
negatives together with a confidence-gated token-level SFT-style likelihood
barrier, yielding outputs with resolved semantic and relational ambiguity and
more precise source grounding.}
  \label{fig:tabpo-pipeline}
\end{figure}

\begin{itemize}[leftmargin=*,itemsep=2pt,topsep=3pt,parsep=0pt,partopsep=0pt]

\item \textbf{We introduce a general task formulation for ontology-driven structured prediction.}
We define a task-agnostic structured-output setting in which correctness depends
on semantic labelling, textual grounding, and relational linking under a
task-specific ontology (Section~\ref{sec:general_task_formulation}).

\item \textbf{We mathematically characterize why sequence-level DPO is mismatched to low-edit-distance structured preferences.}
Through a token-level decomposition of the DPO gradient, we show that standard
DPO can allocate update mass to non-critical serialization tokens rather than
the sparse tokens that determine structured correctness, leading to
\emph{gradient dilution}. We further show that the aggregate
preferred-over-rejected margin can improve even when the likelihood of rare or
under-confident preferred schema tokens decreases, causing
\emph{preferred-token erosion} (Appendix~\ref{app:gradient_dilution}).

\item \textbf{We construct confusion-aware structured preference pairs.}
We create minimally perturbed, schema-valid hard negatives from empirical
SFT ontology-level confusions and expert-adjudicated ambiguity patterns,
targeting realistic semantic-label, evidence-grounding, and relational-linking
errors (Section~\ref{sec:confusion_pref}).

\item \textbf{We propose Token-Adaptive Barrier Preference Optimization.}
TAB-PO augments preference optimization with a confidence-gated SFT-style
likelihood regularizer on under-confident preferred tokens. Together with
confusion-aware preferences, TAB-PO reduces gradient dilution, protects
vulnerable ontology-bound tokens from likelihood erosion, and preserves the
structural scaffolding required for valid serialized outputs
(Section~\ref{sec:tabpo}).

\item \textbf{We provide comprehensive empirical and diagnostic evidence.}
Across PV-Miner and SciERC, and across Llama and Qwen models from 1.5B to 70B
parameters, TAB-PO consistently improves over SFT and outperforms strong
sequence-level and token-level DPO-family baselines. Mechanistic diagnostics
further substantiate these gains by showing that TAB-PO concentrates
optimization on schema-critical tokens while maintaining preference separation
(Section~\ref{sec:experiments_results}; Section~\ref{sec:mechanistic_diagnostics}).

\end{itemize}

Figure~\ref{fig:tabpo-pipeline} provides an overview of the TAB-PO framework, showing how the modular prompt-engineered interface, supervised structured initialization, confusion-aware preference construction, and token-adaptive barrier objective work together for ontology-driven structured prediction.

\begin{figure}[!t]
    \centering
    \includegraphics[width=0.99\linewidth]{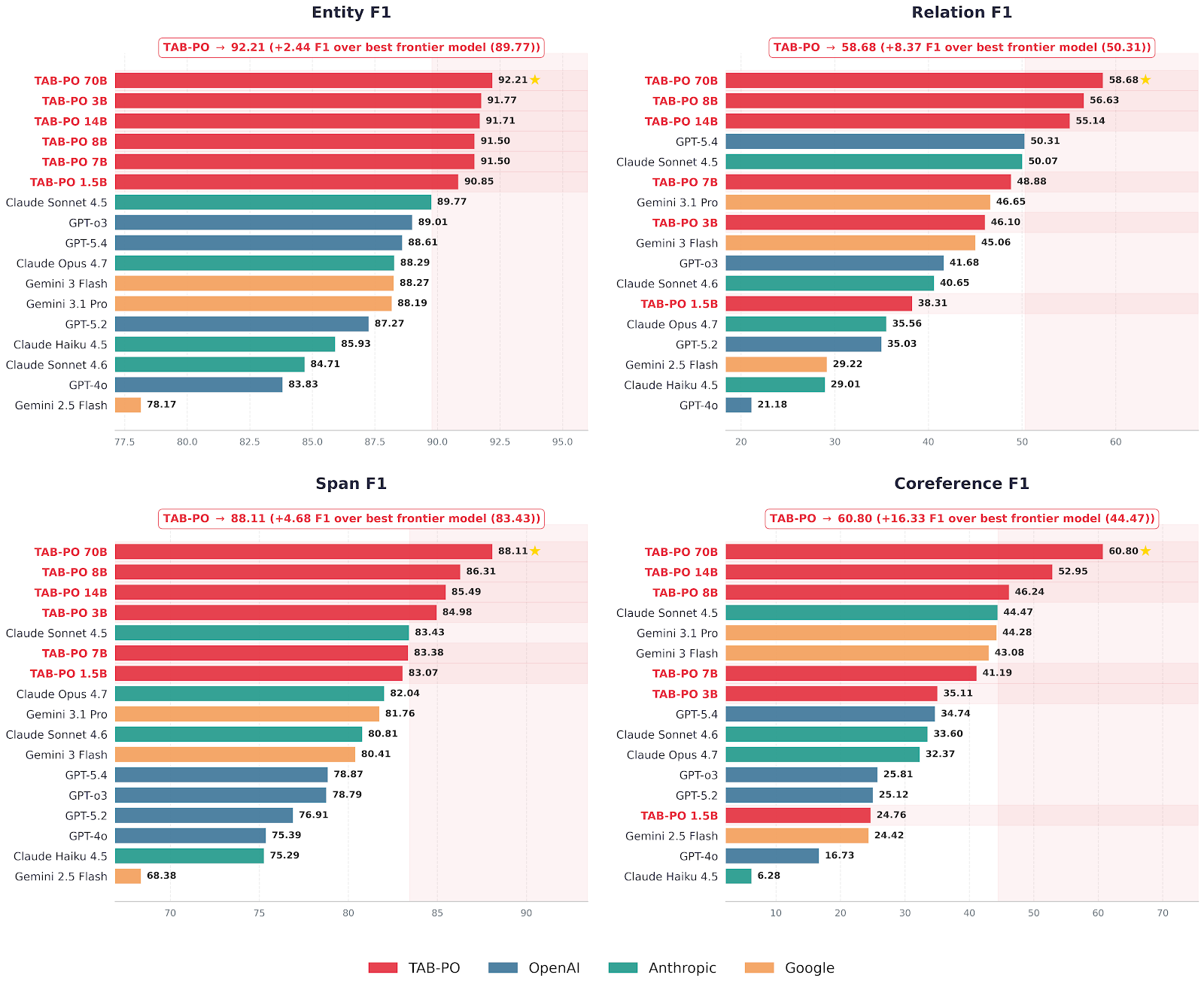}
\caption{
SciERC F1 comparison between TAB-PO models and frontier models across Entity, Relation, Span, and Coreference extraction. 
TAB-PO 70B achieves the best performance across all four evaluation dimensions, with particularly large gains on Relation F1 and Coreference F1, the most structurally demanding SciERC tasks.
These results demonstrate the effectiveness of token-adaptive barrier preference optimization for ontology-constrained structured prediction.
}
    \label{fig:tabpo_frontier_scierc_f1}
\end{figure}

\section{General Task Formulation for Ontology-Driven Structured Prediction}
\label{sec:general_task_formulation}

We introduce a general formulation for ontology-driven structured prediction,
covering tasks in which the desired output is not free-form text but a
schema-valid set of records constrained by a task-specific ontology. Given an
input sequence $s$, optional task metadata $m$, and an ontology $\mathcal{O}$,
the goal is to predict a finite structured output
$\widehat{Y}=\{y_1,y_2,\dots,y_N\}$. The generation model $f_\theta$ maps
\begin{equation}
    f_\theta : (s, m, \mathcal{O}) \;\longmapsto\;
    \widehat{Y} = \{y_1, y_2, \dots, y_N\}.
\end{equation}

Each predicted record has the abstract form
$y_i=(\ell_i,\Pi_i,R_i)$, where the three components correspond to
semantic labelling, textual grounding, and relational linking.

\paragraph{Semantic label.}
The semantic-label component $\ell_i$ is drawn from the ontology-defined label
space. Depending on the task, $\ell_i$ may denote an entity type, a relation
type, or a hierarchical label tuple. For hierarchical ontologies, we write
$\ell_i=(\kappa_i,\varsigma_i)$, where the child label $\varsigma_i$ must be
valid under the ontology mapping $\mathcal{H}$, i.e.,
$\varsigma_i\in\mathcal{H}(\kappa_i)$.

\paragraph{Textual grounding.}
Let $\mathcal{S}(s)$ denote the set of admissible evidence spans in the input
sequence. The grounding component $\Pi_i\subseteq\mathcal{S}(s)$ contains the
span or spans that anchor record $y_i$ to the source text.

\paragraph{Relational linking.}
The relational-linking component $R_i$ stores optional typed links from record
$y_i$ to other records or grounded mentions. These links encode task-specific
inter-record structure, such as relation arguments, directionality,
coreference clusters, or other ontology-defined constraints. If a task does
not require relational links, then $R_i=\emptyset$.

The ontology $\mathcal{O}$ determines which semantic labels, grounding spans,
and relational links are valid. Concrete task-specific instantiations of this
general formulation for PV-Miner and SciERC is presented in 
Appendix~\ref{app:task_specific_formulations}.

\section{Modular Prompt Engineering for Structured Prediction} 
\label{sec:prompt_engineering} 

We next introduce a modular prompt-engineering framework for
ontology-constrained structured prediction. The framework is instantiated using
replaceable task-specific placeholders, allowing the same prompt architecture to
be adapted to any structured prediction task. It targets common failure
modes of in-context structured generation, including format drift, label
ambiguity, reasoning shortcuts, evidence hallucination, metadata-conditioned
confusion, and invalid relational links.

The prompt decomposes structured prediction into six explicitly controlled
modules. \textbf{M1} (XML structuring) partitions global instructions and
task-specific constraints into semantically tagged blocks that establish scope
boundaries for role, performance target, task definition, ontology, and output
requirements \cite{white2023prompt}. \textbf{M2} (disambiguation rules)
encodes expert-derived decision boundaries for confusable ontology labels,
grounding boundaries, and relational-link validity
\cite{pang2023guideline,sainz2024gollie}. \textbf{M3} (reasoning scaffold)
defines a structured verification routine that orders context analysis,
candidate decomposition, semantic labelling, textual grounding, relational-link
validation, and final consistency checking before output emission
\cite{wei2022chain}. \textbf{M4} (metadata-aware decision logic) converts
task metadata into explicit control variables that restrict the valid search
space for labels, evidence spans, and links
\cite{kong2024better,ouyang2022training}. \textbf{M5} (schema contract)
declares machine-parseable validity constraints for records, labels,
groundings, relational links, and task-specific coverage requirements
\cite{li2024simple,sainz2024gollie}. \textbf{M6} (quality gate) defines a
single-turn pre-emission audit for parseability, ontology validity, grounding
fidelity, relational-link consistency, disambiguation-rule compliance, and
expert defensibility \cite{madaan2023selfrefine,huang2024cannot}.

Appendix~\ref{app:Prompt_Engineered_Instruction} provides the detailed
component-level explanation of the modular prompt-engineering framework, and
Figure~\ref{fig:prompt_template} in the appendix presents the corresponding
general-purpose template.

\section{Supervised Structured Initialization}
\label{sec:sft}

After defining the prompt-engineered structured prediction interface, we perform an initial supervised adaptation stage in which the model learns to emit the corresponding gold schema-valid output under this interface. All subsequent preference-optimization stages build on the resulting supervised model $p_{\mathrm{SFT}}$, which is kept fixed as the reference policy. 

Given a gold record set $Y=\{y_1,\ldots,y_N\}$, we serialize it into a
JSON completion string $Y_s$ and concatenate it with the instantiated prompt
$x$, where $x$ contains the prompt-engineered instruction populated with the
input sequence $s$, and optional metadata $m$. Let $z_{1:T}=\operatorname{Tok}(x \,\Vert\, Y_s)$ denote the
tokenized causal-LM training sequence, where $\operatorname{Tok}(\cdot)$ is the
model tokenizer and $\Vert$ denotes string concatenation. Following
completion-only instruction tuning, we compute loss only over tokens belonging
to the serialized completion $Y_s$:

\begin{equation}
\begin{aligned}
\mathcal{L}_{\mathrm{SFT}}(\theta)
=
-\mathbb{E}_{(x,Y_s)\sim\mathcal{D}_{\mathrm{SFT}}}
\Bigg[
\frac{
\sum_{t=1}^{T} r_t
\log p_{\theta}(z_t \mid z_{<t})
}{
\sum_{t=1}^{T} r_t
}
\Bigg].
\end{aligned}
\end{equation}
where $r_t \in \{0,1\}$ indicates whether token $z_t$ belongs to the
completion $Y_s$. The resulting model $p_{\mathrm{SFT}}$ serves as the
initialization for preference optimization and is kept fixed as the reference policy in TAB-PO.

\section{Confusion-Aware Preference Construction}
\label{sec:confusion_pref}

Having obtained the supervised model $p_{\mathrm{SFT}}$, we next construct
preference pairs that expose the model to realistic structured prediction
errors. Preference optimization is most effective when the rejected completion
represents a plausible failure mode rather than an arbitrary corrupted output.
In ontology-driven structured generation, residual SFT errors are often
\emph{structured confusions}: the model produces a schema-valid output that is
mostly correct but substitutes one plausible ontology-bound decision for
another, such as a confusable label, an imprecise evidence span, or an
incorrect relational link. These errors are especially challenging because the
preferred and rejected outputs may share nearly all serialization tokens and
differ only at a small number of task-critical decision points. We therefore
construct preference triples $(x,Y^{+},Y^{-})$, where $Y^{+}$ is the gold
structured output and $Y^{-}$ is a minimally perturbed, schema-valid alternative
designed to isolate a realistic ontology-level error.

Our construction follows a hybrid protocol. First, we include
expert-curated preferences that capture ambiguity patterns identified
during annotation review and adjudication (40\%). Second, we generate
confusion-aware synthetic preferences from empirical errors made by the
SFT model on a held-out validation split (60\%). This design ensures that preference
optimization focuses on realistic ontology-level confusions rather than
uninformative random negatives.

\paragraph{Empirical confusion extraction on validation data.}
Using the supervised model $p_{\mathrm{SFT}}$, we perform inference on a
held-out validation set and compare each predicted structured output
$\widehat{Y}$ with the gold structured output $Y$. Because semantic-label
confusions are meaningful only when the corresponding records refer to the same
textual evidence, we first align gold and predicted records using relaxed
span-level matching. For a predicted span
$\hat{\pi}\in\widehat{\Pi}_k$ and a gold span $\pi\in\Pi_j$, let
$\operatorname{Tok}(\cdot)$ denote the set of tokens in a span. We use a relaxed span-matching criterion that treats two spans as matched
when either span token set contains the other, or when their partial token
overlap exceeds a threshold. For the latter case, we compute token-level
Jaccard similarity as
\begin{equation}
\operatorname{Jaccard}(\hat{\pi},\pi)
=
\frac{
|\operatorname{Tok}(\hat{\pi})\cap \operatorname{Tok}(\pi)|
}{
|\operatorname{Tok}(\hat{\pi})\cup \operatorname{Tok}(\pi)|
}.
\end{equation}
Given a relaxed matching threshold $\gamma$, we define

\begin{equation}
\begin{gathered}
\operatorname{Match}_{\Pi,\gamma}(\hat{\pi},\pi)
=
\mathbb{I}\big[
\mathrm{Tok}(\pi)\subseteq \mathrm{Tok}(\hat{\pi})
\;\vee\;
\mathrm{Tok}(\hat{\pi})\subseteq \mathrm{Tok}(\pi)
\;\vee\;
\mathrm{Jaccard}(\hat{\pi},\pi)\ge \gamma
\big].
\end{gathered}
\end{equation}
Using this span-matching criterion, we form the set of matched
gold--predicted record pairs:

\begin{equation}
\begin{gathered}
\mathcal{M}_{\gamma}(Y,\widehat{Y}) =
\{(j,k):\exists\,\pi\in\Pi_j,
\exists\,\hat{\pi}\in\widehat{\Pi}_k\ \mathrm{s.t.}\ 
\operatorname{Match}_{\Pi,\gamma}(\hat{\pi},\pi)=1\}.
\end{gathered}
\end{equation}
In our implementation, we set $\gamma=0.6$. We then collect mismatched semantic
label pairs only over these span-aligned records:

\begin{equation}
\begin{aligned}
\mathcal{A}^{\neq}(Y,\widehat{Y})
=
\{(\ell_j,\widehat{\ell}_k):\;
(j,k)\in \mathcal{M}_{0.6}(Y,\widehat{Y}),
\ell_j\neq\widehat{\ell}_k
\}.
\end{aligned}
\end{equation}
Aggregating these mismatches over the validation set yields a
frequency-weighted empirical confusion distribution over ontology labels.

\paragraph{Preference triples construction.}
The empirical confusion distribution provides a task-specific estimate of the
model's most likely ontology-level failure modes. We therefore use it to guide
negative construction, ensuring that rejected outputs remain schema-valid while
targeting realistic low-separation errors. Each preference triple shares the same input $x$ and differs only in the
structured output. We set $Y^{+}=Y$ and construct a minimally perturbed
$Y^{-}$ that preserves schema validity while introducing a realistic structured
error. We use four different perturbation families:
\begin{enumerate}[leftmargin=1.2em,itemsep=2pt,topsep=2pt]
\item \textbf{Semantic confusion with preserved grounding.}
This perturbation changes one ontology label while preserving the grounding and
relational structure of the selected record.

\noindent\textbf{Step 1: estimate label vulnerability.}
Let $C(\ell,\tilde{\ell})$ denote the number of times the SFT model predicts
$\tilde{\ell}$ when the gold semantic label is $\ell$ on the validation set.
We define the total confusion count for label $\ell$ as
\begin{equation}
c(\ell)=\sum_{\ell'\neq \ell} C(\ell,\ell').
\end{equation}

\noindent\textbf{Step 2: select a vulnerable record.}
For a structured output $Y^{+}=\{y^{+}_1,\ldots,y^{+}_N\}$ with
$y^{+}_i=(\ell_i,\Pi_i,R_i)$, we sample a record index from the
vulnerability-weighted distribution
\begin{equation}
\label{eq:index_sample}
q(i\mid Y^{+})
=
\frac{c(\ell_i)+\epsilon}
{\sum_{j=1}^{N}\left(c(\ell_j)+\epsilon\right)},
\end{equation}
where $\epsilon>0$ provides smoothing. This assigns higher probability to
records whose labels are more frequently confused by $p_{\mathrm{SFT}}$.

\noindent\textbf{Step 3: sample a confusable replacement label.}
Given the selected record $y_i^{+}=(\ell_i,\Pi_i,R_i)$, we sample an
ontology-valid replacement label $\tilde{\ell}_i$ from the empirical confusion
distribution
\begin{equation}
\begin{aligned}
p_{\mathrm{conf}}(\tilde{\ell}\mid \ell_i)
=
\frac{C(\ell_i,\tilde{\ell})+\epsilon}
{\sum_{\ell'}
\left(C(\ell_i,\ell')+\epsilon\right)} .
\end{aligned}
\end{equation}

\noindent\textbf{Step 4: construct the rejected output.}
We preserve the grounding and relational components and define
$\tilde{y}_i=(\tilde{\ell}_i,\Pi_i,R_i)$. The rejected output is
\begin{equation}
Y^{-}=Y^{+}\setminus\{y_i^{+}\}\cup\{\tilde{y}_i\}.
\end{equation}
Thus, the preferred and rejected outputs differ in a single semantic decision
while retaining the same evidence spans and relational links.

\item \textbf{Missing-record perturbation.}
We delete a record from the preferred output to simulate under-extraction:
\begin{equation}
Y^{-}=Y^{+}\setminus\{y_i^{+}\},
\end{equation}
where $y_i^{+}$ is selected according to Eq.~\ref{eq:index_sample}.

\item \textbf{Extra-record perturbation.}
We insert a spurious but schema-valid record to simulate over-extraction:
\begin{equation}
Y^{-}=Y^{+}\cup\{\tilde{y}\},
\end{equation}
where $\tilde{y}=(\tilde{\ell},\tilde{\Pi},\tilde{R})$ is constructed using an
ontology-valid semantic label $\tilde{\ell}$, a candidate grounding set
$\tilde{\Pi}$ sampled from the source sequence $s$, and optional relational
links $\tilde{R}$, subject to the task-specific ontology, relational-linking
rules, and schema-validity constraints.

\item \textbf{Relational-link perturbation.}

We modify a relational link to simulate relational-linking confusion:
\begin{equation}
Y^{-}=Y^{+}\setminus\{y_i^{+}\}\cup\{\tilde{y}_i\},
\end{equation}
where $y_i^{+}=(\ell_i,\Pi_i,R_i)$ is selected according to
Eq.~\ref{eq:index_sample}, and
$\tilde{y}_i=(\ell_i,\Pi_i,\tilde{R}_i)$ preserves the semantic label and
grounding spans while replacing $R_i$ with a schema-valid but incorrect
relational-link set $\tilde{R}_i$.
\end{enumerate}

For intuitive examples of how these perturbation families produce low-separation, schema-valid hard negatives, see the PV-Miner and SciERC illustrations in Appendix~\ref{append:nega_con}.

\section{Token-Adaptive Barrier Preference Optimization}
\label{sec:tabpo}

\paragraph{Motivation.}
The preference construction described above yields low-separation structured
preference pairs, where preferred and rejected completions often differ in only
a small number of schema-defining decisions while sharing most of the same
serialized output. Standard sequence-level DPO is poorly matched to this
regime. First, its gradients can be diluted across JSON
scaffolding and non-critical serialization tokens rather than concentrated on
the sparse tokens that determine structured correctness. Second, because DPO
optimizes only a relative sequence-level margin, it can improve the aggregate
preference margin while reducing the likelihood of some correct preferred
tokens. We refer to these effects as \emph{gradient dilution} and
\emph{preferred-token erosion}. Appendix~\ref{app:gradient_dilution} provides the mathematical intuition behind
these two failure modes and derives the corresponding token-level decomposition
of the DPO gradient.

\paragraph{Objective.}
Let \(Y_s^{+}\) and \(Y_s^{-}\) denote the serialized preferred and rejected
completion strings obtained from \(Y^{+}\) and \(Y^{-}\). We define the
serialized preference dataset as \(\mathcal{D}_{\mathrm{pref}}\), whose elements
are \((x,Y_s^{+},Y_s^{-})\). For each serialized completion \(Y_s\), let
\(u=\operatorname{Tok}(Y_s)=(u_1,\ldots,u_T)\) denote its completion-token
sequence. We write
the completion log-likelihood as
\begin{equation}
\mu_{\theta}(Y_s\mid x)
\triangleq
\sum_{t=1}^{T}
\log p_{\theta}(u_t\mid x,u_{<t}).
\label{eq:completion_loglik}
\end{equation}

\begin{algorithm}[!t]
\caption{Token-Adaptive Barrier Preference Optimization}
\label{alg:tabpo}
\begin{algorithmic}[1]
\Require SFT data \(\mathcal{D}_{\mathrm{SFT}}\), validation data
\(\mathcal{D}_{\mathrm{val}}\), expert preferences
\(\mathcal{D}_{\mathrm{expert}}\), ontology \(\mathcal{O}\), relaxed span-matching threshold \(\gamma\), smoothing constant \(\epsilon\), preference temperature \(\beta\), confidence threshold \(\tau\), barrier weight \(\lambda\)
\Ensure TAB-PO policy \(p_{\theta}\)

\State Train \(p_{\mathrm{SFT}}\) on \(\mathcal{D}_{\mathrm{SFT}}\) using
completion-only supervised loss \(\mathcal{L}_{\mathrm{SFT}}\).
\State Initialize \(p_{\theta}\leftarrow p_{\mathrm{SFT}}\) and keep
\(p_{\mathrm{SFT}}\) fixed as the reference model.

\Statex \textbf{Confusion-aware preference construction}
\State Run \(p_{\mathrm{SFT}}\) on \(\mathcal{D}_{\mathrm{val}}\), align records
with \(\operatorname{Match}_{\Pi,\gamma}\), and estimate confusion statistics
\(C(\ell,\tilde{\ell})\), \(c(\ell)\), and
\(p_{\mathrm{conf}}(\tilde{\ell}\mid \ell)\).
\State Using \(C(\ell,\tilde{\ell})\), \(c(\ell)\), and
\(p_{\mathrm{conf}}(\tilde{\ell}\mid \ell)\), construct
\(\mathcal{D}_{\mathrm{pref}}\) by combining
\(\mathcal{D}_{\mathrm{expert}}\) with schema-valid hard negatives generated
from semantic-confusion, missing-record, extra-record, and relational-link
perturbations.

\Statex \textbf{TAB-PO preference optimization}
\For{each minibatch \(\mathcal{B}\subset\mathcal{D}_{\mathrm{pref}}\)}
    \State For each preferred completion \(Y_s^{+}\), compute the detached
    confidence gate
    \[
    g_t^{\theta}(x,u^{+})
    =
    \mathbb{I}
    \left[
    \log p_{\theta}(u_t^{+}\mid x,u_{<t}^{+})<\log\tau
    \right].
    \]
    \State Update \(p_{\theta}\) by minimizing the minibatch TAB-PO objective
    \[
    \begin{aligned}
    \widehat{\mathcal{L}}_{\mathrm{TAB\text{-}PO}}(\theta;\mathcal{B})
    =
    \frac{1}{|\mathcal{B}|}
    \sum_{(x,Y_s^{+},Y_s^{-})\in\mathcal{B}}
    \Bigg[
    &
    -\log\sigma
    \Bigg(
    \beta
    \Big(
    \mu_{\theta}(Y_s^{+}\mid x)
    -
    \mu_{\theta}(Y_s^{-}\mid x)
    \\
    &
    -
    \mu_{\mathrm{SFT}}(Y_s^{+}\mid x)
    +
    \mu_{\mathrm{SFT}}(Y_s^{-}\mid x)
    \Big)
    \Bigg)
    \\
    &+
    \lambda
    \frac{
    \sum_{t=1}^{T^{+}}
    g_t^{\theta}(x,u^{+})
    \left[-\log p_{\theta}(u_t^{+}\mid x,u_{<t}^{+})\right]
    }{
    \max\left(
    1,\sum_{t=1}^{T^{+}}g_t^{\theta}(x,u^{+})
    \right)
    }
    \Bigg].
    \end{aligned}
    \]
\EndFor
\State \Return \(p_{\theta}\)
\end{algorithmic}
\end{algorithm}

Here, \(\mu_{\mathrm{SFT}}\) denotes the same completion log-likelihood
computed under the fixed supervised reference model \(p_{\mathrm{SFT}}\).
For each preference triple $(x,Y_s^{+},Y_s^{-})$, we define the
reference-adjusted advantage as the difference between the policy
log-likelihood margin and the corresponding SFT-reference log-likelihood margin:
\begin{equation}
\begin{aligned}
\Delta^{}_{\theta}
\triangleq\;
\left[
\mu_{\theta}(Y_s^{+}\mid x)
-
\mu_{\theta}(Y_s^{-}\mid x)
\right]
-
\left[
\mu_{\mathrm{SFT}}(Y_s^{+}\mid x)
-
\mu_{\mathrm{SFT}}(Y_s^{-}\mid x)
\right],
\end{aligned}
\label{eq:tabpo_delta}
\end{equation}

 The corresponding
preference loss is
\begin{equation}
\begin{aligned}
\mathcal{L}_{\mathrm{pref}}(\theta)
=
-
\mathbb{E}_{\mathcal{D}_{\mathrm{pref}}}
\left[
\log \sigma
\left(
\beta
\Delta^{}_{\theta}
\right)
\right].
\end{aligned}
\label{eq:tabpo_pref_loss}
\end{equation}

To address gradient dilution and preferred-token erosion, TAB-PO adds a
confidence-gated token barrier on the preferred completion. The barrier
activates only when the current policy is under-confident on preferred
completion tokens, according to a probability threshold \(\tau\in(0,1)\).

For the preferred serialized completion \(Y_s^{+}\), let
\(u^{+}=\operatorname{Tok}(Y_s^{+})
=(u_1^{+},\ldots,u_{T^{+}}^{+})\). We define the per-token gate as
\begin{equation}
\begin{aligned}
g_t^{\theta}(x,u^{+})
\triangleq
\mathbb{I}
\left[
\log p_{\theta}(u_t^{+}\mid x,u_{<t}^{+})
<
\log \tau
\right].
\end{aligned}
\label{eq:tabpo_gate}
\end{equation}

The gate identifies preferred-token positions whose likelihood falls below the
confidence threshold. TAB-PO then applies an SFT-style likelihood-restoration
penalty only to these activated positions, averaging the preferred-token
negative log-likelihood over the selected tokens:
\begin{equation}
\begin{aligned}
\mathcal{L}_{\mathrm{barrier}}(\theta)
=
\mathbb{E}_{(x,Y_s^{+},Y_s^{-})\sim\mathcal{D}_{\mathrm{pref}}}
\left[
\frac{
\sum_{t=1}^{T^{+}}
g_t^{\theta}(x,u^{+})
\left[-\log p_{\theta}(u_t^{+}\mid x,u_{<t}^{+})\right]
}{
\max\left(1,\sum_{t=1}^{T^{+}}g_t^{\theta}(x,u^{+})\right)
}
\right].
\end{aligned}
\label{eq:tabpo_barrier}
\end{equation}

When no preferred token is selected by the gate, the numerator is zero and the
barrier contributes no loss. In implementation, the binary gate
\(g_t^{\theta}\) is detached from the computation graph; gradients are
propagated only through the selected preferred-token negative log-likelihood
terms.

The final TAB-PO objective combines reference-adjusted preference learning with
the confidence-gated supervised restoration term:
\begin{equation}
\mathcal{L}_{\mathrm{TAB\text{-}PO}}(\theta)
=
\mathcal{L}_{\mathrm{pref}}(\theta)
+
\lambda
\mathcal{L}_{\mathrm{barrier}}(\theta),
\label{eq:tabpo_final}
\end{equation}
where \(\lambda\geq 0\) controls the strength of the barrier. Algorithm~\ref{alg:tabpo} provides a compact summary of the TAB-PO procedure, showing how validation-set confusion statistics and expert preferences are used to construct schema-valid hard negatives, which are then optimized with the confidence-gated token-adaptive barrier objective.

\subsection{Mechanistic Diagnostics}
\label{sec:mechanistic_diagnostics}

\begin{figure}[!h]
\centering

\includegraphics[width=\linewidth,height=0.11\textheight,keepaspectratio]{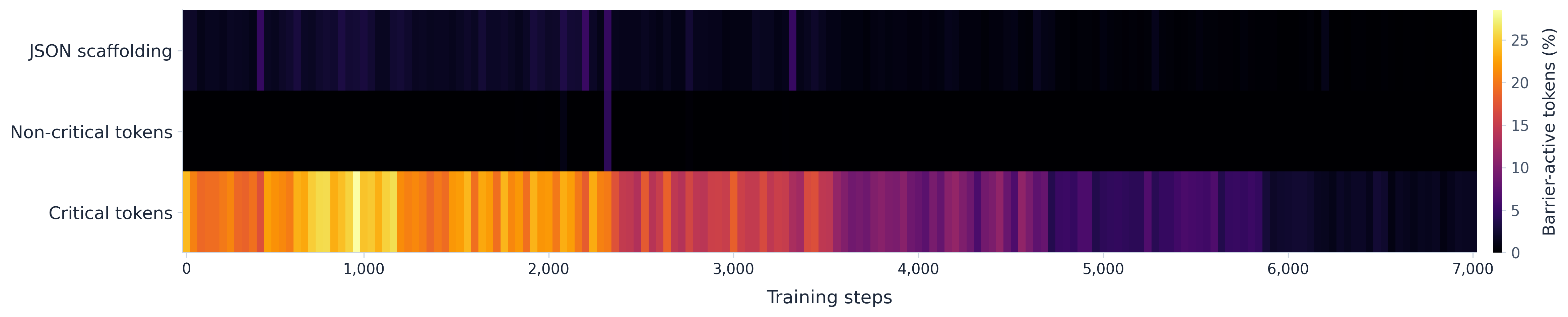}

\includegraphics[width=\linewidth,height=0.22\textheight,keepaspectratio]{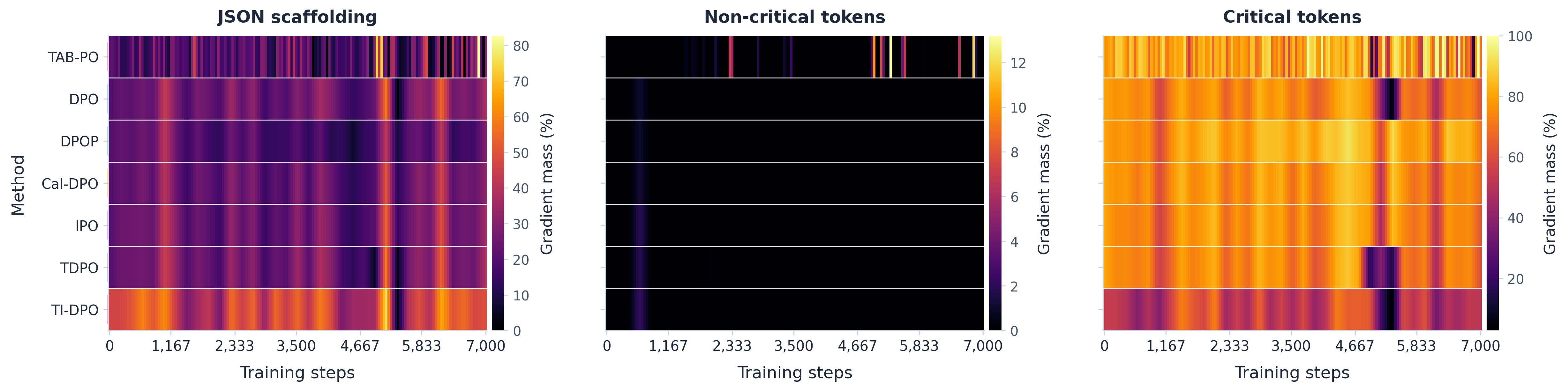}

\includegraphics[width=\linewidth,height=0.23\textheight,keepaspectratio]{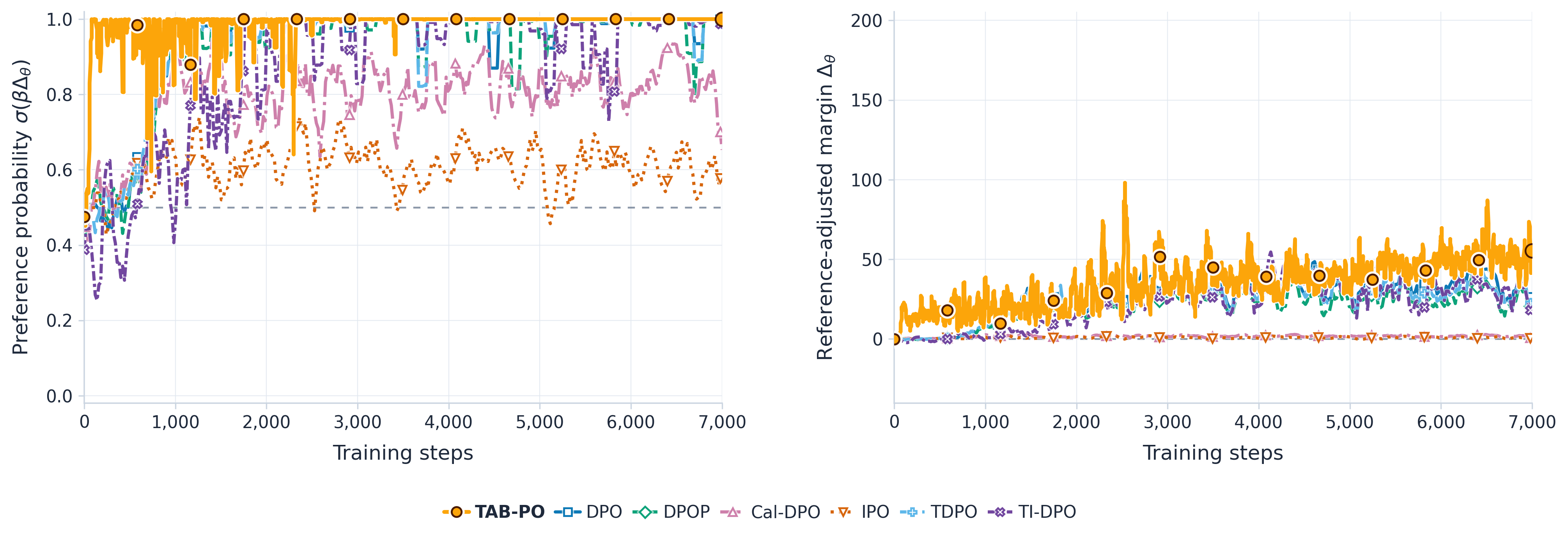}

\caption{
Mechanistic diagnostics for TAB-PO and preference-optimization variants.
The top panel shows TAB-PO barrier activity across JSON scaffolding,
non-critical serialization tokens, and critical schema tokens.
The middle panel shows gradient-mass allocation across the same token
categories for TAB-PO and DPO-family baselines.
The bottom panel shows preference dynamics, with preference probability
$\sigma(\beta\Delta_\theta)$ and reference-adjusted margin $\Delta_\theta$.
Together, these diagnostics show that TAB-PO concentrates optimization on
schema-critical tokens while maintaining preferred-over-rejected separation.
}
\label{fig:tabpo_training_diagnostics}
\end{figure}

TAB-PO is designed to correct mechanism-level failures that arise when
preference optimization is applied to low-separation structured outputs:
gradient dilution and preferred-token erosion. We therefore audit the training dynamics directly,
rather than relying only on downstream F1. The diagnostics separate completion
tokens into three categories: \emph{critical schema tokens}, which determine
semantic labels, textual groundings, relation labels, relation arguments,
sentence identifiers, or coreference links; \emph{JSON scaffolding tokens},
which encode braces, brackets, commas, colons, and quotation marks; and
\emph{non-critical serialization tokens}, such as repeated field names or other
formatting tokens that support parseability but are not themselves scored
structured decisions.

Figure~\ref{fig:tabpo_training_diagnostics} reports three complementary
diagnostics. Each diagnostic isolates a different part of the proposed
mechanism: where the barrier activates, where gradient mass is allocated, and
whether preference separation is preserved while vulnerable preferred tokens are
restored.

\begin{enumerate}[leftmargin=1.35em,itemsep=3pt,topsep=3pt]

\item \textbf{Barrier activity localizes likelihood restoration.}
The barrier-activity panel measures the fraction of preferred tokens whose
current policy probability falls below the confidence threshold $\tau$ and
therefore activates the TAB-PO restoration term. TAB-PO activates most strongly
on critical schema tokens, especially early in optimization, while remaining
weak on non-critical serialization tokens. This shows that the confidence gate
is not simply adding uniform SFT regularization; instead, it selectively
restores the token positions where erosion would most affect structured
correctness. JSON-scaffolding tokens receive weaker but nonzero activation,
which is desirable when structural tokens must be protected to preserve
machine-parseable outputs.

\item \textbf{Gradient mass shifts toward schema-critical decisions.}
The gradient-mass panels examine where preference-optimization updates are
allocated. Sequence-level and token-level DPO-family baselines allocate visible
update mass to JSON scaffolding and other non-critical serialization tokens. In
contrast, TAB-PO shifts a larger share of the learning signal toward critical
schema tokens---the sparse positions that determine ontology labels, grounded
spans, relation labels, and linking decisions. This directly supports the
intended role of TAB-PO in reducing gradient dilution.

\item \textbf{Preference separation is preserved without token erosion.}
The preference-dynamics panels track both the preference probability
$\sigma(\beta\Delta_\theta)$ and the reference-adjusted margin
$\Delta_\theta$. TAB-PO maintains strong preferred-over-rejected separation
while the barrier remains active on under-confident preferred tokens. Thus, the
restoration term does not suppress preference learning; instead, it protects
vulnerable preferred tokens while the model continues to increase the
reference-adjusted preference margin.

\end{enumerate}

Together, these diagnostics show that TAB-PO improves the allocation of
optimization signal, preserves token-level stability, and maintains
sequence-level preference separation.

\section{Experimental Results}
\label{sec:experiments_results}

\paragraph{Scoring overview.}
Figure~\ref{fig:tabpo-pipeline-1} illustrates the two complementary structured prediction tasks evaluated in this paper, and Appendix~\ref{append:metric}
provides the corresponding task-specific scoring definitions. PV-Miner
evaluates hierarchical clinical communication mining through coarse Code F1
(parent-level semantic labelling), fine-grained Sub-code F1 (child-level
semantic labelling), and grounded Span F1 (textual grounding). SciERC evaluates
scientific information extraction through Entity F1 (entity semantic
labelling), Relation F1 (typed relational linking), Span F1 (entity textual
grounding), and Coreference F1 (cross-mention relational linking). Thus, the
reported metrics separate semantic labelling, textual grounding, and relational
linking quality.

We first verify that the modular prompt-engineered interface provides a stronger
zero-shot initialization before supervised or preference-based fine-tuning.
Appendix~\ref{app:Prompt_Engineered_Instruction} reports the full zero-shot
comparison between baseline and prompt-engineered instructions in
Table~\ref{tab:zeroshot_prompt_baseline_side_by_side}. Prompt engineering
improves nearly all model--metric cells on PV-Miner and all cells on SciERC,
supporting its role as an effective structured prediction interface before
SFT and TAB-PO.

\subsection{TAB-PO Consistently Improves over SFT}

Table~\ref{tab:dpo_variants_side_by_side} reports the main post-SFT preference
optimization results. On PV-Miner, TAB-PO improves over the corresponding SFT
baseline by an average of +3.84 F1 across 18 model--metric cells, with the
largest gains on the semantic hierarchy: +4.49 Code F1 and +6.25 Sub-code F1 on
average.

On SciERC, the improvements are larger. TAB-PO improves over SFT by an average
of +5.77 F1 across 24 model--metric cells, with gains concentrated on the most
structurally demanding dimensions: +7.93 Relation F1 and +10.13 Coreference F1
on average. Entity F1 and Span F1 also improve by +1.22 and +3.80,
respectively.

\definecolor{posgreen}{HTML}{2E7D32}
\definecolor{negred}{HTML}{C62828}

\newcommand{\upcell}[3]{\cellcolor{posgreen!#3}#1 ($\uparrow$ #2)}
\newcommand{\downcell}[3]{\cellcolor{negred!#3}#1 ($\downarrow$ #2)}
\newcommand{\bupcell}[3]{\cellcolor{posgreen!#3}#1 ($\uparrow$ #2)}
\newcommand{\bdowncell}[3]{\cellcolor{negred!#3}#1 ($\downarrow$ #2)}

\begin{snwidetable}[!t]
\centering
\caption{F1 results of TAB-PO compared with recent sequence-level and token-level DPO-family methods on PV Miner and SciERC.
Parentheses report the change from the corresponding SFT baseline for the same model and task. 
Green cells indicate improvements and red cells indicate degradations; darker shading indicates a larger absolute change. All the results are the mean over 3 runs with different random seeds.}
\label{tab:dpo_variants_side_by_side}
\scriptsize
\setlength{\tabcolsep}{2.0pt}
\renewcommand{\arraystretch}{1.05}

\vspace{1mm}

\resizebox{\linewidth}{!}{%
\begin{tabular}[t]{@{}llccc@{}}
\toprule
\multicolumn{5}{c}{\textbf{(a) PV Miner}} \\
\midrule
Model & Method & Code & Sub-code & Span \\
\midrule
\multirow{7}{*}{\shortstack[l]{Llama-3.1\\8B}}
 & Cal-DPO & \downcell{77.43}{1.67}{26} & \upcell{74.51}{0.05}{8} & \downcell{86.83}{0.11}{8} \\
 & DPO & \upcell{79.35}{0.25}{12} & \upcell{75.81}{1.35}{26} & \upcell{87.24}{0.30}{12} \\
 & DPOP & \upcell{79.43}{0.33}{12} & \upcell{75.59}{1.13}{26} & \upcell{87.28}{0.34}{12} \\
 & IPO & \downcell{12.75}{66.35}{80} & \downcell{6.25}{68.21}{80} & \downcell{6.00}{80.94}{80} \\
 & TDPO & \downcell{77.89}{1.21}{26} & \downcell{73.77}{0.69}{18} & \downcell{85.38}{1.56}{26} \\
 & TI-DPO & \upcell{79.80}{0.70}{18} & \upcell{75.75}{1.29}{26} & \upcell{87.18}{0.24}{8} \\
 & TAB-PO & \bupcell{82.99}{3.89}{38} & \bupcell{80.62}{6.16}{52} & \bupcell{87.67}{0.73}{18} \\
\midrule

\multirow{7}{*}{\shortstack[l]{Llama-3.2\\3B}}
 & Cal-DPO & \downcell{65.51}{9.61}{66} & \downcell{65.90}{3.74}{38} & \downcell{81.46}{2.29}{38} \\
 & DPO & \downcell{74.80}{0.32}{12} & \upcell{70.68}{1.04}{26} & \upcell{84.48}{0.73}{18} \\
 & DPOP & \downcell{74.89}{0.23}{8} & \upcell{71.08}{1.44}{26} & \upcell{84.25}{0.50}{18} \\
 & IPO & \downcell{52.78}{22.34}{80} & \downcell{42.69}{26.95}{80} & \downcell{49.80}{33.95}{80} \\
 & TDPO & \downcell{72.61}{2.51}{38} & \upcell{70.14}{0.50}{18} & \downcell{82.72}{1.03}{26} \\
 & TI-DPO & \downcell{74.88}{0.24}{8} & \upcell{70.52}{0.88}{18} & \upcell{84.25}{0.50}{18} \\
 & TAB-PO & \upcell{80.49}{5.37}{52} & \upcell{78.08}{8.44}{66} & \upcell{85.29}{1.54}{26} \\
\midrule

\multirow{7}{*}{\shortstack[l]{Llama-3.3\\70B}}
 & Cal-DPO & \downcell{79.77}{3.92}{38} & \downcell{75.49}{4.64}{52} & \downcell{86.21}{2.38}{38} \\
 & DPO & \downcell{83.40}{0.29}{12} & \downcell{79.77}{0.36}{12} & \upcell{88.60}{0.01}{8} \\
 & DPOP & \downcell{83.63}{0.06}{8} & \upcell{80.55}{0.42}{12} & \downcell{88.40}{0.19}{8} \\
 & IPO & \downcell{8.52}{75.17}{80} & \downcell{3.21}{76.92}{80} & \downcell{11.02}{77.57}{80} \\
 & TDPO & \downcell{83.33}{0.36}{12} & \downcell{79.70}{0.43}{12} & \downcell{88.44}{0.15}{8} \\
 & TI-DPO & \downcell{83.67}{0.02}{8} & \upcell{80.38}{0.25}{12} & \bupcell{88.69}{0.10}{8} \\
 & TAB-PO & \bupcell{84.97}{1.28}{26} & \bupcell{82.71}{2.58}{38} & \downcell{88.03}{0.56}{18} \\
\midrule

\multirow{7}{*}{\shortstack[l]{Qwen2.5\\1.5B}}
 & Cal-DPO & \downcell{65.84}{7.56}{52} & \downcell{64.74}{2.17}{38} & \downcell{81.97}{2.32}{38} \\
 & DPO & \downcell{68.30}{5.10}{52} & \downcell{65.36}{1.55}{26} & \upcell{84.61}{0.32}{12} \\
 & DPOP & \downcell{69.78}{3.62}{38} & \downcell{66.66}{0.25}{12} & \downcell{84.17}{0.12}{8} \\
 & IPO & \downcell{61.43}{11.97}{66} & \downcell{51.47}{15.44}{66} & \downcell{75.77}{8.52}{66} \\
 & TDPO & \downcell{67.74}{5.66}{52} & \downcell{64.49}{2.42}{38} & \downcell{83.12}{1.17}{26} \\
 & TI-DPO & \downcell{72.47}{0.93}{18} & \downcell{66.70}{0.21}{8} & \upcell{84.63}{0.34}{12} \\
 & TAB-PO & \bupcell{78.76}{5.36}{52} & \bupcell{74.01}{7.10}{52} & \bupcell{85.35}{1.06}{26} \\
\midrule

\multirow{7}{*}{\shortstack[l]{Qwen2.5\\7B}}
 & Cal-DPO & \downcell{74.67}{2.72}{38} & \downcell{71.82}{1.67}{26} & \downcell{85.64}{1.31}{26} \\
 & DPO & \downcell{75.69}{1.70}{26} & \downcell{71.62}{1.87}{26} & \downcell{85.95}{1.00}{26} \\
 & DPOP & \downcell{75.84}{1.55}{26} & \downcell{71.99}{1.50}{26} & \downcell{86.10}{0.85}{18} \\
 & IPO & \downcell{43.44}{33.95}{80} & \downcell{32.21}{41.28}{80} & \downcell{30.92}{56.03}{80} \\
 & TDPO & \downcell{76.01}{1.38}{26} & \downcell{71.76}{1.73}{26} & \downcell{86.10}{0.85}{18} \\
 & TI-DPO & \downcell{76.84}{0.55}{18} & \downcell{71.91}{1.58}{26} & \downcell{86.18}{0.77}{18} \\
 & TAB-PO & \bupcell{81.86}{4.47}{52} & \bupcell{78.31}{4.82}{52} & \bupcell{87.32}{0.37}{12} \\
\midrule

\multirow{7}{*}{\shortstack[l]{Qwen2.5\\14B}}
 & Cal-DPO & \upcell{77.82}{2.13}{38} & \upcell{75.11}{4.00}{52} & \upcell{86.91}{0.95}{18} \\
 & DPO & \upcell{79.06}{3.37}{38} & \upcell{74.88}{3.77}{38} & \upcell{87.07}{1.11}{26} \\
 & DPOP & \upcell{79.11}{3.42}{38} & \upcell{74.91}{3.80}{38} & \upcell{87.05}{1.09}{26} \\
 & IPO & \downcell{31.86}{43.83}{80} & \downcell{33.77}{37.34}{80} & \downcell{19.70}{66.26}{80} \\
 & TDPO & \upcell{79.12}{3.43}{38} & \upcell{74.99}{3.88}{38} & \upcell{86.92}{0.96}{18} \\
 & TI-DPO & \upcell{79.11}{3.42}{38} & \upcell{74.99}{3.88}{38} & \upcell{87.11}{1.15}{26} \\
 & TAB-PO & \bupcell{82.26}{6.57}{52} & \bupcell{79.50}{8.39}{66} & \bupcell{87.56}{1.60}{26} \\
\bottomrule
\end{tabular}%
\hspace{6pt}%
\begin{tabular}[t]{@{}llcccc@{}}
\toprule
\multicolumn{6}{c}{\textbf{(b) SciERC}} \\
\midrule
Model & Method & Entity & Relation & Span & Coreference \\
\midrule
\multirow{7}{*}{\shortstack[l]{Llama-3.1\\8B}}
 & Cal-DPO & \upcell{90.73}{0.09}{8} & \upcell{50.70}{0.15}{8} & \downcell{85.38}{0.02}{8} & \upcell{41.23}{0.99}{18} \\
 & DPO & \upcell{90.93}{0.29}{12} & \upcell{51.06}{0.51}{18} & \upcell{86.01}{0.61}{18} & \upcell{40.64}{0.40}{12} \\
 & DPOP & \upcell{91.24}{0.60}{18} & \upcell{51.04}{0.49}{12} & \upcell{85.91}{0.51}{18} & \upcell{41.06}{0.82}{18} \\
 & IPO & \downcell{90.50}{0.14}{8} & \downcell{50.50}{0.05}{8} & \upcell{85.89}{0.49}{12} & \upcell{40.91}{0.67}{18} \\
 & TDPO & \upcell{90.83}{0.19}{8} & \upcell{51.12}{0.57}{18} & \upcell{85.94}{0.54}{18} & \downcell{39.26}{0.98}{18} \\
 & TI-DPO & \downcell{90.32}{0.32}{12} & \downcell{49.54}{1.01}{26} & \upcell{85.85}{0.45}{12} & \upcell{41.56}{1.32}{26} \\
 & TAB-PO & \bupcell{91.50}{0.86}{18} & \bupcell{56.63}{6.08}{52} & \bupcell{86.31}{0.91}{18} & \bupcell{46.24}{6.00}{52} \\
\midrule

\multirow{7}{*}{\shortstack[l]{Llama-3.2\\3B}}
 & Cal-DPO & \upcell{90.49}{0.65}{18} & \upcell{39.19}{0.54}{18} & \downcell{82.14}{1.11}{26} & \upcell{27.69}{0.55}{18} \\
 & DPO & \upcell{90.12}{0.28}{12} & \upcell{39.56}{0.91}{18} & \upcell{83.36}{0.11}{8} & \upcell{30.98}{3.84}{38} \\
 & DPOP & \upcell{90.36}{0.52}{18} & \upcell{39.81}{1.16}{26} & \upcell{83.42}{0.17}{8} & \upcell{28.45}{1.31}{26} \\
 & IPO & \upcell{90.45}{0.61}{18} & \upcell{39.23}{0.58}{18} & \downcell{82.94}{0.31}{12} & \upcell{30.23}{3.09}{38} \\
 & TDPO & \downcell{89.81}{0.03}{8} & \upcell{40.88}{2.23}{38} & \downcell{83.13}{0.12}{8} & \upcell{29.58}{2.44}{38} \\
 & TI-DPO & \downcell{89.59}{0.25}{12} & \downcell{35.39}{3.26}{38} & \downcell{83.21}{0.04}{8} & \downcell{26.64}{0.50}{18} \\
 & TAB-PO & \bupcell{91.04}{1.20}{26} & \bupcell{46.67}{8.02}{66} & \bupcell{84.98}{1.73}{26} & \bupcell{35.11}{7.97}{52} \\
\midrule

\multirow{7}{*}{\shortstack[l]{Llama-3.3\\70B}}
 & Cal-DPO & \downcell{90.83}{0.12}{8} & \upcell{54.55}{1.24}{26} & \downcell{86.24}{0.09}{8} & \upcell{49.72}{0.14}{8} \\
 & DPO & \upcell{91.69}{0.74}{18} & \upcell{54.51}{1.20}{26} & \downcell{86.26}{0.07}{8} & \upcell{50.64}{1.06}{26} \\
 & DPOP & \upcell{91.22}{0.27}{12} & \downcell{52.83}{0.48}{12} & \downcell{85.98}{0.35}{12} & \downcell{48.40}{1.18}{26} \\
 & IPO & \downcell{90.65}{0.30}{12} & \upcell{54.34}{1.03}{26} & \downcell{86.07}{0.26}{12} & \upcell{50.24}{0.66}{18} \\
 & TDPO & \upcell{90.99}{0.04}{8} & \upcell{53.74}{0.43}{12} & \upcell{87.02}{0.69}{18} & \downcell{46.72}{2.86}{38} \\
 & TI-DPO & \downcell{89.00}{1.95}{26} & \upcell{53.82}{0.51}{18} & \upcell{86.80}{0.47}{12} & \downcell{45.13}{4.45}{52} \\
 & TAB-PO & \bupcell{92.21}{1.26}{26} & \bupcell{58.68}{5.37}{52} & \bupcell{88.11}{1.78}{26} & \bupcell{60.80}{11.22}{66} \\
\midrule

\multirow{7}{*}{\shortstack[l]{Qwen2.5\\1.5B}}
 & Cal-DPO & \downcell{89.25}{0.02}{8} & \upcell{24.11}{1.60}{26} & \downcell{67.14}{0.33}{12} & \upcell{11.88}{2.15}{38} \\
 & DPO & \upcell{89.32}{0.05}{8} & \upcell{25.16}{2.65}{38} & \upcell{68.85}{1.38}{26} & \upcell{12.05}{2.32}{38} \\
 & DPOP & \downcell{88.41}{0.86}{18} & \upcell{26.95}{4.44}{52} & \upcell{70.83}{3.36}{38} & \upcell{13.12}{3.39}{38} \\
 & IPO & \upcell{89.46}{0.19}{8} & \upcell{23.96}{1.45}{26} & \upcell{67.59}{0.12}{8} & \upcell{11.09}{1.36}{26} \\
 & TDPO & \downcell{89.03}{0.24}{8} & \upcell{25.38}{2.87}{38} & \upcell{68.56}{1.09}{26} & \upcell{13.28}{3.55}{38} \\
 & TI-DPO & \downcell{89.22}{0.05}{8} & \upcell{23.22}{0.71}{18} & \upcell{73.54}{6.07}{52} & \upcell{10.89}{1.16}{26} \\
 & TAB-PO & \bupcell{90.85}{1.58}{26} & \bupcell{38.31}{15.80}{66} & \bupcell{83.07}{15.60}{66} & \bupcell{24.76}{15.03}{66} \\
\midrule

\multirow{7}{*}{\shortstack[l]{Qwen2.5\\7B}}
 & Cal-DPO & \upcell{90.21}{0.63}{18} & \upcell{41.23}{0.64}{18} & \upcell{82.62}{1.43}{26} & \downcell{29.99}{1.28}{26} \\
 & DPO & \upcell{89.70}{0.12}{8} & \downcell{40.40}{0.19}{8} & \downcell{81.08}{0.11}{8} & \upcell{31.63}{0.36}{12} \\
 & DPOP & \upcell{90.23}{0.65}{18} & \upcell{42.79}{2.20}{38} & \upcell{81.63}{0.44}{12} & \upcell{35.66}{4.39}{52} \\
 & IPO & \upcell{89.99}{0.41}{12} & \upcell{41.21}{0.62}{18} & \downcell{80.97}{0.22}{8} & \upcell{32.25}{0.98}{18} \\
 & TDPO & \upcell{89.60}{0.02}{8} & \upcell{40.63}{0.04}{8} & \downcell{80.76}{0.43}{12} & \upcell{33.68}{2.41}{38} \\
 & TI-DPO & \downcell{89.50}{0.08}{8} & \upcell{42.97}{2.38}{38} & \bupcell{83.42}{2.23}{38} & \downcell{28.47}{2.80}{38} \\
 & TAB-PO & \bupcell{91.50}{1.92}{26} & \bupcell{48.88}{8.29}{66} & \upcell{83.38}{2.19}{38} & \bupcell{41.19}{9.92}{66} \\
\midrule

\multirow{7}{*}{\shortstack[l]{Qwen2.5\\14B}}
 & Cal-DPO & \upcell{91.30}{0.17}{8} & \downcell{50.65}{0.45}{12} & \upcell{85.31}{0.70}{18} & \upcell{45.01}{2.67}{38} \\
 & DPO & \downcell{91.07}{0.06}{8} & \downcell{49.77}{1.33}{26} & \upcell{85.43}{0.82}{18} & \upcell{43.41}{1.07}{26} \\
 & DPOP & \downcell{90.67}{0.46}{12} & \upcell{51.76}{0.66}{18} & \bupcell{85.48}{0.87}{18} & \upcell{46.49}{4.15}{52} \\
 & IPO & \downcell{90.85}{0.28}{12} & \downcell{50.78}{0.32}{12} & \upcell{85.38}{0.77}{18} & \upcell{45.49}{3.15}{38} \\
 & TDPO & \downcell{91.05}{0.08}{8} & \downcell{49.15}{1.95}{26} & \downcell{84.60}{0.01}{8} & \upcell{46.31}{3.97}{38} \\
 & TI-DPO & \downcell{90.93}{0.20}{8} & \downcell{50.92}{0.18}{8} & \upcell{84.67}{0.06}{8} & \upcell{43.60}{1.26}{26} \\
 & TAB-PO & \bupcell{91.64}{0.51}{18} & \bupcell{55.14}{4.04}{52} & \upcell{85.20}{0.59}{18} & \bupcell{52.95}{10.61}{66} \\
\bottomrule
\end{tabular}%
}
\end{snwidetable}

\subsection{TAB-PO Outperforms Sequence-Level and Token-Level DPO Variants}

TAB-PO also outperforms recent sequence-level and token-level DPO-family
baselines. Against the strongest non-TAB-PO baseline for each model--metric
cell, TAB-PO wins 39 of 42 comparisons overall: 17 of 18 on PV-Miner and 22 of
24 on SciERC. Notably, TAB-PO wins all semantic-labelling and relational-linking
comparisons, covering PV-Miner Code/Sub-code and SciERC Entity/Relation/Coreference
metrics.

The advantage is largest on token-critical structured decisions. On PV-Miner,
TAB-PO exceeds the best non-TAB-PO baseline by +4.09 Code F1 and +5.33 Sub-code
F1 on average. On SciERC, it exceeds the best non-TAB-PO baseline by +6.01
Relation F1 and +7.07 Coreference F1 on average. These results support the
central design of TAB-PO. Broader related-work positioning of TAB-PO with respect to preference-based
alignment, DPO-family objectives, token-level preference optimization, and
verifier-based RL alternatives is provided in Appendix~\ref{app:related_work}.

\subsection{TAB-PO Surpasses Frontier Proprietary Models on SciERC}

Because PV-Miner is privacy-constrained, frontier model comparisons are
conducted only on SciERC. Figure~\ref{fig:tabpo_frontier_scierc_f1} shows that TAB-PO
70B surpasses the strongest frontier proprietary model on every SciERC metric:
+2.44 Entity F1, +8.37 Relation F1, +4.68 Span F1, and +16.33 Coreference F1.

The frontier comparison also shows that TAB-PO gains are not limited to the
70B model. Using the macro average across Entity, Relation, Span, and
Coreference F1, TAB-PO 14B reaches 71.32 F1, exceeding every frontier baseline
and surpassing the best frontier-per-metric macro score of 67.00 by +4.33 F1.
Even TAB-PO 3B reaches 64.49 macro F1, surpassing 9 of the 11 frontier
baselines, approximately the 82nd percentile of the frontier-model distribution.
These results indicate that TAB-PO improves structured prediction quality not
only at the largest scale, but also for substantially smaller open-weight
models.

\paragraph{Ablation summary.}
We further assess whether TAB-PO's gains are attributable to its proposed design
components rather than incidental optimization choices. Across PV-Miner and
SciERC, performance is strongest when TAB-PO combines confusion-aware hard
negatives with expert-preferred chosen outputs. Increasing the number of
rejected outputs per input improves ontology-critical metrics, suggesting that
exposure to multiple plausible errors provides a stronger preference signal. In
contrast, weaker preference construction, less targeted perturbations, overly
permissive or overly restrictive confidence thresholds, and reduced barrier
strength lead to larger drops. These trends support the central mechanism of
TAB-PO: effective preference optimization for structured generation requires
both realistic hard negatives and token-level protection for under-confident
preferred decisions. The full diagnostic ablation study is reported in
Appendix~\ref{app:ablations}.

Additional auxiliary loss components for structured prediction, including token
weighting, length normalization, and class-balanced reweighting, are described
in Appendix~\ref{app:aux_loss_components}. These components address recurring
challenges in structured prediction, including sparse task-critical tokens,
variable-length evidence spans, and highly imbalanced ontology label distribution.

\section{Discussion}

TAB-PO is designed for settings in which preferred and rejected outputs are
low-separation structured objects, and where correctness depends on a sparse set
of schema-defining tokens. This makes the method especially suitable for
ontology-driven structured prediction, a regime we evaluate through two
complementary task settings: clinical communication mining in PV-Miner and
scientific information extraction in SciERC. These tasks differ substantially in
domain, ontology structure, and output dependencies, yet both expose the same
underlying optimization challenge: small token-level changes can determine
semantic labels, textual grounding, and relational links. The consistency of
TAB-PO's gains across these settings suggests that its mechanism is not tied to
a single ontology or dataset format, but to a broader class of token-critical
structured prediction problems.

A natural direction for future work is to extend TAB-PO beyond structured
information extraction to decision-oriented LLM systems in which a small number
of tokens can alter downstream actions. Relevant settings include clinical safety triage,
legal and regulatory compliance review, insurance claim assessment, quality
auditing, decision-support workflows, and tool-calling systems. In such settings,
the same low-separation preference geometry can arise: two outputs may be
nearly identical in surface form but differ in a decisive label, action, rule
interpretation, risk category, or tool argument. Extending TAB-PO to these
domains would test whether confusion-aware preference construction and
confidence-gated likelihood restoration can improve not only structured
prediction accuracy, but also the reliability of LLM-assisted decision-making
systems in high-stakes applications.

\section{Conclusion}

TAB-PO demonstrates that effective preference optimization for structured
generation requires objectives and preference pairs that reflect and leverage
the geometry of structured correctness. It addresses this setting through two
coordinated components: confusion-aware preference construction, which creates
schema-valid hard negatives from realistic SFT ontology-level error patterns
together with expert-guided ambiguity analysis, and a confidence-gated token
barrier, which protects under-confident preferred tokens from likelihood erosion
while reducing gradient dilution across non-critical serialization tokens. These mechanisms translate into consistent gains on the dimensions most aligned
with token-critical structured decisions. On SciERC, all six model
configurations improve both Relation and Coreference F1 over their SFT
baselines, with even the smallest gains across models still amounting to +4.04 and
+6.00 points, respectively. The benefit holds across model scales from 1.5B to
70B parameters: Qwen2.5-1.5B improves SciERC macro F1 by +12.00 points after
TAB-PO, while Llama-3.3-70B improves by +4.91 points. These results show that
TAB-PO is a strong fit for ontology-driven structured prediction.

\section*{Statements and Declarations}

\bmhead{Acknowledgements}

This work was supported by the Patient-Centered Outcomes Research Institute
(PCORI) under Award No. ME-2023C2-31367 to Samah Fodeh. We thank the
Institutional Review Board (IRB), Yale New Haven Health System (YNHH), the
Texas Association of Charitable Clinics (TXACC), and Cleveland Clinic (CC) for
facilitating institutional review, data governance, and authorized data access
for this study. The clinical component of this study used patient--provider
secure-message data obtained from these participating clinical sites and was
conducted under strict institutional governance, applicable ethical oversight
requirements, and approved privacy safeguards. Data access, processing, and
analysis were conducted in accordance with approved protocols, data-use
requirements, and institutional data-protection standards.

\bmhead{Funding}

This work was supported by the Patient-Centered Outcomes Research Institute
(PCORI) under Award No. ME-2023C2-31367 to Samah Fodeh.

\bmhead{Competing interests}

The authors declare no competing interests.

\bmhead{Author contributions}

Samah Fodeh conceived the study, supervised the overall research and data analysis plan, and critically revised the manuscript. Linhai Ma and Ganesh Puthiaraju developed and implemented the TAB-PO method. Linhai Ma, Ganesh Puthiaraju, and Afshan Khan prepared the initial manuscript draft. Srivani Talakokkul, Sreeraj Ramachandran, Elyas Irankhah, and Aimee Kendall Roundtree contributed to the task design, domain interpretation, and critical review and revision of the manuscript. All authors reviewed and approved the final version of the manuscript.

\bmhead{Ethics approval}

This study used de-identified patient--provider secure-message data obtained
from participating clinical sites, including Yale New Haven Health System,
Cleveland Clinic, and the Texas Association of Charitable Clinics, through
approved institutional governance procedures. All procedures were approved by the Yale University IRB and conducted in
accordance with relevant laws, institutional policies, and ethical guidelines
for the analysis of de-identified clinical data. The project was reviewed under
the applicable institutional governance process and determined not to constitute
human-subjects research. The privacy rights of individuals were strictly
observed, and no directly identifying patient information is reported in the
manuscript.

\bmhead{Consent to participate}

Informed consent was not applicable because the project was determined not to
constitute human-subjects research and used de-identified data under approved
institutional governance procedures.

\bmhead{Consent for publication}

Not applicable. The manuscript reports only aggregate results and does not
include identifiable individual-level patient information.

\bmhead{Data availability}

The SciERC dataset used in this study is publicly available from its original
data source. The experiments on SciERC can be reproduced using the public
SciERC data together with the code that will be made publicly available. The
PV-Miner clinical dataset generated and analyzed during the current study is
not publicly available because it is derived from privacy-protected
patient--provider secure-message data obtained from Yale New Haven Health
System, Cleveland Clinic, and the Texas Association of Charitable Clinics.
These data are subject to strict institutional governance, IRB restrictions,
data-use agreements, and privacy-protection obligations, and therefore cannot
be shared publicly. 

\bmhead{Code availability}

The code developed for supervised fine-tuning, confusion-aware preference
construction, TAB-PO optimization, baseline preference-optimization methods,
inference, and metric evaluation will be made publicly available through a
public GitHub repository.

\bibliography{custom}%

\backmatter

\begin{appendices}

\clearpage

\section{Ablation Study}
\label{app:ablations}

\definecolor{posgreen}{HTML}{2E7D32}
\definecolor{negred}{HTML}{C62828}
\definecolor{BaselineBlue}{HTML}{E8F1FA}
\definecolor{DiagHeader}{HTML}{EEF3F8}

\providecommand{\upcell}[3]{\cellcolor{posgreen!#3}#1 ($\uparrow$ #2)}
\providecommand{\downcell}[3]{\cellcolor{negred!#3}#1 ($\downarrow$ #2)}
\providecommand{\basecell}[1]{\cellcolor{BaselineBlue}#1}

\newcommand{\axiscell}[1]{\makecell[l]{#1}}

\begin{snwidetable}[!htbp]
\centering
\captionsetup{width=\linewidth,justification=raggedright,singlelinecheck=false}
\caption{
Diagnostic ablations for TAB-PO on PV-Miner and SciERC.
For PV-Miner, the reference configuration uses one rejected output per input example,
Code and Sub-code semantic-label confusions, confusion-aware hard-negative construction
with expert-preferred chosen outputs, $\tau=0.87$, $\lambda=0.5$, and $\beta=0.5$.
For SciERC, the reference configuration uses one rejected output per input example,
Entity and Relation semantic-label/relational-linking confusions, confusion-aware hard-negative
construction with expert-preferred chosen outputs, $\tau=0.87$, $\lambda=0.5$, and $\beta=0.5$.
Parentheses report absolute F1 change from the corresponding task-specific reference configuration.
Green cells indicate improvement and red cells indicate degradation; darker shading indicates larger relative change.
}
\label{tab:tabpo_pvminer_scierc_diagnostics}
\scriptsize
\setlength{\tabcolsep}{1.8pt}
\renewcommand{\arraystretch}{1.12}

\vspace{1mm}

\resizebox{\linewidth}{!}{%
\begin{tabular}[t]{@{}p{2.35cm} p{3.25cm} c c c@{}}
\toprule
\multicolumn{5}{c}{\textbf{(a) PV-Miner}} \\
\midrule
\textbf{Diagnostic axis} &
\textbf{Setting} &
\textbf{Code} &
\textbf{Sub-code} &
\textbf{Span} \\
\midrule

\rowcolor{BaselineBlue}
\textbf{Reference} &
\makecell[l]{One rejected output per input example;\\
Code+Sub-code confusion;\\
confusion-aware+expert-preferred;\\
$\tau=0.87$, $\lambda=0.5$, $\beta=0.5$}
&
\basecell{80.49}
&
\basecell{78.08}
&
\basecell{85.29}
\\

\midrule
\rowcolor{DiagHeader}
\multicolumn{5}{l}{\textbf{A. Preference data preparation diagnostics}} \\
\midrule

\axiscell{Number of \\rejected outputs \\per input \\example}
& 2
&
\upcell{82.42}{1.93}{26}
&
\upcell{79.89}{1.81}{26}
&
\upcell{86.07}{0.78}{12}
\\

&
3
&
\upcell{\textbf{83.64}}{3.15}{38}
&
\upcell{\textbf{82.60}}{4.52}{52}
&
\upcell{\textbf{86.97}}{1.68}{18}
\\
\midrule

\axiscell{Preference\\construction}
& Expert-preferred rejected output
&
\downcell{79.24}{1.25}{18}
&
\downcell{76.94}{1.14}{18}
&
\upcell{85.44}{0.15}{8}
\\
\midrule

\axiscell{Confusion-aware\\preference\\construction}
& Extra record
&
\downcell{80.01}{0.48}{12}
&
\downcell{77.88}{0.20}{8}
&
\downcell{84.59}{0.70}{12}
\\

&
Missing record
&
\upcell{80.82}{0.33}{8}
&
\downcell{77.77}{0.31}{8}
&
\upcell{86.50}{1.21}{18}
\\

\midrule
\rowcolor{DiagHeader}
\multicolumn{5}{l}{\textbf{B. TAB-PO objective diagnostics}} \\
\midrule

\axiscell{Barrier\\confidence\\threshold}
& $\tau=0.25$
&
\downcell{61.58}{18.91}{80}
&
\downcell{48.58}{29.50}{80}
&
\downcell{78.81}{6.48}{66}
\\

&
$\tau=0.50$
&
\downcell{78.50}{1.99}{26}
&
\downcell{76.45}{1.63}{26}
&
\downcell{85.01}{0.28}{8}
\\

&
$\tau=0.95$
&
\downcell{80.31}{0.18}{8}
&
\downcell{77.90}{0.18}{8}
&
\upcell{85.72}{0.43}{12}
\\
\midrule

\axiscell{Barrier weight}
& $\lambda=0.1$
&
\downcell{79.13}{1.36}{18}
&
\downcell{76.99}{1.09}{18}
&
\downcell{84.93}{0.36}{8}
\\

&
$\lambda=2.0$
&
\upcell{81.55}{1.06}{18}
&
\upcell{80.81}{2.73}{26}
&
\upcell{86.63}{1.34}{18}
\\
\midrule

\axiscell{Preference\\temperature}
& $\beta=0.25$
&
\upcell{81.07}{0.58}{12}
&
\upcell{78.11}{0.03}{8}
&
\upcell{85.67}{0.38}{8}
\\

&
$\beta=0.75$
&
\downcell{79.85}{0.64}{12}
&
\upcell{78.33}{0.25}{8}
&
\upcell{86.21}{0.92}{18}
\\

\noalign{\vskip 2.9ex}

\bottomrule
\end{tabular}%
\hspace{5pt}%
\begin{tabular}[t]{@{}p{2.35cm} p{3.25cm} c c c c@{}}
\toprule
\multicolumn{6}{c}{\textbf{(b) SciERC}} \\
\midrule
\textbf{Diagnostic axis} &
\textbf{Setting} &
\textbf{Entity} &
\textbf{Relation} &
\textbf{Span} &
\textbf{Coreference} \\
\midrule

\rowcolor{BaselineBlue}
\textbf{Reference} &
\makecell[l]{One rejected output per input example;\\
Entity+Relation confusion;\\
confusion-aware+expert-preferred;\\
$\tau=0.87$, $\lambda=0.5$, $\beta=0.5$}
&
\basecell{91.04}
&
\basecell{46.67}
&
\basecell{84.98}
&
\basecell{35.11}
\\

\midrule
\rowcolor{DiagHeader}
\multicolumn{6}{l}{\textbf{A. Preference data preparation diagnostics}} \\
\midrule

\axiscell{Number of \\rejected outputs \\per input \\example}
& 2
&
\upcell{\textbf{91.41}}{0.37}{12}
&
\upcell{46.90}{0.23}{12}
&
\downcell{84.94}{0.04}{8}
&
\upcell{\textbf{37.04}}{1.93}{52}
\\

&
3
&
\upcell{91.37}{0.33}{12}
&
\upcell{\textbf{47.24}}{0.57}{26}
&
\upcell{85.33}{0.35}{12}
&
\upcell{36.06}{0.95}{38}
\\
\midrule

\axiscell{Preference\\construction}
& Expert-preferred rejected output
&
\downcell{90.79}{0.25}{12}
&
\downcell{44.29}{2.38}{52}
&
\downcell{84.14}{0.84}{18}
&
\downcell{33.54}{1.57}{52}
\\
\midrule

\axiscell{Confusion-aware\\pair generation}
& Missing entity
&
\downcell{90.81}{0.23}{12}
&
\upcell{46.80}{0.13}{12}
&
\upcell{\textbf{85.93}}{0.95}{26}
&
\downcell{34.98}{0.13}{12}
\\

&
Extra entity
&
\downcell{90.72}{0.32}{12}
&
\downcell{44.89}{1.78}{38}
&
\downcell{84.45}{0.53}{18}
&
\downcell{34.14}{0.97}{38}
\\

&
Extra relation
&
\downcell{90.99}{0.05}{8}
&
\downcell{45.65}{1.02}{38}
&
\downcell{84.91}{0.07}{8}
&
\downcell{33.51}{1.60}{52}
\\

&
Extra coreference
&
\downcell{90.74}{0.30}{12}
&
\downcell{45.18}{1.49}{38}
&
\downcell{84.23}{0.75}{18}
&
\downcell{33.92}{1.19}{38}
\\

\midrule
\rowcolor{DiagHeader}
\multicolumn{6}{l}{\textbf{B. TAB-PO objective diagnostics}} \\
\midrule

\axiscell{Barrier\\confidence\\threshold}
& $\tau=0.25$
&
\downcell{90.05}{0.99}{26}
&
\downcell{43.13}{3.54}{52}
&
\downcell{83.76}{1.22}{26}
&
\downcell{23.71}{11.40}{80}
\\

&
$\tau=0.50$
&
\downcell{90.31}{0.73}{18}
&
\downcell{43.98}{2.69}{52}
&
\downcell{83.88}{1.10}{26}
&
\downcell{26.87}{8.24}{80}
\\

&
$\tau=0.95$
&
\downcell{90.79}{0.25}{12}
&
\downcell{46.33}{0.34}{18}
&
\downcell{84.62}{0.36}{12}
&
\downcell{34.70}{0.41}{26}
\\
\midrule

\axiscell{Barrier weight}
& $\lambda=0.1$
&
\downcell{90.55}{0.49}{18}
&
\downcell{45.04}{1.63}{38}
&
\downcell{84.64}{0.34}{12}
&
\downcell{33.23}{1.88}{52}
\\

&
$\lambda=2.0$
&
\upcell{91.21}{0.17}{8}
&
\upcell{46.80}{0.13}{12}
&
\downcell{84.50}{0.48}{18}
&
\upcell{35.41}{0.30}{18}
\\
\midrule

\axiscell{Preference\\temperature}
& $\beta=0.25$
&
\downcell{91.03}{0.01}{8}
&
\downcell{46.31}{0.36}{18}
&
\upcell{85.10}{0.12}{8}
&
\upcell{35.28}{0.17}{12}
\\

&
$\beta=0.75$
&
\downcell{90.34}{0.70}{18}
&
\downcell{45.80}{0.87}{26}
&
\downcell{84.20}{0.78}{18}
&
\downcell{34.73}{0.38}{26}
\\

\bottomrule
\end{tabular}%
}
\end{snwidetable}

This appendix provides an empirical stress test of TAB-PO's proposed mechanism.
We separately evaluate its two central design choices: confusion-aware
preference construction and confidence-gated likelihood restoration. We then
analyze sensitivity to the number of rejected outputs per input, the type of
preference construction, the type of confusion-aware perturbation, and key
optimization hyperparameters, including the preference temperature, barrier
confidence threshold, and barrier strength. Together, these ablations assess
whether TAB-PO's gains arise from its intended mechanism rather than from
incidental negative-sampling choices or hyperparameter effects.

Table~\ref{tab:tabpo_pvminer_scierc_diagnostics} analyzes the components of
TAB-PO. Increasing the number of rejected outputs per input example improves PV
Miner by +1.51 F1 with two negatives and +3.12 F1 with three negatives on
average. On SciERC, the corresponding gains are smaller but consistently
positive, with +0.62 F1 for two negatives and +0.55 F1 for three negatives.
This suggests that additional hard negatives can improve coverage, but the
central gains do not come merely from adding more preference pairs.

The confidence-gated barrier is essential. Lowering the threshold to
$\tau=0.25$ causes large average drops of -18.30 F1 on PV Miner and -4.29 F1 on
SciERC, while $\tau=0.50$ also degrades performance. A weak barrier
($\lambda=0.1$) reduces performance on both tasks, with average drops of -0.94
F1 on PV Miner and -1.09 F1 on SciERC. These results show that the barrier must
activate on the appropriate under-confident preferred tokens with sufficient
strength. Conversely, moderate changes to the preference temperature $\beta$
produce smaller effects, indicating that TAB-PO is less sensitive to the margin
temperature than to the confidence-gated anchoring mechanism.

The preference-construction diagnostics further rule out a simple data-only
explanation. Replacing the default confusion-aware construction with
expert-preferred negative construction reduces average performance by -0.75 F1
on PV Miner and -1.26 F1 on SciERC. Extra-record and missing-record perturbations
produce mixed effects, whereas the full confusion-aware construction provides
the strongest and most stable reference configuration. These ablations support
the intended mechanism: TAB-PO works because it combines realistic
ontology-level hard negatives with a token-level barrier that selectively
protects under-confident preferred schema tokens.

\section{Mathematical Intuition for Gradient Dilution and Preferred-Token Erosion}
\label{app:gradient_dilution}

This appendix gives the mathematical intuition behind two failure modes that can arise when DPO-style preference optimization is applied to low-separation structured outputs: gradient dilution and preferred-token erosion. The main
paper reports empirical mechanism-level diagnostics for TAB-PO and DPO-family variants in Section~\ref{sec:mechanistic_diagnostics}; here we focus only on why these failure modes arise under DPO.

Given a preference triple, let $Y^{+}$ and $Y^{-}$
denote the preferred and rejected structured record sets for the same input,
where each record has the form $y_i=(\ell_i,\Pi_i,R_i)$. Let $x$ denote the
input prompt. We serialize the two structured outputs as completion strings
$Y_s^{+}$ and $Y_s^{-}$. Let
$u^{+}=(u_1^{+},\ldots,u_{T^{+}}^{+})$ and
$u^{-}=(u_1^{-},\ldots,u_{T^{-}}^{-})$ denote the completion-token subsequences
corresponding to $Y_s^{+}$ and $Y_s^{-}$, respectively. The reference model is
the supervised model $p_{\mathrm{SFT}}$.

For a single preference pair, standard DPO minimizes

\begin{align}
\mathcal{L}_{\mathrm{DPO}}(\theta)
=
-\log \sigma
\Biggl(
\beta
\Biggl[
\log
\frac{
p_{\theta}(Y_s^{+}\mid x)
}{
p_{\mathrm{SFT}}(Y_s^{+}\mid x)
}
-
\log
\frac{
p_{\theta}(Y_s^{-}\mid x)
}{
p_{\mathrm{SFT}}(Y_s^{-}\mid x)
}
\Biggr]
\Biggr),
\label{eq:standard_dpo}
\end{align}

where $\beta>0$ controls the preference strength. Define the reference-adjusted
preference margin

\begin{align}
\Delta_{\theta}(x,Y_s^{+},Y_s^{-})
=
\log
\frac{
p_{\theta}(Y_s^{+}\mid x)
}{
p_{\mathrm{SFT}}(Y_s^{+}\mid x)
}
-
\log
\frac{
p_{\theta}(Y_s^{-}\mid x)
}{
p_{\mathrm{SFT}}(Y_s^{-}\mid x)
}.
\label{eq:standard_dpo_margin}
\end{align}

The gradient of the DPO loss is

\begin{align}
\nabla_{\theta}\mathcal{L}_{\mathrm{DPO}}
=
-\beta \sigma(-\beta\Delta_{\theta})
\Bigl[
\nabla_{\theta}\log p_{\theta}(Y_s^{+}\mid x)
-
\nabla_{\theta}\log p_{\theta}(Y_s^{-}\mid x)
\Bigr].
\end{align}
Using the autoregressive factorization
\[
\log p_{\theta}(Y_s^{\pm}\mid x)
=
\sum_{t=1}^{T^{\pm}}
\log p_{\theta}(u_t^{\pm}\mid x,u_{<t}^{\pm}),
\]
we obtain

\begin{align}
\nabla_{\theta}\mathcal{L}_{\mathrm{DPO}}
=
-\beta \sigma(-\beta\Delta_{\theta})
\Biggl[
\sum_{t=1}^{T^{+}}
\nabla_{\theta}\log p_{\theta}
\bigl(
u_t^{+}
\mid x,u_{<t}^{+}
\bigr)
-
\sum_{t=1}^{T^{-}}
\nabla_{\theta}\log p_{\theta}
\bigl(
u_t^{-}
\mid x,u_{<t}^{-}
\bigr)
\Biggr].
\end{align}
Let $t_0$ denote the first completion-token position at which $u^{+}$ and
$u^{-}$ differ, or the first position at which one completion terminates. For
$t<t_0$, the preferred and rejected completions contain identical tokens under
identical autoregressive prefixes, so their gradient terms cancel. The remaining
DPO update is therefore determined by tokens at and after the first divergence. The number of canceled pre-divergence tokens is task- and serialization-dependent. When these shared preferred tokens are under-confident under the SFT initialization, standard DPO provides no direct restoration signal because their preferred and rejected gradient contributions cancel exactly. Thus, SFT uncertainty can remain uncorrected even before the first structured divergence. TAB-PO addresses this gap by applying confidence-gated restoration to under-confident preferred tokens, including tokens whose DPO gradient contribution vanishes due to pre-divergence cancellation.

We group these remaining positions into three categories. Critical tokens are
tokens that directly determine structured prediction correctness, including
ontology labels, grounded evidence spans, relation labels, relation arguments,
sentence identifiers, and coreference links. JSON-scaffolding tokens are purely
structural serialization tokens, such as braces, brackets, commas, colons, and
quotation marks. Non-critical tokens include schema field names and other
formatting or repeated tokens that support parseability but are not themselves
scored task decisions.

Let $\mathcal{C}^{a}$ denote the critical token positions and
$\mathcal{N}^{a}$ denote the union of JSON-scaffolding and non-critical token
positions for $a\in\{+,-\}$ after the first divergence. The remaining DPO
gradient can be decomposed as

\begin{align}
G_{\mathrm{crit}}
=
\sum_{t\in\mathcal{C}^{+}}
\nabla_{\theta}\log p_{\theta}
\bigl(
u_t^{+}\mid x,u_{<t}^{+}
\bigr)
-
\sum_{t\in\mathcal{C}^{-}}
\nabla_{\theta}\log p_{\theta}
\bigl(
u_t^{-}\mid x,u_{<t}^{-}
\bigr).
\end{align}

\begin{align}
G_{\mathrm{non}}
=
\sum_{t\in\mathcal{N}^{+}}
\nabla_{\theta}\log p_{\theta}
\bigl(
u_t^{+}\mid x,u_{<t}^{+}
\bigr)
-
\sum_{t\in\mathcal{N}^{-}}
\nabla_{\theta}\log p_{\theta}
\bigl(
u_t^{-}\mid x,u_{<t}^{-}
\bigr).
\end{align}
Thus, the post-divergence DPO gradient can be summarized as
\begin{equation}
\nabla_{\theta}\mathcal{L}_{\mathrm{DPO}}
\propto
-
\left(
G_{\mathrm{crit}}+G_{\mathrm{non}}
\right).
\end{equation}

In low-separation structured preferences, the number of tokens that directly
encode the scored decision is small relative to the number of serialization and
non-critical tokens that remain in the completion:
\begin{equation}
|\mathcal{C}^{+}|+|\mathcal{C}^{-}|
\ll
|\mathcal{N}^{+}|+|\mathcal{N}^{-}|.
\end{equation}

Moreover, after the first divergence, even surface-identical JSON punctuation
or schema field names may occur under different autoregressive prefixes, so
their gradient terms need not cancel. Standard DPO can therefore allocate update
mass to tokens that do not determine annotation correctness. We refer to this
gradient-allocation mismatch as \emph{gradient dilution}.

This analysis also exposes a related failure mode: \emph{preferred-token
erosion}. Standard DPO optimizes only the relative sequence-level margin between
the preferred and rejected completions, and therefore does not impose a lower
bound on the likelihood of any individual preferred token. Let
$\alpha_t^{+}(\theta)=\log p_{\theta}(u_t^{+}\mid x,u_{<t}^{+})$ and
$\alpha_t^{-}(\theta)=\log p_{\theta}(u_t^{-}\mid x,u_{<t}^{-})$. Expanding
Eq.~\ref{eq:standard_dpo_margin}, we have

\begin{align}
\Delta_{\theta}
=
\sum_{t=1}^{T^{+}}
\left[
\alpha_t^{+}(\theta)-\alpha_t^{+}(\mathrm{SFT})
\right]
-
\sum_{t=1}^{T^{-}}
\left[
\alpha_t^{-}(\theta)-\alpha_t^{-}(\mathrm{SFT})
\right].
\end{align}
This aggregate margin can increase even when some preferred-token likelihoods
decrease. To see this, consider an update from $\theta$ to $\theta'$, and define
$\delta_t^{+}=\alpha_t^{+}(\theta')-\alpha_t^{+}(\theta)$ and
$\delta_t^{-}=\alpha_t^{-}(\theta')-\alpha_t^{-}(\theta)$. Since
$p_{\mathrm{SFT}}$ is fixed,
\begin{equation}
\Delta_{\theta'}-\Delta_{\theta}
=
\sum_{t=1}^{T^{+}}\delta_t^{+}
-
\sum_{t=1}^{T^{-}}\delta_t^{-}.
\end{equation}
Thus, for any preferred token $k$ with $\delta_k^{+}<0$, the margin can still
increase whenever
\begin{equation}
\sum_{t\neq k}\delta_t^{+}
-
\sum_{t=1}^{T^{-}}\delta_t^{-}
>
|\delta_k^{+}|.
\end{equation}
A successful DPO update therefore need not preserve the absolute likelihood of
every correct preferred token. This is especially problematic for
ontology-driven structured prediction: rare or ontology-specific schema tokens
are often precisely the tokens on which the SFT model is least confident. The
model may improve the relative preference margin while becoming less reliable
at generating the correct structured record. Thus, a purely relative
sequence-level objective provides no token-level protection for under-confident
preferred schema tokens.

The diagnostics in Figure~\ref{fig:tabpo_training_diagnostics} support this
mechanism-level interpretation. TAB-PO's barrier activity is concentrated on
critical schema tokens, with much weaker activation on JSON scaffolding and
non-critical tokens. The gradient-mass heatmaps show that sequence-level DPO
variants can allocate substantial update mass to serialization or non-critical
tokens, whereas TAB-PO concentrates learning signal on the sparse tokens that
determine structured prediction correctness avoiding \emph{gradient dilution}. The preference-probability and
reference-adjusted-margin curves further show that TAB-PO maintains strong
preference separation while applying targeted restoration to under-confident
preferred tokens avoiding \emph{token erosion}.

\section{Auxiliary Loss Components for Structured Prediction}
\label{app:aux_loss_components}

TAB-PO targets ontology-driven structured prediction, where most serialized
tokens encode schema scaffolding while only a small subset determines semantic
labelling, textual grounding, or relational linking correctness. In addition to
the core preference objective and confidence-gated barrier, we consider three
auxiliary loss components that address recurring pathologies in structured
prediction datasets: sparse critical tokens, large variation in completion
length, and long-tailed ontology-label distributions.

\subsection{Token-Weighted Preference Aggregation}
\label{app:token_weighting}

Structured prediction errors are often localized to a few schema-defining
tokens: semantic-label tokens, textual-grounding tokens, relation labels and direction pointers. Standard sequence-level aggregation weights these tokens the same as JSON punctuation,
field names, and repeated schema scaffolding. Motivated by loss-reweighting
methods that increase the influence of informative or under-emphasized training
signals~\cite{helm2025token}, we introduce token weights that
concentrate the preference signal on tokens that determine structured prediciton correctness.

For each serialized completion $Y_s$, define token-index sets
$\mathcal{T}_{\mathrm{SL}}(Y_s)$, $\mathcal{T}_{\mathrm{TG}}(Y_s)$, and
$\mathcal{T}_{\mathrm{RL}}(Y_s)$ for semantic labelling, textual grounding, and
relational linking, respectively. The remaining completion tokens are assigned
to $\mathcal{T}_{0}(Y_s)$. We define
\begin{equation}
\label{eq:aux_token_weights}
\begin{aligned}
w_t(Y_s)
=
w_{\mathrm{SL}}\mathbb{I}[t\in\mathcal{T}_{\mathrm{SL}}(Y_s)]
+
w_{\mathrm{TG}}\mathbb{I}[t\in\mathcal{T}_{\mathrm{TG}}(Y_s)]
&+
w_{\mathrm{RL}}\mathbb{I}[t\in\mathcal{T}_{\mathrm{RL}}(Y_s)]
\\&+
w_{0}\mathbb{I}[t\in\mathcal{T}_{0}(Y_s)],
\end{aligned}
\end{equation}
where
$w_{\mathrm{SL}},w_{\mathrm{TG}},w_{\mathrm{RL}}\geq w_0\geq 0$. The token-weighted completion log-likelihood is
\begin{equation}
\label{eq:aux_weighted_likelihood}
\begin{aligned}
\mu_\theta^{w}(Y_s\mid x)
=
\sum_{t=1}^{T}
w_t(Y_s)
\log p_\theta(u_t\mid x,u_{<t}).
\end{aligned}
\end{equation}
Replacing the unweighted sequence likelihood with
$\mu_\theta^w$ gives the weighted reference-adjusted advantage
\begin{equation}
\label{eq:aux_weighted_delta}
\begin{aligned}
\Delta_\theta^{w}(x,Y_s^+,Y_s^-)
=
\Big[
\mu_\theta^{w}(Y_s^+\mid x)
-
\mu_\theta^{w}(Y_s^-\mid x)
\Big]
-
\Big[
\mu_{\mathrm{SFT}}^{w}(Y_s^+\mid x)
-
\mu_{\mathrm{SFT}}^{w}(Y_s^-\mid x)
\Big],
\end{aligned}
\end{equation}
where $p_{\mathrm{SFT}}$ is fixed. The corresponding preference loss is
\begin{equation}
\label{eq:aux_weighted_pref_loss}
\begin{aligned}
\mathcal{L}_{\mathrm{pref}}^{w}(\theta)
=
-
\mathbb{E}_{\mathcal{D}_{\mathrm{pref}}}
\left[
\log\sigma\!\left(
\beta\Delta_\theta^{w}(x,Y_s^+,Y_s^-)
\right)
\right].
\end{aligned}
\end{equation}
This weighting changes only how token evidence is aggregated; it does not alter
preference-pair construction. It is intended to increase the effective learning
signal on semantic labels, grounded spans, and relational pointers while
reducing gradient dilution from shared serialization scaffolding.

A corpus-level diagnostic over the canonical serialized gold completions further
supports this design: critical serialized tokens account for only about
$20.0\%$ of PV Miner completion tokens and $13.4\%$ of SciERC completion tokens
on average. Thus, most completion tokens are not directly responsible for the
scored structured decision, making uniform sequence-level aggregation a poor
match for token-critical structured prediction.

\subsection{Per-Example Length Normalization}
\label{app:length_normalization}

Structured outputs can differ sharply in length. Some examples contain a single
short grounded record, whereas others contain many records, long evidence spans,
or multiple relation and coreference links. Length effects are a known source of
bias in sequence scoring and generation~\cite{wu2016google,murray2018correcting}.
In structured prediction, this issue is especially prominent: without
normalization, long serialized completions can contribute more total gradient
mass simply because they contain more completion tokens, thereby biasing the
model toward longer outputs and over-emphasizing examples with many records or
long evidence spans rather than examples with more difficult structured decisions.

For instance, PV Miner completions range from 35 to
1,140 serialized tokens, a $32.6{\times}$ difference, while non-empty SciERC
completions range from 116 to 3,112 serialized tokens, a $26.8{\times}$
difference. This variation is driven by the number of extracted records, span length, and relation/coreference structure. We therefore normalize each weighted
log-likelihood by its own token-weight mass. To make each preference pair contribute at a
comparable scale regardless of serialized length, we normalize each weighted
log-likelihood by its own token-weight mass:
\begin{equation}
\label{eq:aux_weight_mass}
Z_w(Y_s)
=
\max\!\left(
\epsilon,
\sum_{t=1}^{T} w_t(Y_s)
\right),
\end{equation}
where $\epsilon>0$ is a small constant for numerical stability. The length-normalized weighted
completion log-likelihood is
\begin{equation}
\label{eq:aux_length_normalized_likelihood}
\begin{aligned}
\bar{\mu}_\theta^{w}(Y_s\mid x)
&=
\frac{1}{Z_w(Y_s)}
\sum_{t=1}^{T}
w_t(Y_s)
\\
&\qquad
\log p_\theta(u_t\mid x,u_{<t}) .
\end{aligned}
\end{equation}
When length normalization is enabled, the reference-adjusted advantage becomes

\begin{equation}
\label{eq:aux_length_normalized_delta}
\begin{aligned}
\bar{\Delta}_\theta^{w}(x,Y_s^+,Y_s^-)
=
\Big[
\bar{\mu}_\theta^{w}(Y_s^+\mid x)
-
\bar{\mu}_\theta^{w}(Y_s^-\mid x)
\Big]
-
\Big[
\bar{\mu}_{\mathrm{SFT}}^{w}(Y_s^+\mid x)
-
\bar{\mu}_{\mathrm{SFT}}^{w}(Y_s^-\mid x)
\Big].
\end{aligned}
\end{equation}
The length-normalized preference loss is
\begin{equation}
\label{eq:aux_length_normalized_pref_loss}
\begin{aligned}
\mathcal{L}_{\mathrm{pref}}^{w,\mathrm{LN}}(\theta)
=
-
\mathbb{E}_{\mathcal{D}_{\mathrm{pref}}}
\Big[
\log\sigma\!\Big(
\beta\bar{\Delta}_\theta^{w}
(x,Y_s^+,Y_s^-)
\Big)
\Big] .
\end{aligned}
\end{equation}

The same normalization is applied to the confidence-gated barrier. Let
$g_t^\theta(x,u^+)\in\{0,1\}$ denote the TAB-PO gate for preferred token
$u_t^+$. The weighted, length-normalized barrier is

\begin{equation}
\label{eq:aux_weighted_ln_barrier}
\begin{aligned}
\mathcal{L}_{\mathrm{barrier}}^{w,\mathrm{LN}}(\theta)
=
\mathbb{E}_{\mathcal{D}_{\mathrm{pref}}}
\Bigg[
\frac{
\begin{gathered}
\sum_{t=1}^{T^+}
g_t^\theta(x,u^+)
w_t(Y_s^+)
\big[-\log p_\theta(u_t^+\mid x,u_{<t}^+)\big]
\end{gathered}
}{
\max\!\left(
\epsilon,
\sum_{t=1}^{T^+}
g_t^\theta(x,u^+)w_t(Y_s^+)
\right)
}
\Bigg] .
\end{aligned}
\end{equation}

When no preferred tokens are gated, the numerator is zero and the barrier
contributes zero. This normalization ensures that examples with long spans,
many records, or many relational links do not dominate optimization solely
through completion length.

\subsection{Class-Balanced Example Reweighting}
\label{app:class_balancing}

\begin{figure}[!h] \centering \begin{subfigure}[t]{0.48\linewidth} \centering \includegraphics[width=\linewidth]{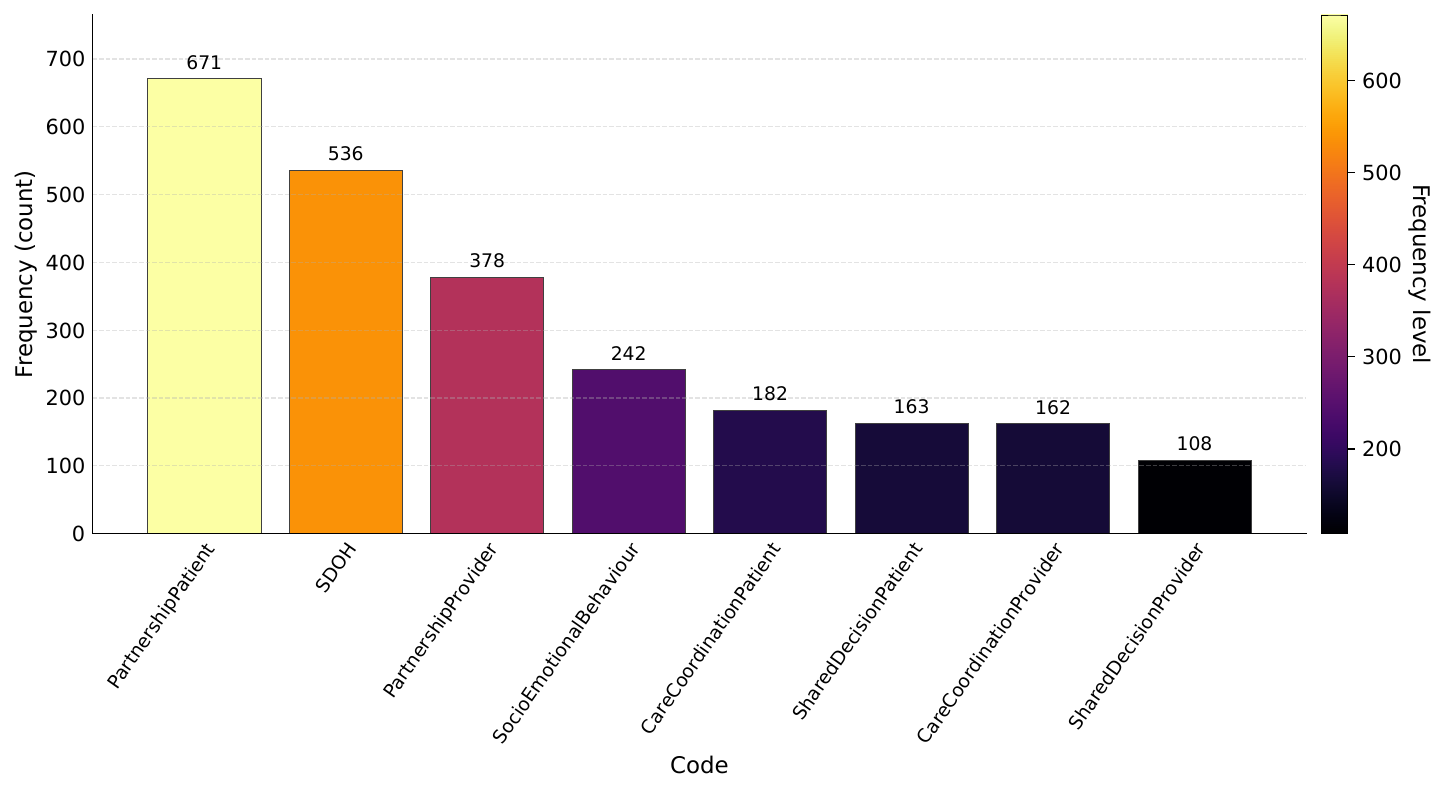} \caption{Code-level PV Miner labels. Counts reflect annotation instances rather than unique messages.} \label{fig:code_dist} \end{subfigure} \hfill \begin{subfigure}[t]{0.48\linewidth} \centering \includegraphics[width=\linewidth]{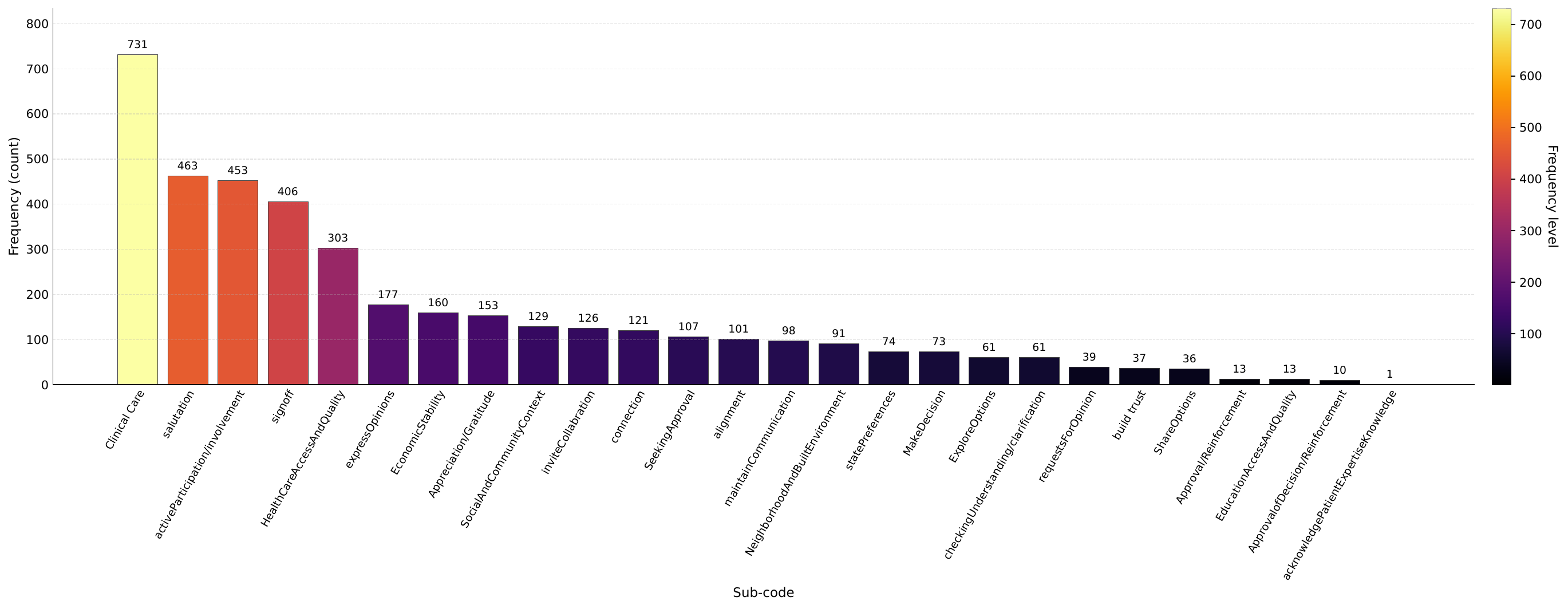} \caption{Sub-code-level PV Miner labels. Sub-codes represent fine-grained categories under the Code-level ontology.} \label{fig:subcode_dist} \end{subfigure} \caption{Frequency distributions of PV Miner structured prediction labels across the annotated patient messages.} \label{fig:pv_label_dist} \end{figure}

\begin{figure}[!h] \centering \begin{subfigure}[t]{0.48\linewidth} \centering \includegraphics[width=\linewidth]{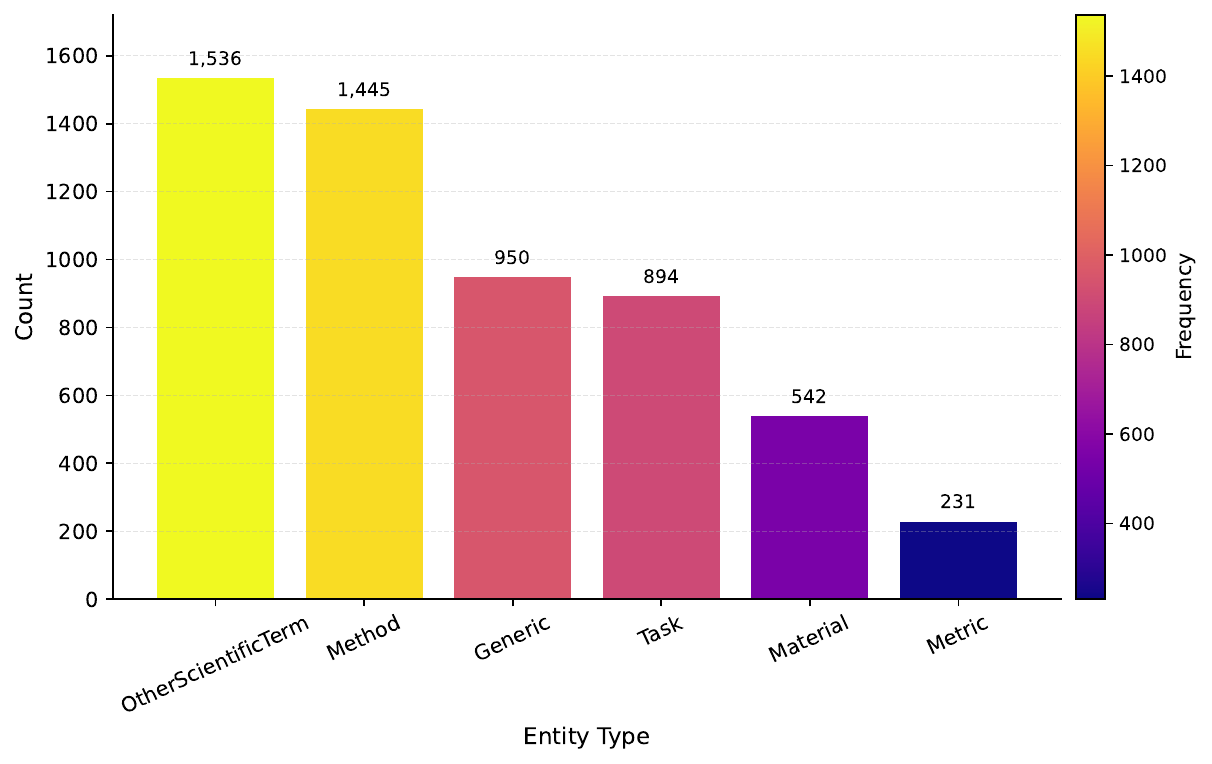} \caption{SciERC entity types. Counts correspond to entity annotation instances in the training split.} \label{fig:scierc_entity_dist} \end{subfigure} \hfill \begin{subfigure}[t]{0.48\linewidth} \centering \includegraphics[width=\linewidth]{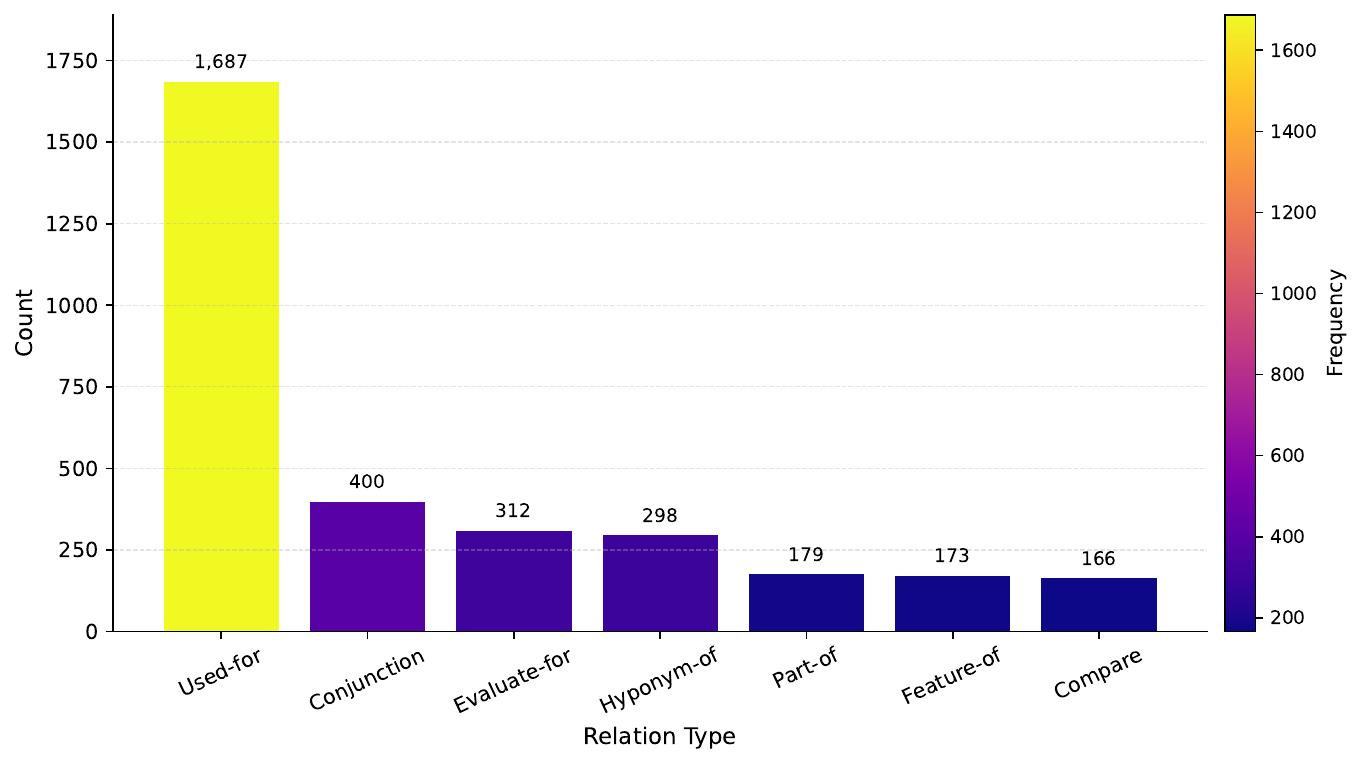} \caption{SciERC relation types. Counts correspond to relation annotation instances in the training split.} \label{fig:scierc_relation_dist} \end{subfigure} \caption{Frequency distributions of SciERC structured prediction labels in the training split.} \label{fig:scierc_entity_relation_dist} \end{figure}

Structured prediction datasets often have long-tailed ontology distributions.
This is apparent in the frequency plots for PV Miner Code and Sub-code labels
(Figures~\ref{fig:code_dist} and~\ref{fig:subcode_dist}) and for SciERC Entity
and Relation types (Figure~\ref{fig:scierc_entity_relation_dist}). In PV Miner,
the most frequent Code, \texttt{PartnershipPatient}, appears 671 times, whereas
the least frequent Code, \texttt{SharedDecisionProvider}, appears 108 times.
The Sub-code distribution is more extreme: \texttt{Clinical Care} appears 731
times, while \texttt{acknowledgePatientExpertiseKnowledge} appears once. In
SciERC, \texttt{OtherScientificTerm} appears 1,536 times compared with 231
occurrences of \texttt{Metric}, and \texttt{Used-for} appears 1,687 times
compared with 166 occurrences of \texttt{Compare}. These skews can cause common
labels to dominate minibatch updates and leave rare semantic or relational
labels under-optimized.

We therefore apply example-level class-balanced reweighting using the
effective-number principle~\cite{cui2019classbalanced}. Let
$L_{\mathrm{bal}}$ denote the task-specific set of ontology labels used for
class balancing. For a hierarchical labelling task, this set may include coarse
and fine-grained semantic labels. For a relational extraction task, it may
include entity labels, relation labels, and, when applicable, coreference or
linking decisions.

For each gold structured output $Y^{(i)}$, let
$\mathrm{Lab}(Y^{(i)})\subseteq L_{\mathrm{bal}}$ denote the set of ontology
labels selected for class balancing that appear in the labels or typed links of
its records. For each such label $\ell\in L_{\mathrm{bal}}$, let $n_\ell$ be
its frequency in the training set. We define
\begin{equation} 
\label{eq:aux_effective_number}
\eta(\ell)
=
\frac{1-\rho}{1-\rho^{n_\ell}},
\qquad
\rho\in[0,1).
\end{equation}

 

The example weight is
\begin{equation}
\label{eq:aux_example_weight}
\begin{aligned}
\omega_i
&=
\max_{\ell\in\mathrm{Lab}(Y^{(i)})}\eta(\ell),
\\
\omega_i
&\leftarrow
\min(\omega_i,\omega_{\max}).
\end{aligned}
\end{equation}
If $\mathrm{Lab}(Y^{(i)})=\emptyset$, we set $\omega_i=1$. The max operation
emphasizes examples containing at least one rare ontology label, while clipping
by $\omega_{\max}$ prevents unstable updates from extremely rare labels.

Let $\mathcal{L}_{\mathrm{TAB\text{-}PO}}^{(i)}(\theta)$ denote the per-example
TAB-PO loss, including the preference term and confidence-gated barrier,
optionally with token weighting and length normalization. For a minibatch of size $B$, the class-balanced objective is
\begin{equation}
\label{eq:aux_class_balanced_loss}
\begin{aligned}
\mathcal{L}_{\mathrm{CB}}(\theta)
=
\frac{
\sum_{i=1}^{B}
\omega_i
\mathcal{L}_{\mathrm{TAB\text{-}PO}}^{(i)}(\theta)
}{
\sum_{i=1}^{B}\omega_i
}.
\end{aligned}
\end{equation}
This normalized weighted average preserves the minibatch loss scale while
increasing the influence of examples containing rare semantic labels, rare
relation labels, or rare linking decisions.

\subsection{Effect of the Auxiliary Components}
\label{app:aux_results}

\definecolor{posgreen}{HTML}{2E7D32}
\definecolor{negred}{HTML}{C62828}
\definecolor{BaselineBlue}{HTML}{E8F1FA}

\newcommand{\poscell}[2]{\cellcolor{posgreen!#1}#2}
\newcommand{\negcell}[2]{\cellcolor{negred!#1}#2}
\renewcommand{\basecell}[1]{\cellcolor{BaselineBlue}#1}

\begin{snwidetable}[!t]
\centering
\scriptsize
\setlength{\tabcolsep}{3pt}
\renewcommand{\arraystretch}{1.08}
\caption{F1 results of Plain TAB-PO and TAB-PO variants with class balancing, length normalization, and token weighting on PV Miner and SciERC. Cells for the modified TAB-PO settings are colored by the F1 change relative to Plain TAB-PO; green indicates improvement and red indicates degradation, with stronger color intensity indicating a larger absolute change.}
\label{tab:tabpo_ablation_class_length_token_side_by_side}

\vspace{2pt}

\resizebox{\linewidth}{!}{%
\begin{tabular}[t]{@{}lccc@{}}
\toprule
\multicolumn{4}{c}{\textbf{(a) PV Miner}} \\
\midrule
\textbf{Setting} 
& \textbf{Code} 
& \textbf{Sub-code} 
& \textbf{Span} \\
\midrule
Plain TAB-PO
& \basecell{80.49}
& \basecell{78.08}
& \basecell{85.29} \\

Plain TAB-PO + Class balancing
& \poscell{12}{81.39}
& \poscell{14}{78.77}
& \poscell{7}{85.59} \\

Plain TAB-PO + Length normalization
& \poscell{15}{81.79}
& \poscell{10}{78.32}
& \poscell{14}{86.37} \\

Plain TAB-PO + Token weighting
& \poscell{8}{80.86}
& \poscell{17}{79.58}
& \poscell{8}{85.69} \\
\bottomrule
\end{tabular}%
\hspace{10pt}%
\begin{tabular}[t]{@{}lcccc@{}}
\toprule
\multicolumn{5}{c}{\textbf{(b) SciERC}} \\
\midrule
\textbf{Setting} 
& \textbf{Entity} 
& \textbf{Relation} 
& \textbf{Span} 
& \textbf{Coreference} \\
\midrule
Plain TAB-PO
& \basecell{91.04}
& \basecell{46.67}
& \basecell{84.98}
& \basecell{35.11} \\

Plain TAB-PO + Class balancing
& \poscell{10}{91.62}
& \poscell{10}{47.21}
& \poscell{2}{85.01}
& \poscell{7}{35.68} \\

Plain TAB-PO + Length normalization
& \poscell{8}{91.38}
& \poscell{7}{46.97}
& \poscell{10}{85.63}
& \negcell{3}{34.83} \\

Plain TAB-PO + Token weighting
& \poscell{8}{91.44}
& \poscell{10}{47.09}
& \poscell{8}{85.12}
& \poscell{15}{35.81} \\
\bottomrule
\end{tabular}%
}
\end{snwidetable}

Table~\ref{tab:tabpo_ablation_class_length_token_side_by_side} shows that each
auxiliary component improves the metric most aligned with its intended role.
Class balancing primarily strengthens ontology-level prediction: on PV Miner,
Code F1 increases from 80.49 to 81.39 and Sub-code F1 increases from 78.08 to
78.77. Token weighting gives the clearest gain on the fine-grained Sub-code
metric, increasing PV Miner Sub-code F1 from 78.08 to 79.58, suggesting that
placing more loss mass on schema-critical label tokens helps preserve
fine-grained validity mappings. Length normalization most directly improves
textual grounding: Span F1 increases from 85.29 to 86.37 on PV Miner and from
84.98 to 85.63 on SciERC. Together, these trends support the intended
decomposition: class balancing addresses long-tailed ontology labels, token
weighting combats critical-token sparsity, and length normalization reduces
optimization bias from variable-length grounded completions.

\section{Prompt Engineered Instruction}
\label{app:Prompt_Engineered_Instruction}

This appendix provides the design rationale and failure-mode analysis for the
six prompt modules (M1--M6) introduced in Section~\ref{sec:prompt_engineering}. The main
text describes the template and its empirical effect; this appendix explains why
each module is included and how the modules interact. The template is designed
for, and can be used across, a broad class of ontology-constrained structured
prediction tasks in which the model must emit schema-valid outputs containing
semantic labels, grounded textual evidence, and, when required, typed relational
links.

\subsection{M1: Hierarchical XML Structuring}
\label{app:prompt-xml}

M1 organizes the instruction into semantically tagged blocks, including
\texttt{<role>}, \texttt{<performance\_target>}, \texttt{<task>}, and
\texttt{<structured\_prediction\_ontology>}. The model does not parse these tags
as formal XML; rather, the tags function as lightweight structural delimiters
that separate global behavioral instructions from local task-specific decision
rules.

This structure serves two purposes. First, it creates explicit scope boundaries
inside a long instruction. Structured prediction prompts often contain role
definitions, ontology inventories, disambiguation rules, metadata controls,
reasoning steps, and output schemas. Without segmentation, these constraints can
interfere with one another or become less salient as the context grows. The
tagged structure mitigates this by assigning each type of instruction a clear
location and semantic scope. Second, the structure supports modular
substitution: a new task can replace the ontology, metadata variables, grounding
rules, relational-link inventory, and output schema while preserving the same
instruction architecture. This improves portability across structured prediction
settings and reduces format drift in long-context generation
\cite{white2023prompt}.

\subsection{M2: Expert-Curated Disambiguation Rules}
\label{app:prompt-disambiguation}

M2 encodes expert-derived decision boundaries for cases that are likely to be
confused by an instruction-tuned model. These rules may distinguish confusable
semantic labels, specify textual-grounding boundaries, define when a candidate
Span is too broad or too narrow, or determine whether a relational link is valid.
For relational tasks, the rules can additionally specify constraints over the
structured relation fields, such as valid evidence
pairings, relation type, link directionality, scope, and
metadata-dependent validity conditions.

The purpose of this module is to operationalize the annotation manual inside the
prompt. Rather than relying on the model to infer boundary conditions from label
names alone, M2 states the decision criteria explicitly. This is important
because structured prediction evaluation is often sensitive to small local
errors: a near-correct label, an over-extended Span, an invalid relation
type, an incorrect evidence pairing, or an
incorrect sentence references in a coreference record can change the scored
output. Crisp disambiguation rules therefore reduce ontology-boundary,
grounding-boundary, and relational-linking errors
\cite{pang2023guideline,sainz2024gollie}.

\subsection{M3: Reasoning Scaffold as Structured Verification}
\label{app:prompt-cot}

M3 provides a task-specialized reasoning scaffold. The scaffold decomposes
prediction into context and metadata analysis, candidate-unit decomposition,
semantic label matching, textual grounding with boundary verification,
relational linking with structured-field validity verification, and final
cross-validation before output emission. Unlike generic chain-of-thought
prompting, the scaffold is not intended to elicit free-form explanation; it is a
verification routine aligned with the fields of the structured output.

This module addresses reasoning shortcuts and incomplete record construction.
A model may identify a plausible label without verifying the evidence Span, or
it may extract valid units without checking whether the relation type,
evidence, and associated sentence identifier fields are
jointly valid under the task schema. M3 forces these decisions to be considered
in sequence: first identify candidate units, then assign labels, then verify
grounding, then validate relational-link fields, and finally check consistency
across the full output. This makes the model less likely to emit records that
are syntactically valid but semantically or relationally inconsistent
\cite{wei2022chain}.

\definecolor{gainGreen}{HTML}{10B981} 
\definecolor{lossRed}{HTML}{F43F5E}   

\renewcommand{\poscell}[2]{\cellcolor{gainGreen!#1!white}\textbf{#2}}
\renewcommand{\negcell}[2]{\cellcolor{lossRed!#1!white}\textbf{#2}}

\begin{snwidetable}[t]
\centering
\scriptsize
\setlength{\tabcolsep}{2.5pt}
\renewcommand{\arraystretch}{1.08}
\caption{Zero-shot F1 results under baseline (Base) and prompt-engineered instructions (Prompt) on PV Miner and SciERC. F1 cells under Prompt are colored by the relative F1 change from baseline to prompt-engineered instruction.  Stronger color intensity indicates a larger relative change.}
\label{tab:zeroshot_prompt_baseline_side_by_side} 

\vspace{2pt}

\resizebox{\linewidth}{!}{%
\begin{tabular}[t]{@{}l cc cc cc@{}}
\toprule
\multicolumn{7}{c}{\textbf{(a) PV Miner}} \\
\midrule
\multirow{2}{*}{\textbf{Model}} 
& \multicolumn{2}{c}{\textbf{Code}} 
& \multicolumn{2}{c}{\textbf{Sub-code}} 
& \multicolumn{2}{c}{\textbf{Span}} \\
\cmidrule(lr){2-3}\cmidrule(lr){4-5}\cmidrule(lr){6-7}
& \textbf{Base} & \textbf{Prompt}
& \textbf{Base} & \textbf{Prompt}
& \textbf{Base} & \textbf{Prompt} \\
\midrule
Llama-3.3-70B   & 60.11 & \poscell{5}{62.25}   & 36.13 & \poscell{9}{43.71}   & 55.04 & \poscell{7}{60.86} \\
Llama-3.1-8B    & 0.00  & \poscell{35}{47.09}  & 0.00  & \poscell{35}{20.84}  & 55.24 & \negcell{5}{54.15} \\
Qwen2.5-7B      & 31.71 & \poscell{12}{42.87}  & 22.43 & \poscell{8}{25.78}   & 40.29 & \poscell{8}{46.72} \\
Qwen2.5-14B     & 56.60 & \poscell{6}{60.60}   & 28.80 & \poscell{11}{37.74}  & 44.68 & \poscell{6}{47.55} \\
Llama-3.2-3B    & 22.96 & \poscell{16}{38.24}  & 10.54 & \poscell{10}{13.22}  & 35.29 & \poscell{7}{38.98} \\
Qwen2.5-1.5B    & 19.20 & \poscell{9}{22.64}   & 1.95  & \poscell{32}{17.85}  & 16.25 & \poscell{11}{21.16} \\
\bottomrule
\end{tabular}%
\hspace{6pt}%
\begin{tabular}[t]{@{}l cc cc cc cc@{}}
\toprule
\multicolumn{9}{c}{\textbf{(b) SciERC}} \\
\midrule
\multirow{2}{*}{\textbf{Model}} 
& \multicolumn{2}{c}{\textbf{Entity}} 
& \multicolumn{2}{c}{\textbf{Relation}} 
& \multicolumn{2}{c}{\textbf{Span}} 
& \multicolumn{2}{c}{\textbf{Coreference}} \\
\cmidrule(lr){2-3}\cmidrule(lr){4-5}\cmidrule(lr){6-7}\cmidrule(lr){8-9}
& \textbf{Base} & \textbf{Prompt}
& \textbf{Base} & \textbf{Prompt}
& \textbf{Base} & \textbf{Prompt}
& \textbf{Base} & \textbf{Prompt} \\
\midrule
Llama-3.3-70B   & 78.59 & \poscell{5}{81.08}   & 14.82 & \poscell{7}{16.13}   & 59.09 & \poscell{7}{65.10}   & 0.29 & \poscell{30}{1.72} \\
Llama-3.1-8B    & 73.95 & \poscell{4}{74.62}   & 5.64  & \poscell{18}{10.09}  & 53.72 & \poscell{8}{60.71}   & 0.00 & \poscell{35}{0.44} \\
Qwen2.5-7B      & 70.23 & \poscell{7}{78.68}   & 10.90 & \poscell{10}{13.33}  & 58.17 & \poscell{8}{66.60}   & 0.88 & \poscell{8}{1.01} \\
Qwen2.5-14B     & 78.83 & \poscell{4}{79.29}   & 12.86 & \poscell{14}{18.82}  & 64.44 & \poscell{6}{68.01}   & 5.08 & \poscell{11}{6.55} \\
Llama-3.2-3B    & 27.04 & \poscell{24}{74.16}  & 0.88  & \poscell{27}{3.27}   & 31.44 & \poscell{14}{45.84}  & 0.00 & \poscell{35}{0.09} \\
Qwen2.5-1.5B    & 62.21 & \poscell{7}{68.17}   & 4.35  & \poscell{10}{5.39}   & 29.98 & \poscell{13}{41.56}  & 0.07 & \poscell{28}{0.31} \\
\bottomrule
\end{tabular}%
}
\end{snwidetable}

\subsection{M4: Metadata-Aware Decision Logic}
\label{app:prompt-metadata}

M4 exposes task metadata as explicit control variables. Depending on the task,
these variables may include speaker role, message direction, sentence
identifier, document identifier, section type, temporal context, modality, or
other task-specific controls. The key design choice is to make metadata visible
to the model as a constraint rather than leaving it as implicit context.

This module narrows the valid search space for structured prediction. Metadata
can determine which labels are admissible, which Spans are in scope, which
sentence identifier values are valid, and which relational records are possible
under the declared schema. By converting latent contextual attributes into
observed control signals, M4 reduces metadata-conditioned confusion and prevents
plausible but invalid predictions. Its advantage is especially clear in tasks
where two records may look lexically similar but differ because of speaker role,
sentence context, document section, or relational structure
\cite{kong2024better,ouyang2022training}.

\subsection{M5: Structured Output Schema Contract}
\label{app:prompt-schema}

M5 specifies the output as a machine-parseable schema contract. The contract
defines the required record fields, the representation of grounded evidence,
semantic-label validity constraints, relational-link validity constraints, and
task-specific coverage requirements. Depending on the task, the schema may
specify fields for semantic labels, grounded evidence units, record identifiers,
typed links between records or spans, sentence- or document-level provenance,
and any additional task-specific attributes required for valid structured
prediction.

The schema contract has both modeling and evaluation advantages. At inference
time, it reduces format drift by giving the model a fixed structure to fill
rather than an unconstrained response format. At evaluation time, it enables
deterministic parsing and reproducible metric computation. M5 therefore protects
the interface between model generation and automatic evaluation
\cite{li2024simple,sainz2024gollie}.

\subsection{M6: Single-Turn Self-Validation Quality Gate}
\label{app:prompt-selfval}

M6 adds a compact validation checklist immediately before final output emission.
The checklist verifies that the output is parseable under the declared schema,
that semantic labels are drawn from the allowed ontology, that grounded evidence
satisfies the task requirements, that relational records satisfy the required constraints, that relevant disambiguation rules have been applied, and that the
final output is defensible under expert review.

This module is an intra-generation audit rather than an iterative
self-refinement procedure. The model is not asked to perform multi-turn
critique-and-revision; instead, it is asked to check concrete validity
conditions before emitting the final answer. This distinction is important
because single-pass inference is more efficient and more reproducible, and
because self-correction without external feedback can be unreliable on
reasoning-intensive tasks. M6 therefore functions as a pre-submission audit: it
does not add new information, but it increases the likelihood that the generated
structure satisfies the constraints already declared in M1--M5
\cite{madaan2023selfrefine,huang2024cannot}.

\subsection{Interactions Between Modules}
\label{app:prompt-interactions}

The six modules are designed to be individually interpretable but jointly
effective. M1 provides the structural anchors that make the other modules easier
to localize and apply. M2 supplies expert boundary conditions that are invoked
during the M3 reasoning scaffold. M4 narrows the search space before label,
grounding, or relational-link decisions are made. M5 converts those decisions
into a strictly parseable schema. M6 then audits the final output against the
preceding constraints.

The modules therefore address complementary failure modes. M1 and M5 reduce
format drift and schema violations. M2 and M4 reduce ontology ambiguity,
grounding-boundary errors, and metadata-conditioned confusion. M3 reduces
reasoning shortcuts by separating semantic labelling, textual grounding, and
relational-link validation into explicit verification steps. M6 reduces
verification omissions by requiring a final single-turn consistency check. The
cross-module advantage is that the prompt does not rely on any single mechanism:
structural organization, expert rules, metadata control, stepwise reasoning,
schema enforcement, and self-validation work together to constrain the model
toward valid structured outputs.

\subsection{Prompt Engineering Improves Zero-Shot Structured Prediction}

Table~\ref{tab:zeroshot_prompt_baseline_side_by_side} shows that the
prompt-engineered instruction substantially improves zero-shot structured 
prediction before any supervised or preference-based fine-tuning. On PV Miner,
the mean F1 across all model--metric cells increases from 29.85 to 39.01,
corresponding to a +9.17 F1 gain and a +30.72\% relative macro improvement.
Prompt engineering improves 17 of 18 PV Miner cells, with the largest average
gains on ontology-level labels: +13.85 Code F1 and +9.88 Sub-code F1, compared
with +3.77 Span F1.

The same trend holds on SciERC. The mean F1 increases from 30.98 to 36.71,
corresponding to a +5.73 F1 gain and a +18.50\% relative macro improvement.
Prompt engineering improves all 24 SciERC model--metric cells. The largest
average gains occur for Entity F1 (+10.86) and Span F1 (+8.50), while Relation
F1 improves by +2.93 and Coreference F1 improves by +0.63. These results show
that structured prompting is an important initialization step: it improves schema
adherence and task decomposition, but it does not by itself solve the harder
post-SFT preference-optimization problem addressed by TAB-PO.

\subsection{Confusion-Matrix Evidence for Prompt-Engineered Instruction}
\label{app:prompt-confusion-analysis}

Table~\ref{tab:zeroshot_prompt_baseline_side_by_side} quantifies the performance
improvements produced by the prompt-engineered instruction over the baseline
instruction. Importantly, the baseline is not an uninformed prompt: it contains
the essential task definition, semantic-label inventory, grounding requirements,
relational-linking requirements where applicable, and input/output schema needed
for structured prediction. The gains therefore do not simply come from adding
missing task information. Instead, they support the value of organizing the same
task knowledge into explicit modules for ontology salience, disambiguation,
metadata-aware control, stepwise verification, schema enforcement, and
self-validation.

\begin{figure}[!t]
    \centering
    \begin{subfigure}[t]{0.45\linewidth}
        \centering
        \includegraphics[width=\linewidth]{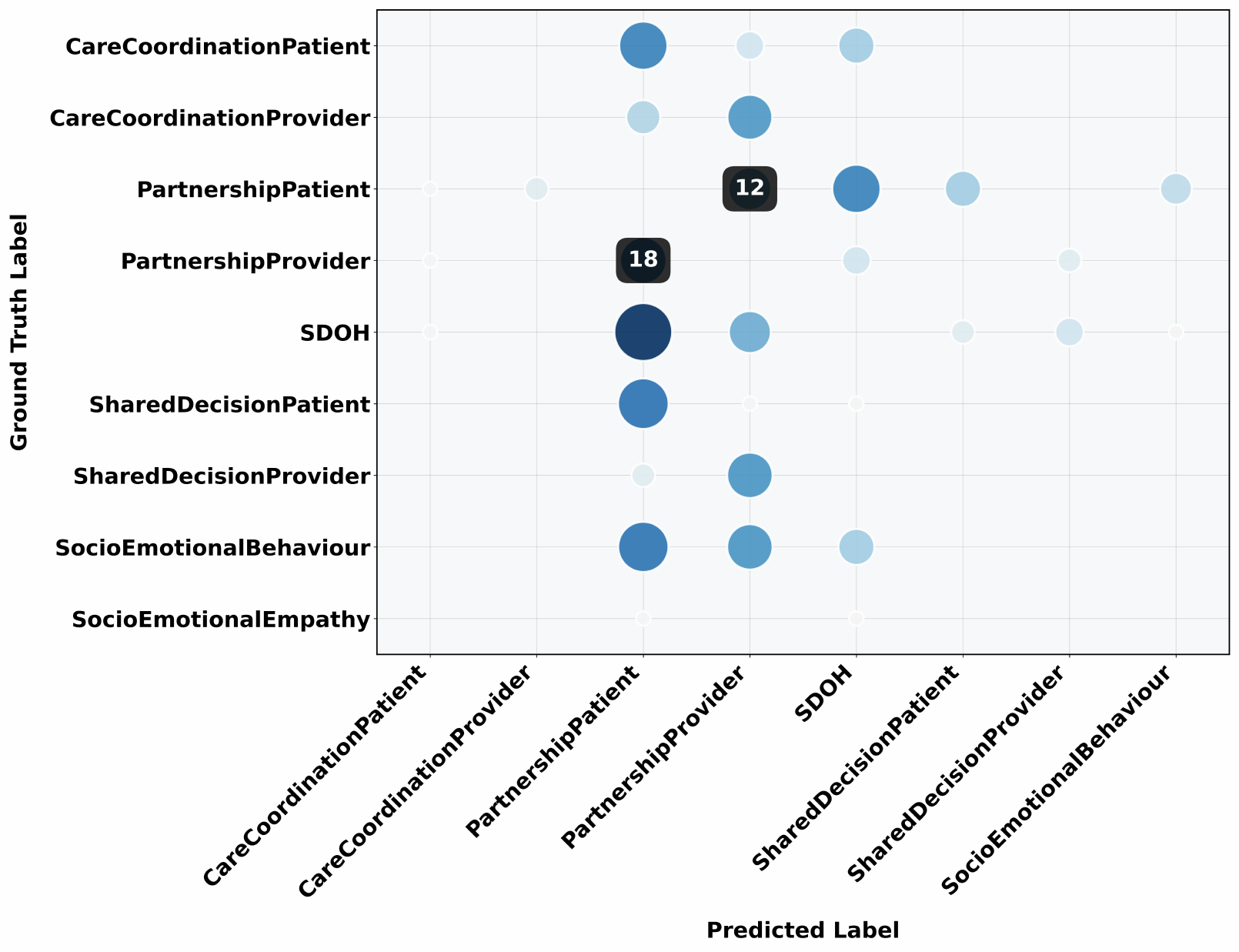}
    \end{subfigure}
    \hfill
    \begin{subfigure}[t]{0.45\linewidth}
        \centering
        \includegraphics[width=\linewidth]{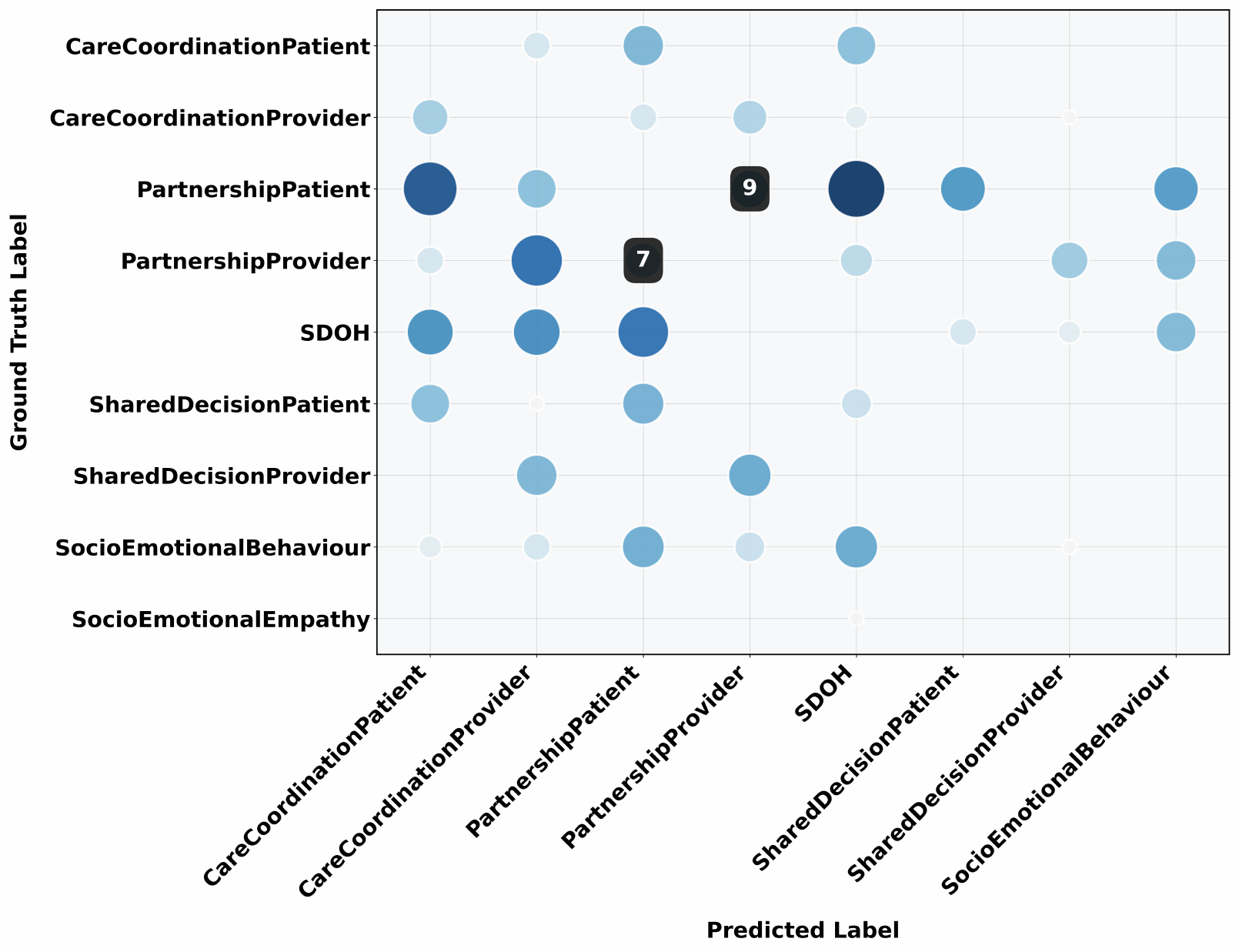}
    \end{subfigure}

    \caption{Code-level confusion matrices comparing baseline instruction outputs
    (left) and prompt-engineered instruction outputs (right) from
    Llama-3.2-3B-Instruct.}
    \label{fig:code-comparison}
\end{figure}

\begin{figure}[!t]
    \centering
    \begin{subfigure}[t]{0.45\linewidth}
        \centering
        \includegraphics[width=\linewidth]{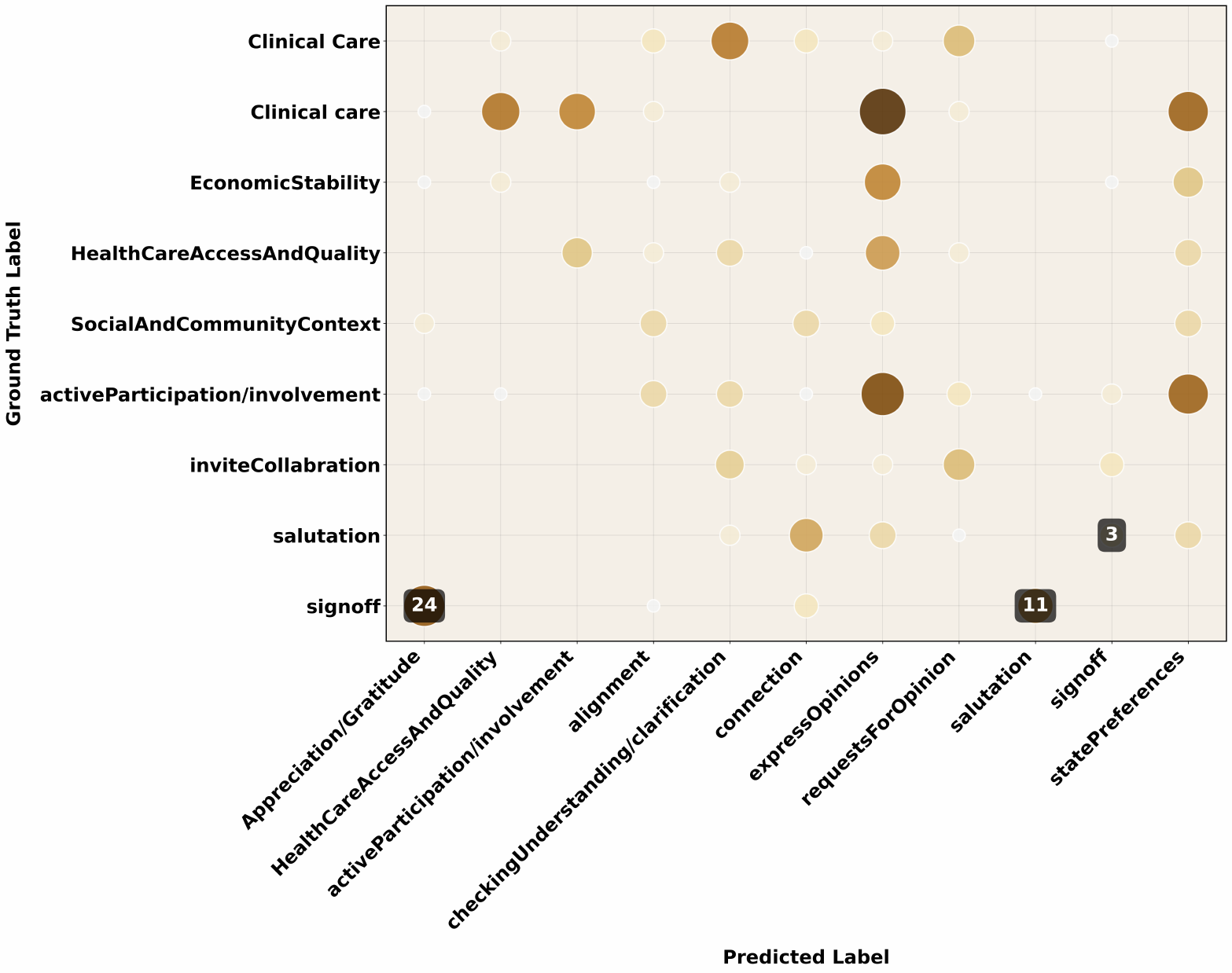}
    \end{subfigure}
    \hfill
    \begin{subfigure}[t]{0.45\linewidth}
        \centering
        \includegraphics[width=\linewidth]{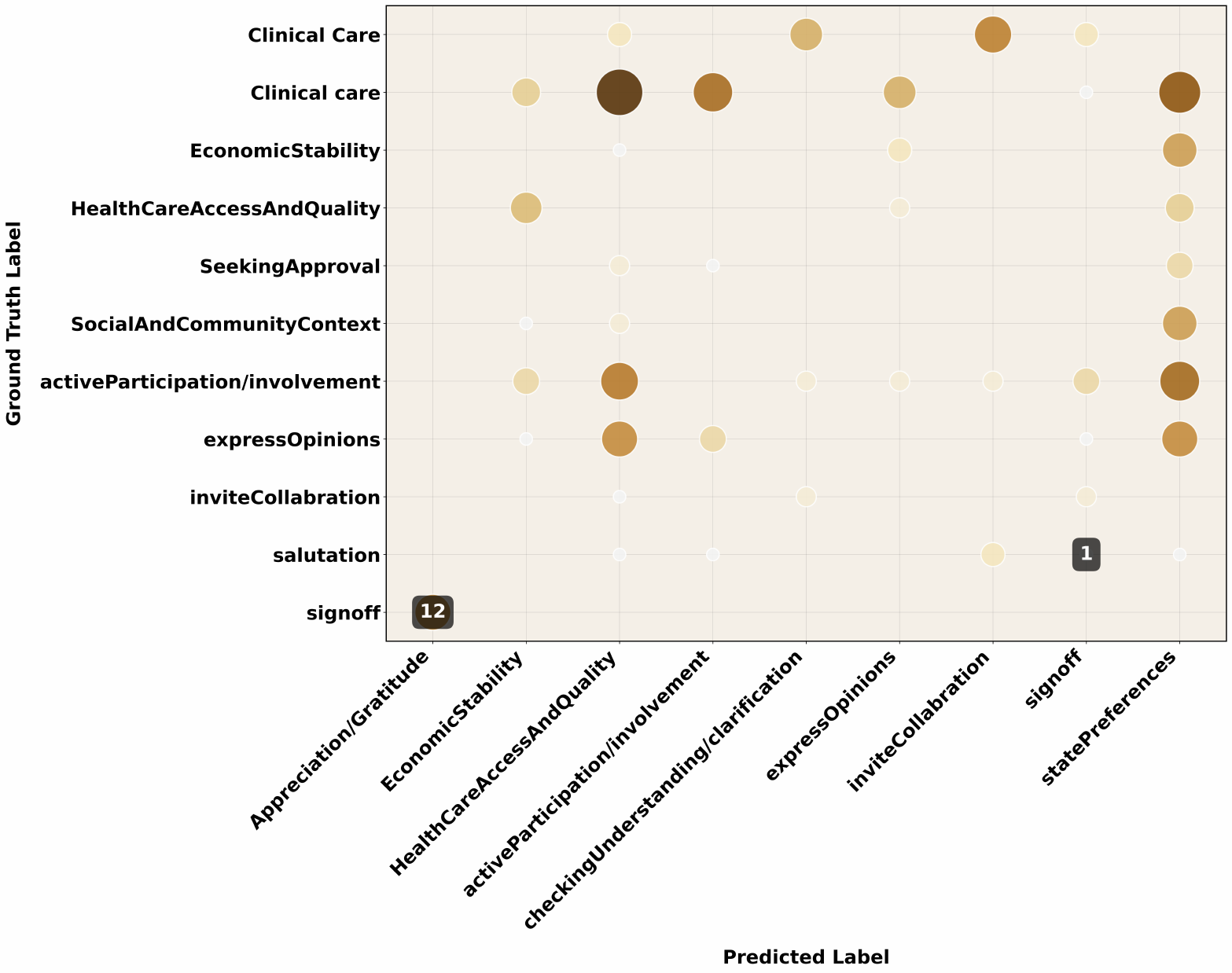}
    \end{subfigure}

    \caption{Sub-code-level confusion matrices comparing baseline instruction
    outputs (left) and prompt-engineered instruction outputs (right) from
    Llama-3.2-3B-Instruct. Bubble size indicates the mismatch count between
    ground-truth and predicted Sub-codes. Only the top 10 most frequently
    confused Sub-codes are shown for clarity.}
    \label{fig:subcode-comparison}
\end{figure}

\begin{figure}[!t]
  \centering
  \includegraphics[width=0.9\linewidth]{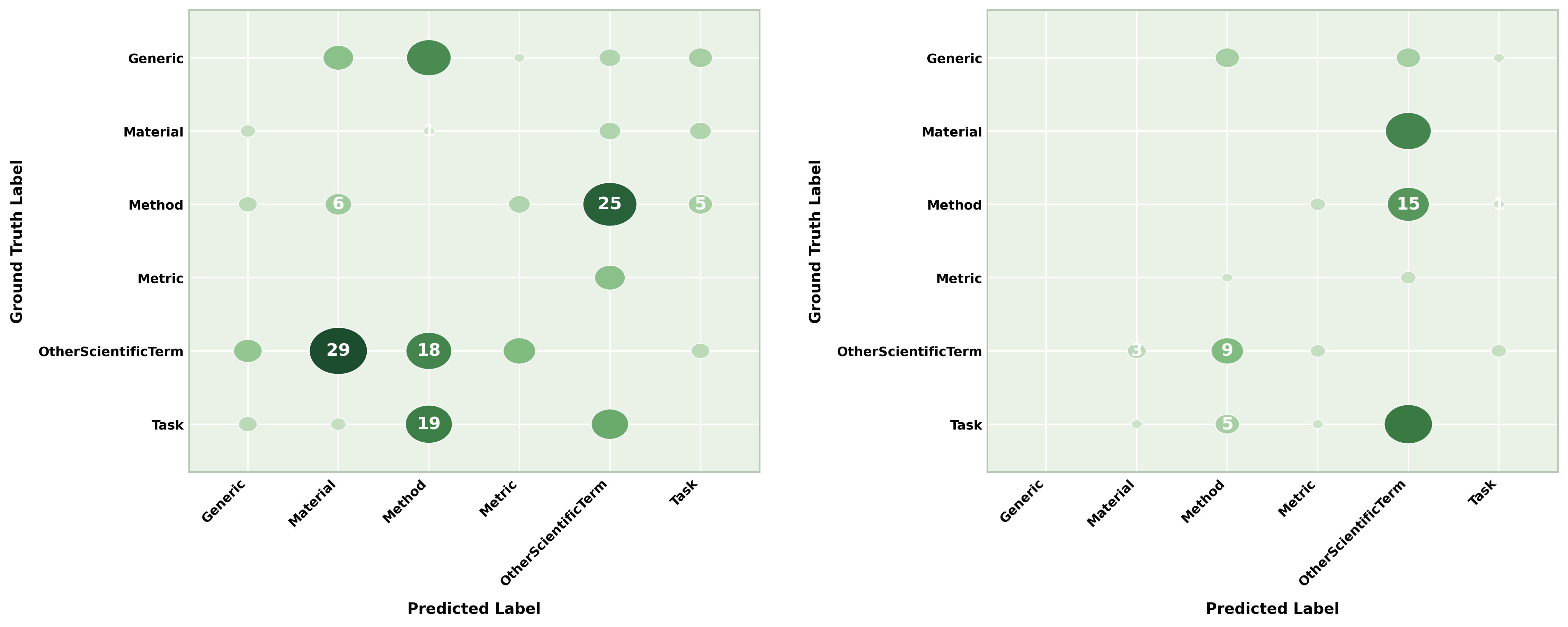}
  \caption{Entity type-level confusion matrices comparing baseline instruction outputs
    (left) and prompt-engineered instruction outputs (right) from
    Llama-3.2-3B-Instruct.}
  \label{fig:entity-comparison}
\end{figure}

The confusion matrices in
Figures~\ref{fig:code-comparison},
\ref{fig:subcode-comparison}, and
\ref{fig:entity-comparison} explain where these gains arise. At the Code level, prompt engineering reduces direction-conditioned confusions
between patient- and provider-oriented categories. For example, confusions
between \texttt{PartnershipPatient} and \texttt{PartnershipProvider} decrease
after prompt engineering. This pattern supports the role of M4, which exposes
metadata and control variables as explicit constraints that narrow the valid
label search space, and M6, which checks whether the final prediction is
consistent with those constraints.

At the Sub-code level, prompt engineering reduces common boundary errors among
semantically adjacent communication functions. The baseline shows prominent
confusions such as \texttt{signoff} $\rightarrow$ \texttt{Appreciation/Gratitude} and \texttt{salutation} $\rightarrow$ \texttt{signoff}; after prompt engineering, these confusions decrease. These
reductions are consistent with M2, which encodes expert-curated decision
boundaries for confusable labels, and M3, which forces candidate units to be
decomposed, labelled, grounded, and cross-checked before output emission.

The SciERC entity-level confusion matrices show a similar effect for scientific
information extraction. Prompt engineering reduces several high-frequency
off-diagonal confusions, including
\texttt{OtherScientificTerm} $\rightarrow$ \texttt{Material},
\texttt{OtherScientificTerm} $\rightarrow$ \texttt{Method},
\texttt{Method} $\rightarrow$ \texttt{OtherScientificTerm}, and
\texttt{Task} $\rightarrow$ \texttt{Method}. These improvements suggest that the
prompt modules help the model separate ontology-near entity types whose surface
forms are often lexically ambiguous. They also support the combined benefit of M1, M2, M3, M5, and M6: structural segmentation improves ontology salience,
expert rules sharpen local decision boundaries, the reasoning scaffold separates
candidate identification from label assignment, the schema contract enforces
valid structured output, and the quality gate encourages final constraint
checking.

\definecolor{tplFrame}{HTML}{4A148C}
\definecolor{tplBack}{HTML}{F3E5F5}
\definecolor{tplTitle}{HTML}{FFFFFF}
\definecolor{tplTitleBg}{HTML}{6A1B9A}
\definecolor{modXML}{HTML}{1565C0}
\definecolor{modDIS}{HTML}{7B1FA2}
\definecolor{modCOT}{HTML}{E65100}
\definecolor{modVAL}{HTML}{C62828}
\definecolor{modLOG}{HTML}{00838F}
\definecolor{modSCH}{HTML}{2E7D32}
\definecolor{slotGray}{HTML}{616161}

\newcommand{\purposehighlight}[2]{%
  \vspace{2pt}
  \noindent\colorbox{#1!12}{%
    \parbox{\dimexpr\linewidth-2\fboxsep\relax}{%
      \rmfamily\footnotesize\textcolor{slotGray}{\textit{#2}}%
    }%
  }%
}

\begin{figure}[!htbp]
\centering

\begin{adjustbox}{max totalsize={0.96\linewidth}{0.96\textheight},center}
\begin{minipage}{0.96\linewidth}
\begin{tcolorbox}[
  enhanced,
  colframe=tplFrame,
  colback=tplBack,
  colbacktitle=tplTitleBg,
  coltitle=tplTitle,
  title={\small\bfseries Prompt-Engineered Instruction Template for Structured Prediction},
  fonttitle=\bfseries\large,
  boxrule=1.2pt,
  arc=3.5mm,
  top=4pt, bottom=4pt, left=8pt, right=8pt,
  drop fuzzy shadow=black!35,
]

\small

\vspace{-4pt}

\begin{tcolorbox}[
  enhanced,
  colback=white,
  colframe=modXML!60,
  boxrule=0.8pt,
  arc=2mm,
  left=6pt, right=6pt, top=4pt, bottom=3pt,
  title={\textcolor{modXML}{\textbf{M1\;\;$\vert$\;\;Hierarchical XML Structuring}}},
  fonttitle=\small\bfseries,
  colbacktitle=modXML!10,
  coltitle=modXML, 
]
\footnotesize\ttfamily
\textcolor{modXML}{<role>} \textcolor{slotGray}{\textit{[Domain expert persona and behavioral prior]}} \textcolor{modXML}{</role>}\\[1pt]
\textcolor{modXML}{<performance\_target>} \textcolor{slotGray}{\textit{[Precision-oriented accuracy expectations for semantic labelling, textual grounding, relational linking, and schema-valid structured prediction]}} \textcolor{modXML}{</performance\_target>}\\[1pt]
\textcolor{modXML}{<task>} \textcolor{slotGray}{\textit{[Task definition, input/output specification, and structured prediction ontology constraints]}} \textcolor{modXML}{</task>}\\[1pt]
\textcolor{modXML}{<structured\_prediction\_ontology>} \textcolor{slotGray}{\textit{[Authoritative ontology with operational definitions for semantic labels, textual grounding units, and relational link types/validity conditions, where applicable]}} \textcolor{modXML}{</structured\_prediction\_ontology>}\\[2pt]
\purposehighlight{modXML}{Separates global instructions from local decision rules; improves constraint salience in long contexts.}
\end{tcolorbox}

\vspace{-8pt}

\begin{tcolorbox}[
  enhanced,
  colback=white,
  colframe=modDIS!60,
  boxrule=0.8pt,
  arc=2mm,
  left=6pt, right=6pt, top=4pt, bottom=3pt,
  title={\textcolor{modDIS}{\textbf{M2\;\;$\vert$\;\;Expert-Curated Disambiguation Rules}}},
  fonttitle=\small\bfseries,
  colbacktitle=modDIS!10,
  coltitle=modDIS,
]
\footnotesize\ttfamily
\textcolor{modDIS}{<disambiguation\_rules>}\\
\quad \textcolor{slotGray}{\textit{[Boundary condition 1: decision criteria for confusable semantic label pair A vs.\ B]}}\\
\quad \textcolor{slotGray}{\textit{[Boundary condition 2: textual grounding boundary, positional, specificity, or semantic cues]}}\\
\quad \textcolor{slotGray}{\textit{[Boundary condition 3: relational linking validity, source/target role, directionality, or scope conditions]}}\\
\quad \textcolor{slotGray}{\quad\dots}\\
\quad \textcolor{slotGray}{\textit{[Boundary condition K: metadata-dependent or context-dependent distinctions]}}\\
\textcolor{modDIS}{</disambiguation\_rules>}\\[2pt]
\purposehighlight{modDIS}{Operationalizes the annotation manual inside the prompt; resolves ontology boundary errors using expert-curated ambiguity adjudication from the annotation process.}
\end{tcolorbox}

\vspace{-8pt}

\begin{tcolorbox}[
  enhanced,
  colback=white,
  colframe=modCOT!60,
  boxrule=0.8pt,
  arc=2mm,
  left=6pt, right=6pt, top=4pt, bottom=3pt,
  title={\textcolor{modCOT}{\textbf{M3\;\;$\vert$\;\;Chain-of-Thought Reasoning Scaffold}}},
  fonttitle=\small\bfseries,
  colbacktitle=modCOT!10,
  coltitle=modCOT,
]
\footnotesize\ttfamily
\textcolor{modCOT}{<reasoning\_process>}\\
\quad \textcolor{slotGray}{\textit{Step 1: Context and metadata analysis}}\\
\quad \textcolor{slotGray}{\textit{Step 2: Candidate unit decomposition and semantic label matching}}\\
\quad \textcolor{slotGray}{\textit{Step 3: Textual grounding with boundary verification}}\\
\quad \textcolor{slotGray}{\textit{Step 4: Relational linking with argument-validity verification}}\\
\quad \textcolor{slotGray}{\textit{Step 5: Cross-validation with loop-back conditions}}\\
\textcolor{modCOT}{</reasoning\_process>}\\[2pt]
\purposehighlight{modCOT}{Imposes a disciplined internal verification sequence before structured output emission.}
\end{tcolorbox}

\vspace{-8pt}

\begin{tcolorbox}[
  enhanced,
  colback=white,
  colframe=modLOG!60,
  boxrule=0.8pt,
  arc=2mm,
  left=6pt, right=6pt, top=4pt, bottom=3pt,
  title={\textcolor{modLOG}{\textbf{M4\;\;$\vert$\;\;Metadata-Aware Decision Logic}}},
  fonttitle=\small\bfseries,
  colbacktitle=modLOG!10,
  coltitle=modLOG,
]
\footnotesize\ttfamily
\textcolor{modLOG}{<metadata\_control>}\\
\quad \textcolor{slotGray}{\textit{[Expose task metadata as explicit control variables, e.g., speaker role, message direction, sentence identifier, document identifier, section type, temporal context, source/target role, modality, and task-specific control fields]}}\\
\textcolor{modLOG}{</metadata\_control>}\\[2pt]
\purposehighlight{modLOG}{Converts metadata values, which are latent contextual attributes, into observed control signals; narrows the valid search space for semantic labelling, textual grounding, and relational linking.}
\end{tcolorbox}

\vspace{-8pt}

\begin{tcolorbox}[
  enhanced,
  colback=white,
  colframe=modSCH!60,
  boxrule=0.8pt,
  arc=2mm,
  left=6pt, right=6pt, top=4pt, bottom=3pt,
  title={\textcolor{modSCH}{\textbf{M5\;\;$\vert$\;\;Structured Output Schema Contract}}},
  fonttitle=\small\bfseries,
  colbacktitle=modSCH!10,
  coltitle=modSCH,
]
\footnotesize\ttfamily
\textcolor{modSCH}{<output\_format>}\\
\quad \textcolor{slotGray}{\textit{[Machine-parseable hard constraints: schema for structured predictions, grounded evidence, semantic label validity, relational linking validity, and task-specific coverage conditions]}}\\
\textcolor{modSCH}{</output\_format>}\\[2pt]
\purposehighlight{modSCH}{Shifts the task from open-ended generation to structured generation with explicit validity conditions.}
\end{tcolorbox}

\vspace{-8pt}

\begin{tcolorbox}[
  enhanced,
  colback=white,
  colframe=modVAL!60,
  boxrule=0.8pt,
  arc=2mm,
  left=6pt, right=6pt, top=4pt, bottom=3pt,
  title={\textcolor{modVAL}{\textbf{M6\;\;$\vert$\;\;Self-Validation Quality Gate}}},
  fonttitle=\small\bfseries,
  colbacktitle=modVAL!10,
  coltitle=modVAL,
]
\footnotesize\ttfamily
\textcolor{modVAL}{<quality\_gate>}\\
\quad \textcolor{slotGray}{\textit{[1. Output is parseable under the declared schema]}}\\
\quad \textcolor{slotGray}{\textit{[2. Every semantic label is drawn from the allowed ontology]}}\\
\quad \textcolor{slotGray}{\textit{[3. Every relational link satisfies source/target argument-type constraints]}}\\
\quad \textcolor{slotGray}{\quad\dots}\\
\quad \textcolor{slotGray}{\textit{[K-1. All disambiguation rules have been applied]}}\\
\quad \textcolor{slotGray}{\textit{[K. The final output is high-confidence and defensible under expert review]}}\\
\textcolor{modVAL}{</quality\_gate>}\\[2pt]
\purposehighlight{modVAL}{Single-turn intra-generation audit; constrains the model to verify concrete failure conditions within one pass.}
\end{tcolorbox}

\vspace{-8pt}
\end{tcolorbox}
\end{minipage}
\end{adjustbox}

\caption{General modular prompt template for ontology-constrained structured prediction with semantic labelling, textual grounding, and relational linking.}
\label{fig:prompt_template}

\end{figure}

\clearpage

\clearpage

\section{Task-Specific Instantiations of the Structured Prediction Formulation}
\label{app:task_specific_formulations}

This appendix instantiates the general ontology-driven structured prediction
formulation introduced in Section~\ref{sec:general_task_formulation} for the
two tasks studied in this work.

\subsection{PV-Miner}
\label{subsec:pv_miner_formulation}

PV-Miner requires the model to extract communication-pattern records from a
patient--provider message. The objective is to predict one or more records,
where each record contains a coarse-grained Code, a valid fine-grained
Sub-code under that Code, and a text Span that grounds the prediction in the
source message. Therefore, PV-Miner instantiates the general formulation as
hierarchical semantic classification with direct textual grounding and no
inter-record relations.

Let $d \in \{\texttt{Y},\texttt{N}\}$ denote the message-direction metadata,
where $d=\texttt{Y}$ denotes provider$\rightarrow$patient and
$d=\texttt{N}$ denotes patient$\rightarrow$provider. Let
$L_{\text{Code}}$ be the set of valid Codes and $L_{\text{Sub-code}}$ be the
set of valid Sub-codes. The PV-Miner ontology defines a hierarchy
\begin{equation}
    \mathcal{H}_{\text{PV}}:
    L_{\text{Code}} \rightarrow 2^{L_{\text{Sub-code}}},
\end{equation}
where $\mathcal{H}_{\text{PV}}(\kappa)$ gives the set of Sub-codes valid under
Code $\kappa$.

The model maps
\begin{equation}
    f_\theta : (s, d, \mathcal{O}_{\text{PV}})
    \;\longmapsto\;
    \widehat{Y}_{\text{PV}}.
\end{equation}
\textbf{Record Instantiation.}
Each PV-Miner record is instantiated as
\begin{equation}
    y_i^{\text{PV}}
    =
    \big(
    (\kappa_i, \varsigma_i),
    \{\pi_i\},
    \emptyset
    \big),
\end{equation}
where
\begin{equation}
    \kappa_i \in L_{\text{Code}},
    \qquad
    \varsigma_i \in \mathcal{H}_{\text{PV}}(\kappa_i),
    \qquad
    \pi_i \in \mathcal{S}(s).
\end{equation}

Thus,
\begin{equation} \begin{aligned} \widehat{Y}_{\text{PV}} = \Big\{ \big( (\kappa, \varsigma), \{\pi\}, \emptyset \big) \;\Big|\; \kappa \in L_{\text{Code}}, \varsigma \in \mathcal{H}_{\text{PV}}(\kappa), \pi \in \mathcal{S}(s) \Big\}. \end{aligned} \end{equation}

Here, $\ell_i = (\kappa_i,\varsigma_i)$ is the ordered hierarchical label
tuple, $\Pi_i=\{\pi_i\}$ is the grounding span, and $R_i=\emptyset$ because
PV-Miner records do not point to other records.

\subsection{SciERC}
\label{subsec:scierc_formulation}

SciERC requires the model to extract a relational information graph from a
scientific document. The objective is to predict typed entity records grounded
in text, typed relation records connecting entity records, and coreference
cluster records grouping entity records that refer to the same underlying
scientific concept. Thus, SciERC instantiates the general formulation as a
record graph with grounded entity records and structural relation/coreference
records. The model maps
\begin{equation}
    f_\theta : (s, \mathcal{O}_{\text{SciERC}})
    \;\longmapsto\;
    \widehat{\mathcal{Y}}_{\text{SciERC}}.
\end{equation}

The predicted SciERC output is partitioned into three record subsets:
\begin{equation}
    \widehat{Y}_{\text{SciERC}}
    =
    \mathcal{E}
    \;\dot{\cup}\;
    \mathcal{R}
    \;\dot{\cup}\;
    \mathcal{C},
\end{equation}
where $\mathcal{E}$ is the set of entity records, $\mathcal{R}$ is the set of
relation records, and $\mathcal{C}$ is the set of coreference records. Each
record still follows the same general format $y_i=(\ell_i,\Pi_i,R_i)$.

\textbf{Record Instantiation.}
Let $L_{\text{Entity}}$ denote the SciERC entity-type inventory,
$L_{\text{Relation}}$ denote the relation-type inventory, and $\mathcal{D}(s)$
denote the set of sentence identifiers in the source document. The predicted
SciERC output is partitioned into entity, relation, and coreference records:
\begin{equation}
    \widehat{Y}_{\text{SciERC}}
    =
    \mathcal{E}
    \;\dot{\cup}\;
    \mathcal{R}
    \;\dot{\cup}\;
    \mathcal{C}.
\end{equation}

Entity records are typed mentions grounded in the source text:
\begin{equation}
    e
    =
    \big(
    t_e,
    \{\pi_e\},
    \emptyset
    \big),
    \qquad
    t_e \in L_{\text{Entity}},
    \quad
    \pi_e \in \mathcal{S}(s).
\end{equation}
Here, $t_e$ is the entity type and $\pi_e$ is the text span that grounds the
entity mention.

Relation records are typed directed links from one grounded entity mention to
another:

\begin{equation}
\begin{aligned}
r
=
\big(
\{\pi_{r1}\},
t_r,
\{\pi_{r2}\}
\big),
t_r
\in L_{\text{Relation}},
\pi_{r1},\pi_{r2}
\in \mathcal{S}(s).
\end{aligned}
\end{equation}

Here, $t_r$ is the relation type, and the ordered pair of grounding sets
$(\{\pi_{r1}\},\{\pi_{r2}\})$ specifies the directed relation from the first
grounded entity mention to the second grounded entity mention.

Coreference records link grounded entity mentions that refer to the same
underlying scientific concept:

\begin{equation}
\begin{aligned}
c
=
\big(
\{\pi_{c1}\},
d_1,
\{\pi_{c2}\},
d_2
\big),
\pi_{c1},\pi_{c2}
\in \mathcal{S}(s),
d_1,d_2
\in \mathcal{I}_{\mathrm{sent}}(s).
\end{aligned}
\end{equation}

Here, $\{\pi_{c1}\}$ and $\{\pi_{c2}\}$ are the grounding sets for the two
linked mentions, and $d_1$ and $d_2$ are their corresponding sentence
identifiers.

\begingroup
\setlength{\abovedisplayskip}{3pt}
\setlength{\belowdisplayskip}{3pt}
\setlength{\abovedisplayshortskip}{2pt}
\setlength{\belowdisplayshortskip}{2pt}
\setlength{\jot}{2pt}

\section{Metric Definitions}
\label{append:metric}

Each task adopts an evaluation strategy tailored to its prediction format and
semantic structure. This appendix defines the task-specific metrics used for
PV-Miner and SciERC. PV-Miner is evaluated through hierarchical semantic-label
prediction and textual grounding, using Code F1, Sub-code F1, and Span F1.
SciERC is evaluated through scientific information extraction, using Entity F1,
Relation F1, Span F1, and Coreference F1 to measure semantic labelling,
textual grounding, and relational linking.

\begin{figure}[!h]
  \centering
  \includegraphics[width=0.95\linewidth]{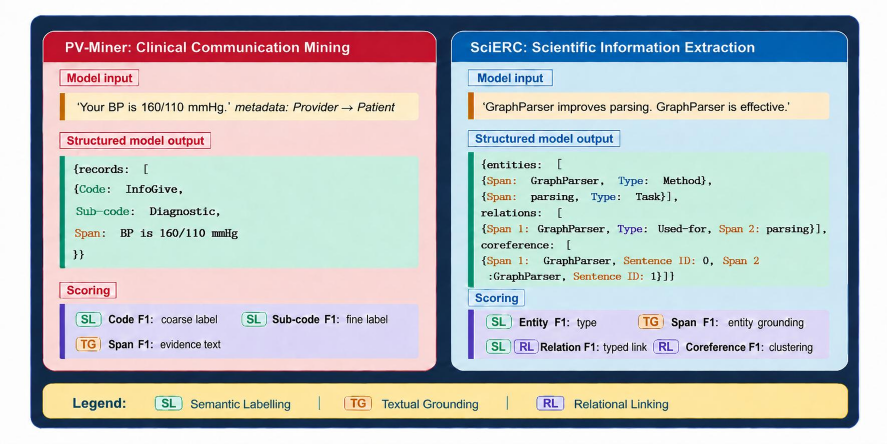}
  \caption{  Examples of the two ontology-constrained structured prediction
  tasks evaluated in this paper.
  \textbf{PV Miner} requires hierarchical clinical-communication labels
  with grounded evidence spans;
  \textbf{SciERC} requires entity typing, span grounding, relation
  extraction, and coreference linking..}
  \label{fig:tabpo-pipeline-1}
\end{figure}

\subsection{PV-Miner}

\paragraph{Code Classification}

The Code classification task is formulated as a multi-label classification problem over a predefined set of communicative Codes. Let $\widehat{\mathcal K}^{(i)}$ denote the predicted Code set and $\mathcal K^{(i)}$ the gold Code set for instance $i$. We compute precision recall, and F1-score as follows:

\begin{equation}
\begin{aligned}
\mathrm{precision}_{\mathrm{Code}}
&=
\frac{
\sum_i \left|
\widehat{\mathcal K}^{(i)}
\cap
\mathcal K^{(i)}
\right|
}{
\sum_i \left|
\widehat{\mathcal K}^{(i)}
\right|
}
\\[1.0em]
\mathrm{recall}_{\mathrm{Code}}
&=
\frac{
\sum_i \left|
\widehat{\mathcal K}^{(i)}
\cap
\mathcal K^{(i)}
\right|
}{
\sum_i \left|
\mathcal K^{(i)}
\right|
}
\\[1.0em]
F1_{\mathrm{Code}}
&=
\frac{
2 \times \mathrm{precision}_{\mathrm{Code}} \times \mathrm{recall}_{\mathrm{Code}}
}{
\mathrm{precision}_{\mathrm{Code}} + \mathrm{recall}_{\mathrm{Code}}
}.
\end{aligned}
\end{equation}

\paragraph{Sub-code Classification}

Sub-code classification is also evaluated as a multi-label task, where each message may be annotated with one or more Sub-codes tied to a parent Code. Let $\widehat{\mathcal V}^{(i)}$ and $\mathcal V^{(i)}$ denote predicted and gold Sub-code sets, respectively. Metrics are calculated using:

\begin{equation}
\begin{gathered}
\mathrm{precision}_{\mathrm{Sub\text{-}code}}
=
\dfrac{
\sum_i \left|
\widehat{\mathcal V}^{(i)}
\cap
\mathcal V^{(i)}
\right|
}{
\sum_i \left|
\widehat{\mathcal V}^{(i)}
\right|
}
\\[1.4em]
\mathrm{recall}_{\mathrm{Sub\text{-}code}}
=
\dfrac{
\sum_i \left|
\widehat{\mathcal V}^{(i)}
\cap
\mathcal V^{(i)}
\right|
}{
\sum_i \left|
\mathcal V^{(i)}
\right|
}
\\[1.4em]
F1_{\mathrm{Sub\text{-}code}}
=
\dfrac{
2\,\mathrm{precision}_{\mathrm{Sub\text{-}code}}\,
\mathrm{recall}_{\mathrm{Sub\text{-}code}}
}{
\mathrm{precision}_{\mathrm{Sub\text{-}code}}
+
\mathrm{recall}_{\mathrm{Sub\text{-}code}}
}.
\end{gathered}
\end{equation}
\paragraph{Span}

For Span extraction, we evaluate each predicted evidence string against gold Spans using a \textbf{relaxed token-level matching strategy}, which combines:
\begin{itemize}
    \setlength{\itemsep}{0pt}
    \setlength{\topsep}{2pt}
    \setlength{\parsep}{0pt}
    \item Full containment (i.e., gold Span is fully included in predicted Span or vice versa).
    \item Jaccard similarity (between predicted and gold Spans), with a threshold of 0.6.
\end{itemize}

Let $\widehat{\Pi}^{(i)}$ and $\Pi^{(i)}$ denote the sets of predicted and gold evidence Spans (strings) for instance $i$.
A predicted Span $\hat{\pi}\in\widehat{\Pi}^{(i)}$ is counted as a \textbf{true positive (TP)} if it matches any gold Span $\pi\in\Pi^{(i)}$ such that
$\mathrm{Tok}(\pi)\subseteq \mathrm{Tok}(\hat{\pi})$ or
$\mathrm{Tok}(\hat{\pi})\subseteq \mathrm{Tok}(\pi)$ or
$\mathrm{Jaccard}(\hat{\pi},\pi)\ge 0.6$.
Spans in $\widehat{\Pi}^{(i)}$ that fail to match any gold Span are counted as \textbf{false positives (FP)},
and Spans in $\Pi^{(i)}$ not matched by any prediction are counted as \textbf{false negatives (FN)}.
Precision, recall, and F1-score are then computed as:

\begin{equation}
\begin{aligned}
\mathrm{precision}_{\mathrm{Span}}
&=
\frac{|\mathrm{TP}|}{|\mathrm{TP}+\mathrm{FP}|}
\\[1.0em]
\mathrm{recall}_{\mathrm{Span}}
&=
\frac{|\mathrm{TP}|}{|\mathrm{TP}+\mathrm{FN}|}
\\[1.0em]
F1_{\mathrm{Span}}
&=
\frac{
2 \times \mathrm{precision}_{\mathrm{Span}} \times \mathrm{recall}_{\mathrm{Span}}
}{
\mathrm{precision}_{\mathrm{Span}} + \mathrm{recall}_{\mathrm{Span}}
}.
\end{aligned}
\end{equation}

\subsection{SciERC}

\paragraph{Entity}

The Entity classification task is formulated as a multi-label classification
problem over a predefined set of scientific entity types. Let
$\widehat{\mathcal T}_{\text{Entity}}^{(i)}$ denote the predicted entity-type
set for instance $i$, and let $\mathcal T_{\text{Entity}}^{(i)}$ denote the
corresponding gold-standard entity-type set, where these sets are obtained by
projecting each entity record onto its label component $t_e$. We compute
precision, recall, and F1-score as follows:

\begin{equation}
\begin{gathered}
\mathrm{precision}_{\mathrm{Entity}}
=
\dfrac{
\sum_i
\left|
\widehat{\mathcal T}_{\mathrm{Entity}}^{(i)}
\cap
\mathcal T_{\mathrm{Entity}}^{(i)}
\right|
}{
\sum_i
\left|
\widehat{\mathcal T}_{\mathrm{Entity}}^{(i)}
\right|
}
\\[1.4em]
\mathrm{recall}_{\mathrm{Entity}}
=
\dfrac{
\sum_i
\left|
\widehat{\mathcal T}_{\mathrm{Entity}}^{(i)}
\cap
\mathcal T_{\mathrm{Entity}}^{(i)}
\right|
}{
\sum_i
\left|
\mathcal T_{\mathrm{Entity}}^{(i)}
\right|
}
\\[1.4em]
F1_{\mathrm{Entity}}
=
\dfrac{
2\,\mathrm{precision}_{\mathrm{Entity}}\,\mathrm{recall}_{\mathrm{Entity}}
}{
\mathrm{precision}_{\mathrm{Entity}}+\mathrm{recall}_{\mathrm{Entity}}
}.
\end{gathered}
\end{equation}

\paragraph{Span}
For Entity Span extraction, we evaluate each predicted evidence string against
gold Entity Spans using a \textbf{relaxed token-level matching strategy}, which
combines:
\begin{itemize}
    \setlength{\itemsep}{0pt}
    \setlength{\topsep}{2pt}
    \setlength{\parsep}{0pt}
    \item Full containment, i.e., the gold Span is fully included in the
    predicted Span or vice versa.
    \item Jaccard similarity between predicted and gold Spans, with a threshold
    of $0.6$.
\end{itemize}

Let $\widehat{\Pi}_e^{(i)}$ and $\Pi_e^{(i)}$ denote the sets of predicted and gold
evidence Spans for instance $i$, where each grounding set satisfies
$\Pi_e^{(i)} \subseteq \mathcal{S}(s_i)$ and anchors the structured record to the
source sequence $s_i$. 

For a predicted Span $\hat{\pi_e}\in\widehat{\Pi_e}^{(i)}$ and a gold Span
$\pi_e\in\Pi_e^{(i)}$, let $\mathrm{Tok}(\cdot)$ denote the set of tokens in the
Span. The relaxed Span-matching function is defined as:

\begin{equation}
\begin{aligned}
\operatorname{Match}_{\Pi_e}(\hat{\pi}_e,\pi_e)
=&
\mathbb{I}\left[
\mathrm{Tok}(\pi_e)\subseteq \mathrm{Tok}(\hat{\pi}_e)
\;\vee\;
\right.
\\&[0.15em]
\hspace{-0.4em}
\left.
\mathrm{Tok}(\hat{\pi}_e)\subseteq \mathrm{Tok}(\pi_e)
\;\vee\;
\mathrm{Jaccard}(\hat{\pi}_e,\pi_e)\ge 0.6
\right]
\end{aligned}
\end{equation}

A predicted Span $\hat{\pi}\in\widehat{\Pi}^{(i)}$ is counted as a
\textbf{true positive (TP)} if it matches any gold Span $\pi\in\Pi^{(i)}$ under
$\operatorname{Match}_{\Pi}$. Spans in $\widehat{\Pi}^{(i)}$ that fail to match
any gold Span are counted as \textbf{false positives (FP)}, and Spans in
$\Pi^{(i)}$ not matched by any prediction are counted as \textbf{false negatives
(FN)}. Precision, recall, and F1-score are then computed as:

\begin{equation}
\begin{gathered}
\mathrm{precision}_{\mathrm{Span}}
=
\dfrac{
|\mathrm{TP}_{\mathrm{Span}}|
}{
|\mathrm{TP}_{\mathrm{Span}}| + |\mathrm{FP}_{\mathrm{Span}}|
}
\\[1.4em]
\mathrm{recall}_{\mathrm{Span}}
=
\dfrac{
|\mathrm{TP}_{\mathrm{Span}}|
}{
|\mathrm{TP}_{\mathrm{Span}}| + |\mathrm{FN}_{\mathrm{Span}}|
}
\\[1.4em]
F1_{\mathrm{Span}}
=
\dfrac{
2\,\mathrm{precision}_{\mathrm{Span}}\,
\mathrm{recall}_{\mathrm{Span}}
}{
\mathrm{precision}_{\mathrm{Span}}
+
\mathrm{recall}_{\mathrm{Span}}
}.
\end{gathered}
\end{equation}

\paragraph{Relation}
Relation extraction is evaluated by jointly checking the relation type and the
ordered grounding Spans for the two relation arguments. Let a predicted relation
record be represented as
$\hat{r}=(\widehat{\Pi}^{(i)}_{r1},\hat{t},\widehat{\Pi}^{(i)}_{r2})$, where
$\widehat{\Pi}^{(i)}_{r1}$ and $\widehat{\Pi}^{(i)}_{r2}$ are the predicted grounding
sets for the first and second relation arguments, and $\hat{t}$ is the predicted
relation type. Similarly, let a gold relation record be represented as
$r=(\Pi^{(i)}_{r1},t,\Pi^{(i)}_{r2})$.

A predicted relation is considered correct only if the relation type matches and
both argument grounding sets match their corresponding gold grounding sets in
the correct order:

\begin{equation}
\begin{aligned}
\operatorname{Match}_{\mathrm{Relation}}(\hat{r},r)
=
\mathbb{I}\Big[
\hat{t}=t
&\wedge\;
\exists\,\hat{\pi}_{r1}\in\widehat{\Pi}^{(i)}_{r1},
\pi_{r1}\in\Pi^{(i)}_{r1}
\\&\text{s.t.}\;
\operatorname{Match}_{\Pi}(\hat{\pi}_{r1},\pi_{r1})=1
\wedge\;
\exists\,\hat{\pi}_{r2}\in\widehat{\Pi}^{(i)}_{r2},
\pi_{r2}\in\Pi^{(i)}_{r2}
\\&\text{s.t.}\;
\operatorname{Match}_{\Pi}(\hat{\pi}_{r2},\pi_{r2})=1
\Big].
\end{aligned}
\end{equation}
The ordering of relation arguments is strict: the predicted first argument
grounding set $\widehat{\Pi}^{(i)}_{r1}$ must match the gold first argument
grounding set $\Pi^{(i)}_{r1}$, and the predicted second argument grounding set
$\widehat{\Pi}^{(i)}_{r2}$ must match the gold second argument grounding set
$\Pi^{(i)}_{r2}$.

After one-to-one matching between predicted and gold relation records,
precision, recall, and F1-score are computed as:

\begin{equation}
\begin{gathered}
\mathrm{precision}_{\mathrm{Relation}}
=
\dfrac{
|\mathrm{TP}_{\mathrm{Relation}}|
}{
|\mathrm{TP}_{\mathrm{Relation}}| + |\mathrm{FP}_{\mathrm{Relation}}|
}
\\[1.4em]
\mathrm{recall}_{\mathrm{Relation}}
=
\dfrac{
|\mathrm{TP}_{\mathrm{Relation}}|
}{
|\mathrm{TP}_{\mathrm{Relation}}| + |\mathrm{FN}_{\mathrm{Relation}}|
}
\\[1.4em]
F1_{\mathrm{Relation}}
=
\dfrac{
2\,\mathrm{precision}_{\mathrm{Relation}}\,
\mathrm{recall}_{\mathrm{Relation}}
}{
\mathrm{precision}_{\mathrm{Relation}}
+
\mathrm{recall}_{\mathrm{Relation}}
}.
\end{gathered}
\end{equation}

\paragraph{Coreference}
Coreference is evaluated at the coreference-link level. Let a predicted
coreference link be represented as
$\hat{c}=(\widehat{\Pi}^{(i)}_{c1},\hat{d}_1,\widehat{\Pi}^{(i)}_{c2},\hat{d}_2)$,
where $\widehat{\Pi}^{(i)}_{c1}$ and $\widehat{\Pi}^{(i)}_{c2}$ are the predicted
grounding sets for the two linked mentions, and $\hat{d}_1$ and $\hat{d}_2$ are
their predicted sentence identifiers. Similarly, let a gold coreference link be
represented as $c=(\Pi^{(i)}_{c1},d_1,\Pi^{(i)}_{c2},d_2)$.

A predicted coreference link is considered correct only if both mention
grounding sets match under relaxed Span matching and both sentence identifiers
match exactly:
\begin{equation} \begin{aligned} \operatorname{Match}_{\text{Coref}}(\hat{c},c) &= \mathbb{I}\Big[ \exists \hat{\pi}_{c1}\in\widehat{\Pi}^{(i)}_{c1},\, \pi_{c1}\in\Pi^{(i)}_{c1}: \operatorname{Match}_{\Pi}(\hat{\pi}_{c1},\pi_{c1})=1 \;\wedge\; \hat{d}_1=d_1 \\ &\qquad\wedge\; \exists \hat{\pi}_{c2}\in\widehat{\Pi}^{(i)}_{c2},\, \pi_{c2}\in\Pi^{(i)}_{c2}:  \operatorname{Match}_{\Pi}(\hat{\pi}_{c2},\pi_{c2})=1 \;\wedge\; \hat{d}_2=d_2 \Big]. \end{aligned} \end{equation}
Thus, coreference evaluation requires relaxed textual agreement for both linked
mention grounding sets and strict agreement for their sentence identifiers.

After one-to-one matching between predicted and gold coreference links,
precision, recall, and F1-score are computed as:

\begin{equation}
\begin{gathered}
\mathrm{precision}_{\mathrm{Coreference}} =
\dfrac{
|\mathrm{TP}_{\mathrm{Coreference}}|
}{
|\mathrm{TP}_{\mathrm{Coreference}}|+|\mathrm{FP}_{\mathrm{Coreference}}|
}
\\[1.2em]
\mathrm{recall}_{\mathrm{Coreference}} =
\dfrac{
|\mathrm{TP}_{\mathrm{Coreference}}|
}{
|\mathrm{TP}_{\mathrm{Coreference}}|+|\mathrm{FN}_{\mathrm{Coreference}}|
}
\\[1.2em]
F1_{\mathrm{Coreference}} =
\dfrac{
2\,\mathrm{precision}_{\mathrm{Coreference}}\,
\mathrm{recall}_{\mathrm{Coreference}}
}{
\mathrm{precision}_{\mathrm{Coreference}}
+
\mathrm{recall}_{\mathrm{Coreference}}
}.
\end{gathered}
\end{equation}

\section{Related Work}
\label{app:related_work}
\paragraph{Preference-based alignment and offline preference optimization.}
Preference-based alignment has become a standard post-supervised fine-tuning strategy for adapting large language models to human or task-specific preferences. Early RLHF pipelines learn a reward model and optimize the policy with online reinforcement learning, but these pipelines require sampling during training and can be computationally expensive and difficult to stabilize \cite{ouyang2022training,bai2022constitutional}. Direct Preference Optimization (DPO) simplifies this pipeline by replacing explicit reward modeling and online RL with an offline, reference-regularized classification objective over preferred and rejected completions \cite{rafailov2023direct}. This formulation is effective for many open-ended generation settings, but it remains fundamentally sequence-level: the preference signal is assigned to entire completions rather than to the few tokens that determine structured correctness.

\paragraph{Sequence-level DPO variants.}
A large body of recent work improves DPO by modifying the sequence-level preference objective. IPO and the broader $\Psi$PO framework analyze preference optimization from a theoretical perspective and replace the logistic classification target with a more stable margin-regression objective \cite{gheshlaghiAzar2024psiPO}. SimPO removes the explicit reference model and uses average sequence log-likelihood as an implicit reward, improving simplicity and length behavior \cite{meng2024simpo}. Cal-DPO argues that contrastive preference objectives can ignore the absolute scale of implicit rewards and introduces calibration to make learned rewards more comparable to target rewards \cite{xiao2024caldpo}. DPO-Positive / DPOP addresses the observation that DPO can increase the relative preference margin while degrading the likelihood of the chosen response, adding a correction that protects preferred completions \cite{pal2024dpop}. These methods address important sequence-level pathologies such as reference dependence, margin instability, calibration drift, and chosen-likelihood degradation. However, they still operate primarily through completion-level scalar margins. In ontology-driven structured prediction, this is insufficient because the preferred and rejected outputs are often near-identical serialized objects whose correctness differs in only a small number of schema-defining tokens.

\paragraph{Low-edit-distance structured preferences.}
Structured extraction tasks differ from open-ended alignment tasks because small local errors can invalidate an otherwise well-formed output. In PV-Miner, a single Code or Sub-code token can change the clinical-communication interpretation of a grounded span. In SciERC, a single entity type, relation label, sentence identifier, or coreference link can determine whether a structured record is correct. These tasks therefore induce low-edit-distance preference pairs: the chosen and rejected completions share most JSON scaffolding, field names, punctuation, and repeated formatting tokens, while differing only at sparse semantic decision points. Under standard DPO, the post-divergence gradient can be spread across both critical and non-critical tokens. This creates gradient dilution, where optimization mass is spent on serialization tokens that do not determine task F1, and preferred-token erosion, where the aggregate preference margin can improve even as the likelihood of rare but correct schema tokens decreases. TAB-PO is designed for this regime rather than for generic response-level preference separation.

\paragraph{Token-level preference optimization.}
Token-level preference optimization methods move beyond purely
completion-level objectives. TDPO decomposes preference optimization at the
token level and introduces token-wise forward KL constraints to improve the
balance between alignment and generation diversity
\cite{zeng2024tdpo}. TI-DPO further argues that tokens differ in
importance and estimates token-importance weights using gradient attribution,
together with an additional triplet-style loss that pulls the policy toward
preferred responses and away from rejected responses
\cite{yang2025tidpo}. These approaches are closer to TAB-PO than
sequence-level DPO variants because they recognize that token positions
contribute unequally to preference learning. However, their token-level
mechanisms are not specialized for ontology-constrained structured outputs.
TDPO is primarily a token-level KL reformulation for general alignment, while
TI-DPO infers token importance from model-derived attribution signals. TAB-PO
instead uses the structure of the task itself: it constructs minimally
perturbed, schema-valid rejected outputs from empirical SFT confusions, and it
applies a confidence-gated barrier only to preferred tokens whose current
likelihood falls below a threshold. This distinction is reflected in the
training diagnostics: TAB-PO achieves stronger preference-probability
convergence while maintaining a more stable reference-adjusted margin, rather
than increasing the sequence-level margin by sacrificing under-confident
schema tokens. Thus, TAB-PO does not merely ask which tokens are important in
general; it asks which schema-defining preferred tokens are currently at risk
of probability erosion and protects those tokens during preference learning.

\paragraph{Verifier-based RL and GRPO-style alternatives.}
Recent reasoning-oriented alignment methods use reinforcement learning with
verifiable rewards, including GRPO-style optimization for mathematical
reasoning \cite{shao2024deepseekmath}. More generally, RLVR is effective
when candidate outputs can be scored by reliable programmatic or binary
verifiers, such as final-answer checks in mathematics, executable tests in
code, or other domains with objective outcome-level rewards
\cite{wen2025rlvr}. Ontology-driven clinical and scientific annotation has a
different supervision structure. The gold output is available, but the
clinically or semantically meaningful distinction is often local,
hierarchical, and ambiguity-sensitive rather than reducible to a single
scalar reward. In this setting, expert-adjudicated and confusion-aware
preference pairs provide a more targeted signal than online scalar rewards:
they encode exactly which structured alternative should be preferred and
preserve the low-edit-distance contrast that exposes label, span, relation,
and coreference errors. TAB-PO also has a practical computational advantage:
after supervised fine-tuning, its preference data can be prepared at
near-zero marginal cost from existing gold records, SFT validation errors,
and empirical confusion tables, without online rollout sampling, reward-model
training, external verifier execution, or group-based policy optimization.
TAB-PO is therefore complementary to verifier-based RL, but it is tailored to
the offline post-SFT setting where gold structured records and empirical
model confusions are already available.

\paragraph{Positioning of TAB-PO.}
TAB-PO differs from prior preference-optimization methods along two axes.
First, its preference data are confusion-aware: rejected outputs are
minimally perturbed, schema-valid alternatives sampled from the SFT model's
empirical ontology-level confusions and expert-observed ambiguity patterns.
This makes optimization focus on realistic structured errors rather than
arbitrary negative completions. Second, its objective is token-critical: the
confidence-gated barrier restores likelihood only for under-confident
preferred tokens, preserving the \emph{critical schema-defining tokens} that
determine structured correctness, including labels, spans, relation
arguments, sentence identifiers, and coreference links. This protection is
localized: confident tokens can still support preference separation, while
under-confident semantic tokens are prevented from eroding below a useful
likelihood floor. This combination directly addresses the two failure modes
that arise in low-edit-distance structured preferences: gradient dilution
over non-critical serialization tokens and erosion of rare preferred schema
tokens. As a result, TAB-PO is not simply another DPO variant for general
alignment; it is a post-SFT preference objective specialized for
ontology-constrained structured generation.

\section{Targeted TAB-PO Negative Construction}
\label{append:nega_con}

This appendix illustrates how TAB-PO constructs targeted negative outputs for
ontology-driven structured prediction. Given an input $x$ and gold structured
output $Y^{+}$, TAB-PO constructs a rejected output $Y^{-}$ by introducing a
small, schema-valid perturbation that reflects a realistic structured prediction
error. The goal is not to create arbitrary corrupted outputs, but to produce
low-separation hard negatives: outputs that remain parseable and ontology-valid
while differing from the gold output in the semantic label, textual grounding,
or relational-linking decision that determines correctness. Figures~\ref{fig:pvminer_negative_construction},
\ref{fig:scierc_negative_construction}, and
\ref{fig:scierc_negative_construction-1} below illustrate representative perturbations for
PV-Miner and SciERC.

\definecolor{semconf}{HTML}{C62828}
\definecolor{missrec}{HTML}{00897B}
\definecolor{extrarec}{HTML}{1565C0}

\newcommand{\perturbsection}[2]{%
\vspace{0.55em}
\begin{tcolorbox}[
  enhanced,
  colback=#1!6!white,
  colframe=#1!65!black,
  boxrule=0.45pt,
  arc=1.2mm,
  left=1.0mm,
  right=1.0mm,
  top=0.7mm,
  bottom=0.7mm,
  borderline west={1.6pt}{0pt}{#1!85!black}
]
{\sffamily\footnotesize\bfseries #2}
\end{tcolorbox}
\vspace{0.15em}
}

\newcommand{\perturbnote}[2]{%
\begin{tcolorbox}[
  enhanced,
  colback=#1!7!white,
  colframe=#1!45!black,
  boxrule=0.35pt,
  arc=1.0mm,
  left=1.0mm,
  right=1.0mm,
  top=0.55mm,
  bottom=0.55mm,
  borderline west={1.2pt}{0pt}{#1!80!black}
]
{\sffamily\footnotesize\emph{#2}}
\end{tcolorbox}
}

\newtcolorbox{dpobox}[1][]{
  enhanced,
  colback=teal!3!white,
  colframe=teal!60!black,
  colbacktitle=teal!85!black,
  fonttitle=\bfseries\sffamily,
  title={Targeted TAB-PO negative construction for PV Miner},
  boxrule=0.9pt,
  arc=2.2mm,
  left=1.2mm,right=1.2mm,top=1.0mm,bottom=1.0mm,
  #1
}

\begin{figure}[!h]
\centering
\begin{adjustbox}{max totalsize={0.96\linewidth}{0.96\textheight},center}
\begin{minipage}{0.96\linewidth}

\begin{dpobox}

{\sffamily\footnotesize
\textbf{Legend:}\;
\textcolor{semconf}{\rule{0.9em}{0.9em}}\;Semantic labelling confusion\hspace{0.9em}
\textcolor{missrec}{\rule{0.9em}{0.9em}}\;Missing record perturbation\hspace{0.9em}
\textcolor{extrarec}{\rule{0.9em}{0.9em}}\;Extra record perturbation
}

\vspace{0.6em}
{\sffamily\footnotesize
\textbf{Metadata:} TO\_PAT\_YN=Y \textit{(Provider$\rightarrow$Patient)}\\
\textbf{Input message:} \emph{No worries! Thank you for letting me know. Best wishes, Dr.\ Person1}
}

\vspace{0.6em}
\hrule
\vspace{0.6em}

{\sffamily\footnotesize\textbf{Ground truth:}}
\begin{alltt}\footnotesize
[
  \{Code: PartnershipProvider, Sub-code: connection, Span: "No worries!"\},
  \{Code: PartnershipProvider, Sub-code: Appreciation/Gratitude,
                               Span: "Thank you for letting me know"\},
  \{Code: PartnershipProvider, Sub-code: signoff, Span: "Best wishes"\}
]
\end{alltt}

\perturbsection{semconf}{1) Semantic labelling confusion}

\perturbnote{semconf}{
A semantic label is perturbed while the grounded span is retained.
Code--Sub-code mapping is preserved under Semantic labelling confusion.
}

\begin{alltt}\footnotesize
[
  \{Code: \textcolor{semconf}{PartnershipPatient}, Sub-code: connection, Span: "No worries!"\},
  \{Code: PartnershipProvider, Sub-code: \textcolor{semconf}{signoff},
                               Span: "Thank you for letting me know"\},
  \{Code: PartnershipProvider, Sub-code: signoff, Span: "Best wishes"\}
]
\end{alltt}

\perturbsection{missrec}{2) Missing record perturbation}

\perturbnote{missrec}{
A valid ground-truth record is removed from the structured output.
}

\begin{alltt}\footnotesize
[
  \{Code: PartnershipProvider, Sub-code: connection, Span: "No worries!"\},
  \{Code: PartnershipProvider, Sub-code: Appreciation/Gratitude,
                               Span: "Thank you for letting me know"\}
]
\end{alltt}

{\sffamily\footnotesize
\textbf{Removed:}
\textcolor{missrec}{\texttt{\{Code: PartnershipProvider, Sub-code: signoff, Span: "Best wishes"\}}}
}

\perturbsection{extrarec}{3) Extra record perturbation}

\perturbnote{extrarec}{
An additional validly formatted but unsupported record is inserted.
The Span for the extra record is chosen from the input message.
}

\begin{alltt}\footnotesize
[
  \{Code: PartnershipProvider, Sub-code: connection, Span: "No worries!"\},
  \{Code: PartnershipProvider, Sub-code: Appreciation/Gratitude,
                               Span: "Thank you for letting me know"\},
  \{Code: PartnershipProvider, Sub-code: signoff, Span: "Best wishes"\},
  \textcolor{extrarec}{\{Code: SDOH, Sub-code: SocialAndCommunityContext, Span: "No worries"\}}
]
\end{alltt}

\end{dpobox}

\end{minipage}
\end{adjustbox}
\caption{
PV-Miner example of targeted TAB-PO negative construction.
}
\label{fig:pvminer_negative_construction}
\end{figure}

\clearpage

\definecolor{sciercback}{HTML}{FFF8E1}
\definecolor{sciercframe}{HTML}{8D6E63}
\definecolor{scierctitle}{HTML}{6D4C41}

\definecolor{scisemconf}{HTML}{C62828}
\definecolor{scirelconf}{HTML}{6B8E23}
\definecolor{scimissrec}{HTML}{00897B}
\definecolor{sciextrarec}{HTML}{1565C0}

\newcommand{\sciperturbsection}[2]{%
\vspace{0.55em}
\begin{tcolorbox}[
  enhanced,
  colback=#1!8!white,
  colframe=#1!65!black,
  boxrule=0.45pt,
  arc=1.2mm,
  left=1.0mm,
  right=1.0mm,
  top=0.65mm,
  bottom=0.65mm,
  borderline west={1.7pt}{0pt}{#1!85!black}
]
{\sffamily\footnotesize\bfseries #2}
\end{tcolorbox}
\vspace{0.15em}
}

\newcommand{\sciperturbnote}[2]{%
\begin{tcolorbox}[
  enhanced,
  colback=#1!7!white,
  colframe=#1!45!black,
  boxrule=0.35pt,
  arc=1mm,
  left=1.0mm,
  right=1.0mm,
  top=0.55mm,
  bottom=0.55mm,
  borderline west={1.2pt}{0pt}{#1!80!black}
]
{\sffamily\footnotesize\emph{#2}}
\end{tcolorbox}
}

\newtcolorbox{sciercdpobox}[1][]{
  enhanced,
  colback=sciercback,
  colframe=sciercframe,
  colbacktitle=scierctitle,
  fonttitle=\bfseries\sffamily,
  title={Targeted TAB-PO negative construction for SciERC},
  boxrule=0.9pt,
  arc=2.2mm,
  left=1.2mm,right=1.2mm,top=1.0mm,bottom=1.0mm,
  #1
}

\newtcolorbox{sciercdpoboxcontinued}[1][]{
  enhanced,
  colback=sciercback,
  colframe=sciercframe,
  boxrule=0.9pt,
  arc=2.2mm,
  left=1.2mm,right=1.2mm,top=1.0mm,bottom=1.0mm,
  #1
}

\begin{figure}[!h]
\centering
\begin{adjustbox}{max totalsize={0.98\linewidth}{0.96\textheight},center}
\begin{minipage}{0.98\linewidth}

\begin{sciercdpobox}

{\sffamily\footnotesize
\textbf{Legend:}\;
\textcolor{scisemconf}{\rule{0.9em}{0.9em}}\;Semantic labelling confusion\hspace{0.7em}
\textcolor{scirelconf}{\rule{0.9em}{0.9em}}\;Relational linking confusion\hspace{0.7em}
\textcolor{scimissrec}{\rule{0.9em}{0.9em}}\;Missing record perturbation\hspace{0.7em}
\textcolor{sciextrarec}{\rule{0.9em}{0.9em}}\;Extra record perturbation
}

\vspace{0.6em}
{\sffamily\footnotesize

\textbf{Input sample:} \emph{GraphParser improves parsing. GraphParser is effective.}
}

\vspace{0.6em}
\hrule
\vspace{0.6em}

{\sffamily\footnotesize\textbf{Ground-truth:}}
\begin{alltt}\footnotesize
\{
  entities: [
    \{Span: GraphParser, Type: Method\},
    \{Span: parsing, Type: Task\}
  ],
  relations: [
    \{Span 1: GraphParser, Type: Used-for, Span 2: parsing\}
  ],
  coreference: [
    \{Span 1: GraphParser, Sentence ID: 0,
     Span 2: GraphParser, Sentence ID: 1\}
  ]
\}
\end{alltt}
\sciperturbsection{scisemconf}{1) Semantic labelling confusion}
\sciperturbnote{scisemconf}{
A semantic label is perturbed while the grounded span and structured record are retained.
For SciERC, this can perturb an entity type, or relation type.
}

\begin{alltt}\footnotesize
\{
  entities: [
    \{Span: GraphParser, Type: Method\},
    \{Span: parsing, Type: \textcolor{scisemconf}{Method}\}
  ],
  relations: [
    \{Span 1: GraphParser, Type: \textcolor{scisemconf}{Evaluate-for}, Span 2: parsing\}
  ],
  coreference: [
    \{Span 1: GraphParser, Sentence ID: 0,
     Span 2: GraphParser, Sentence ID: 1\}
  ]
\}
\end{alltt}

\sciperturbsection{scirelconf}{2) Relational linking confusion}

\sciperturbnote{scirelconf}{
A relational link is perturbed while the spans remain grounded in the input. This may involve reversing Span 1 and Span 2, confusing one linked span with another entity span, or modifying a coreference link by substituting another entity span with a valid sentence ID.
}

\begin{alltt}\footnotesize
\{
  entities: [
    \{Span: GraphParser, Type: Method\},
    \{Span: parsing, Type: Task\}
  ],
  relations: [
    \{Span 1: \textcolor{scirelconf}{parsing}, Type: Used-for, Span 2: \textcolor{scirelconf}{GraphParser}\}
  ],
  coreference: [
    \{Span 1: \textcolor{scirelconf}{parsing}, Sentence ID: 0,
     Span 2: GraphParser, Sentence ID: 1\}
  ]
\}
\end{alltt}

\end{sciercdpobox}

\end{minipage}
\end{adjustbox}
\caption{
SciERC semantic-label and relational-link confusion examples for TAB-PO
negative construction.
}
\label{fig:scierc_negative_construction}
\end{figure}

\vspace{0.6em}

\begin{figure}[!h]
\centering
\begin{adjustbox}{max totalsize={0.98\linewidth}{0.96\textheight},center}
\begin{minipage}{0.98\linewidth}

\begin{sciercdpoboxcontinued}

\sciperturbsection{scimissrec}{3) Missing record perturbation}

\sciperturbnote{scimissrec}{
A valid ground-truth record is removed from the structured output. For SciERC, this may involve removing an entity, relation, or coreference link; if an entity is removed, its associated relations and coreference links are also removed to maintain structural validity.
}

\begin{alltt}\footnotesize
\{
  entities: [
    \{Span: GraphParser, Type: Method\}
  ],
  relations: [
    \{Span 1: GraphParser, Type: Used-for, Span 2: parsing\} 
  ],
  coreference: [
    \{Span 1: GraphParser, Sentence ID: 0,
     Span 2: GraphParser, Sentence ID: 1\}
  \\]
\}
\end{alltt}

{\sffamily\footnotesize
\textbf{Removed:}
\textcolor{scimissrec}{\{Span: parsing, Type: Task\}}
}

\sciperturbsection{sciextrarec}{4) Extra record perturbation}

\sciperturbnote{sciextrarec}{
An additional validly formatted but unsupported record is inserted.
For entity-level extra records, the span is chosen from the input text.
}

\begin{alltt}\footnotesize
\{
  entities: [
    \{Span: GraphParser, Type: Method\},
    \{Span: parsing, Type: Task\},
    \textcolor{sciextrarec}{\{Span: effective, Type: Metric\}}
  ],
  relations: [
    \{Span 1: GraphParser, Type: Used-for, Span 2: parsing\}
  ],
  coreference: [
    \{Span 1: GraphParser, Sentence ID: 0,
     Span 2: GraphParser, Sentence ID: 1\}
  ]
\}
\end{alltt}

\end{sciercdpoboxcontinued}

\end{minipage}
\end{adjustbox}
\caption{
SciERC missing-record and extra-record perturbation examples for TAB-PO
negative construction.}
\label{fig:scierc_negative_construction-1}
\end{figure}

\clearpage

PV-Miner focuses on hierarchical clinical annotation, where
negative construction targets Code/Sub-code confusions, missing grounded
records, and unsupported extra records. SciERC extends the same idea to
scientific relational extraction, where perturbations can affect entity labels,
relation labels, relation arguments, coreference links, missing records, and
extra records. Across all perturbation types, the rejected output remains close to the
preferred output while changing the specific ontology-bound decision that
determines correctness. TAB-PO is designed to overcome the disadvantage
of low-separation structured preferences: it converts preference optimization
into a targeted signal for schema-critical tokens rather than allowing the
update to be diluted across shared serialization structure.

\newpage

\endgroup

\end{appendices}

\end{document}